\documentclass[lettersize,journal]{IEEEtran}
\usepackage{algorithmic}
\usepackage{graphicx}
\usepackage{graphics}
\usepackage{textcomp}
\usepackage{subfigure}
\usepackage{tabularx}
\usepackage{times}
\usepackage{multirow}
\usepackage{bm}
\usepackage{latexsym}
\usepackage{booktabs}
\usepackage{threeparttable}
\usepackage{epsf}
\usepackage[vlined,ruled,linesnumbered]{algorithm2e}
\usepackage{color,colortbl}

\usepackage{cite}
\usepackage{picinpar}
\usepackage{amsmath}
\usepackage{url}    
\usepackage[latin1]{inputenc}
\usepackage{soul}
\usepackage{multirow}
\usepackage{pifont}
\usepackage{alltt}
\usepackage{hyperref}
\usepackage{enumerate}
\usepackage{siunitx}
\usepackage{breakurl}
\usepackage{epstopdf}
\usepackage{pbox}
\usepackage{amsmath,amssymb,amsfonts}
\usepackage{amsthm}
\usepackage{mathrsfs}
\usepackage{makecell}
\usepackage{pbox}
\usepackage{supertabular}
\usepackage{multicol}
\usepackage{lipsum}
\usepackage{leftindex}
\newtheorem{theorem}{Theorem}

\newtheorem{remark}{Remark}
\newtheorem{lemma}{Lemma}

\newtheorem{assumption}{Assumption}
\newcommand{\blue}{\textcolor{blue}}
\renewcommand{\arraystretch}{1.5}
\begin{document}
	
	\title{	Relative Pose Estimation for Nonholonomic Robot Formation with UWB-IO Measurements }
	\author{
		\vskip 1em
		Kunrui Ze, 
		Wei Wang, \emph{IEEE Senior Member},
		Shuoyu Yue, 
		Guibin Sun,
		Kexin Liu,
		{Jinhu L\"u}, \emph{IEEE Fellow}
		\thanks{
			%Manuscript received April 19, 2021; revised August 16, 2021.
			This work was supported in part by the National Natural Science Foundation of China under Grants 625B2017, 62141604 and 62373019, in part by the National Key Research and Development Program of China under Grant 2022YFB3305600, in part by the Fundamental Research Funds for the Central Universities under Grant 501XYGG2025103037 and 501XSKC2025103001.
			\emph{(Corresponding authors: Kexin Liu.)}}
		\thanks{
			Kunrui Ze and Guibin Sun are with the School of Automation Science and Electrical Engineering, Beihang University, Beijing 100191, China (e-mail: kr\_ze@buaa.edu.cn; sunguibinx@buaa.edu.cn).
			
			Shuoyu Yue is with the Cavendish Laboratory, University of Cambridge, CB3 0DS, United Kingdom (e-mail:sy481@cam.ac.uk).
			
			Wei Wang, Kexin Liu and {Jinhu L\"u} are with the School of Automation Science and Electrical Engineering, Beihang University, Beijing 100191, China, and also with the Zhongguancun Laboratory, Beijing 100083, China (e-mail: w.wang@buaa.edu.cn; kxliu@buaa.edu.cn; jhlu@iss.ac.cn).}
	}
	\maketitle
	\begin{abstract}
		This article studies the problem of distributed formation control for multiple robots by using onboard Ultra Wide Band (UWB) distance and inertial odometer (IO) measurements. 
		Although this problem has been widely studied, a fundamental limitation of most works is that they require each robot's pose and sensor measurements are expressed in a common reference frame.
		However, it is inapplicable for nonholonomic robot formations due to the practical difficulty of aligning IO measurements of individual robot in a common frame.
		To address this problem, firstly, a concurrent-learning based estimator is firstly proposed to achieve relative localization between neighboring robots in a local frame. 
		Different from most relative localization methods in a global frame, both relative position and orientation in a local frame are estimated with only UWB ranging and IO
		measurements.
		Secondly, to deal with information loss caused by directed communication topology, a cooperative localization algorithm is introduced to estimate the relative pose to the leader robot.
		Thirdly, based on the theoretical results on relative pose estimation, a distributed formation tracking controller is proposed for nonholonomic robots.
		Both 2D and 3D real-world experiments conducted on aerial robots and grounded robots are provided to demonstrate the effectiveness of the proposed method.
	\end{abstract}
	
	
	\markboth{IEEE/ASME TRANSACTIONS ON MECHATRONICS}%
	{}
	
	\definecolor{limegreen}{rgb}{0.2, 0.8, 0.2}
	\definecolor{forestgreen}{rgb}{0.13, 0.55, 0.13}
	\definecolor{greenhtml}{rgb}{0.0, 0.5, 0.0}
	
	\section{Introduction}
	
	Distributed formation control for robot swarms has received significant attention in recent years  due to its wide range of potential applications \cite{Sun2025IEM, Zhao2023TM,xu2023cooperative,li2024urban}. 
	The objective of formation control is to guide robot swarms from an initial configuration to a specified formation through distributed interactions.
	To achieve this objective, various methods have been explored, including those based on consensus \cite{Ze2023TIE}, artificial light field \cite{Chu2023TASE} and mean-shift exploration \cite{Sun2023NC}. 
	However, most of these methods rely on external localization systems, such as GPS, which are not well-suited for practical applications.
	Moreover, external localization systems may be unavailable in certain scenarios, such as indoor or underground environments.  
	In summary, compared to relying on external localization systems, robots that depend solely on onboard measurements are more autonomous.
	
	Extensive researches adopted sensor measurements to estimate relative poses between neighboring robots.
	Based on the type of sensor utilized, formation control methods relying on local measurements can be roughly classified into three categories: relative-position, relative-bearing (or angle), and relative-distance based methods. 
	For the first class methods, depth camera or LADAR can be used to achieve ralative measurements between neighboring robots. 
	Distributed formation control task can be accomplished when the inter-robot measurement topology satisfies specific conditions such as
	bearing rigidity \cite{Zhao2019TAC} or non-collinear neighors condition \cite{Wang2024Auto}.
	To relax the topology condition, recent work \cite{Ning2024IJRR} enhance the observability of relative pose of neighbor robots by more effectively utilizing visual measurement information.
	However, all these methods require expensive and complex sensors and object detection algorithms, significantly increasing both complexity and hardware costs. 
	Additionally, these methods face challenges like limited field of view and high environmental requirements, including lighting conditions \cite{Xie2023TRO}.
	
	\begin{figure}[!t]\centering
		\includegraphics[scale=1.066]{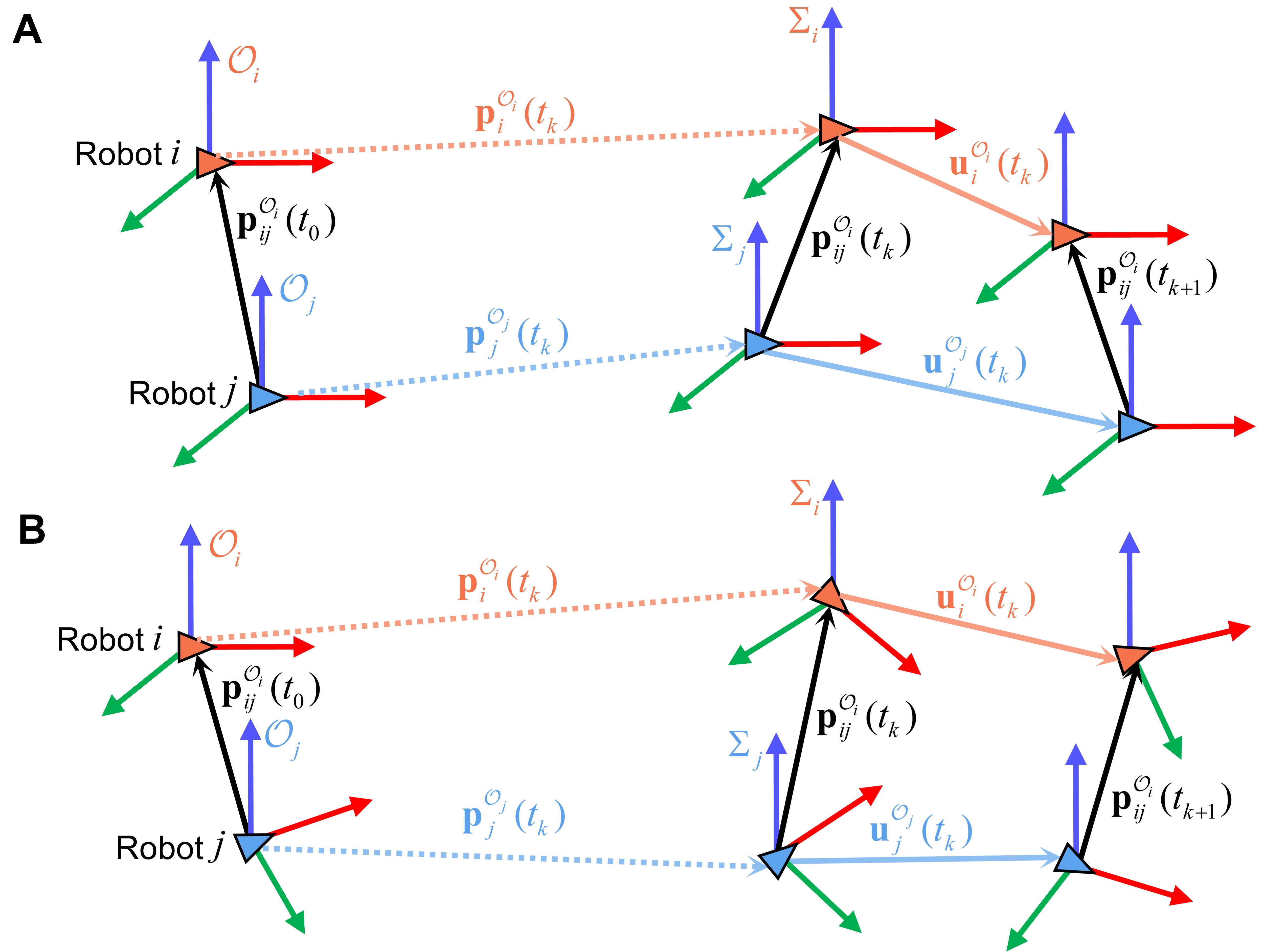}
		\caption{Geometric relationship between the displacement and UWB distance measurements of the two robots. A. All the robot have a global orientation \cite{Xie2019TCNS,dong2025tmech,Liu2023Auto,xiong2025adaptive,Chen2023TIE,Liu2023RAL}. B. Each robot is in its local frame.}\centering
		\label{rl_figure}
		\vspace{-0.5cm}
	\end{figure}
	
	Distance measurement sensors like Ultra Wide Band (UWB) inherently address the challenges \cite{Mehdifar2020PPB, Chen2024TCNS}. 
	One of the primary difficulties of formation control that relies on distance measurements is accurately estimating.
	To estimate relative position, traditional methods typically derive the leader's relative position using distance measurements from multiple neighbors \cite{lv2022TNNLS}. However, this method inevitably imposes conditional constraints on the measurement topology, such as the requirement of rigidity.
	Similarly, a common strategy is to place multiple UWB tags on each robot \cite{Cossette2022IROS,Shalaby2024TRO} or install UWB anchors in the environment \cite{cano2023TRO}, which may sacrifice the scalability of the robot swarm.
	Another strategy involves introducing the inertial odometer (IO) measurement to achieve relative localization among robots \cite{Xie2019TCNS}. 
	Using a relative position estimator based on UWB distance and local IO measurements, distributed formation control of robot swarms has been achieved in \cite{dong2025tmech, Chen2023TIE, Liu2023Auto, Ze2025TASE}.
	To handle the sensor characteristics, an abnormal data removal mechanism is proposed in \cite{Chen2023TIE}.
	
	All the aforementioned methods assume that the IO frames of all robots are directionally aligned, as illustrated in Fig. \ref{rl_figure} A. 
	This assumption is usually achieved by equipping robots with sensors such as compasses, which are highly susceptible to interference from ambient magnetic field variations \cite{silic2019correcting}. 
	For a nonholonomic robot with its own local frame, the orientation of this frame changes dynamically as the robot moves, resulting in no consistently aligned direction. 
	As a result, relative pose estimation between nonholonomic robots must be conducted within local frame, as shown in Fig. \ref{rl_figure} B, which is challenging relying solely on relative distance measurements.
	Fortunately, by increasing the IO displacement of each robot within its odometer frame, the relative yaw angle between the two robots can be estimated \cite{Xie2023TRO}.
	The core concept involves formulating an optimization problem that relates real-time distance measurements and the relative pose between robots.
	%In these methods, only UWB distance and local IO measurements are adopted to estimate relative pose estimation between two robots.
	Nevertheless, all these methods \cite{Liu2023Auto,Wang2023TM,Xie2023TRO} require that the relative motion between robots satisfy the persistently excited (PE) condition to ensure the convergence of the relative localization estimators.
	In other words, these methods require that robots maintain a non-zero relative velocity at all times. This is typically achieved through the introduction of continuous excitation noise \cite{Liu2023Auto} and periodic oscillation signals \cite{Xie2019TCNS,Xie2023TRO} of varying amplitudes into the control inputs of the robots. 
	The disadvantage lies in the contradiction between the motion needed for PE localization enhancement and relatively stationary formation control tasks, which can result in uneven motion trajectories and diminish control performance.
	To address this issue, a rotating UWB tag has been used to achieve the PE condition \cite{Liu2023RAL}. 
	However, this solution significantly increases the complexity of the sensor equipment on each robot and raises the failure rate. 
	So far, 
	there are few studies focused on the implementation of distributed formation control for multiple nonholonomic robots using only UWB and IO measurements. 
	
	\blue{The primary challenge lies in achieving distributed pose estimation based on relative distance measurements across different local frames without necessitating the PE condition.
		To address the above challenge, the contributions can be summarized as follows.
		First, %the proposed distributed formation control law relies solely on inter-robot distance measurements. 
		unlike existing formation control methods for nonholonomic robots that require knowledge of inter-robot relative positions \cite{Zhao2023TM,Wang2024Auto,Ma2024TCST} or relative orientations \cite{Wang2024Auto, Zou2024Auto}, our method utilizes only UWB distance and IO measurements.}
	This reduces the sensor requirements for each robot and broadens applicability across various scenarios.
	Second, to facilitate relative pose estimation between robots in a local frame, a concurrent-learning based relative pose estimator is proposed. 
	Different from recent methods\cite{Liu2023Auto, Zhao2023TM} that estimate relative position or orientation in a global frame, our approach focuses on estimating relative pose in a local context. 
	%In contrast to PE relative motion based methods \cite{Liu2023Auto, Wang2023TM}, our method effectively incorporates both current and historical information to relax the PE condition to a sufficiently excited (SE) condition in the relative pose estimation problem. 
	%This can effectively improve the smoothness of robot trajectories and enhance control performance. 
	\blue{In contrast to PE-dependent relative-motion methods \cite{Liu2023Auto, Wang2023TM}, our approach integrates online measurement information with selected historical data in the relative pose estimation process. This allows the convergence condition to be characterized by a sufficiently excited (SE) condition rather than the classical persistent excitation (PE) condition. Consequently, the estimator places less demand on sustained inter-robot excitation, which is beneficial for achieving smoother motion and better closed-loop control performance.}
	Additionally, the proposed data-driven methods demonstrate superior noise robustness compared to those relying solely on current information \cite{Liu2023Auto}.
	We provide both theoretical analysis and real-world experiments involving mobile robots and drones are provided to validate the effectiveness of the proposed methods.
	
	The rest of this article is organized as follows. In Section II, the distributed formation control task for nonholonomic robots is formulated. In Section III, the main results for inter-robot relative pose estimation, leader relative pose estimator design and the distributed formation controller are introduced, followed by the performance evaluation and experimental results in Section IV. Finally, Section V concludes this article.
	
	\textbf{Notations:} A transformation matrix in frame $\{A\}$ is denoted as $\textbf{T}^A :=  \begin{bmatrix}
		\textbf{R}^A, & \textbf{p}^A \\
		\textbf{0}^{T}, & 1 \\
	\end{bmatrix}  \in SE(3)$,
	where $\textbf{p}^A \in \mathcal{R}^{3}$ and $\textbf{R}^A \in SO(3)$ are the position vector and rotation matrix in frame $\{A\}$, respectively. Denote $\textbf{T}^A_B$ and $\textbf{R}^A _B$ as the transformation and rotation matrices from frame $\{B\}$ to frame $\{A\}$, respectively. 
	For vector $\textbf{x}$, $x[i]$ and $x[i:j]$ denote the $i$-th element and the elements from $i$ to $j$ of $\textbf{x}$.
	For ease of reading, we list the main symbols of this article in Appendix \ref{Technical details}-A.
	
	\section{Preliminaries and problem formulation}
	
	%\begin{figure}[!t]\centering
	%	\includegraphics[scale=1.066]{figures/system.jpg}
	%	\caption{Control object.}\centering
	%	\label{notations}
	%	\vspace{-0.5cm}
	%\end{figure}
	
	\subsection{Robot Models}
	
	Consider a group of $N+1$ nonholonomic robots in $\mathbb{R}^2$ or $\mathbb{R}^3$. For $i=0,...,N$, the kinematics of robot $i$ is described as
	\begin{align}
		\label{model}
		\left\{
		\begin{aligned}
			\dot{\textbf{p}}^{\mathcal{O}_{i}}_{i}(t) &=  \begin{bmatrix}
				\cos \phi_{\Sigma_{i}}^{\mathcal{O}_{i}}(t),& 0\\
				\sin \phi_{\Sigma_{i}}^{\mathcal{O}_{i}}(t),& 0\\
				0,& 1
			\end{bmatrix} 
			\begin{bmatrix}
				v_{i,h} \\
				v_{i,z}
			\end{bmatrix}\\
			\dot{\phi}_{\Sigma_{i}}^{\mathcal{O}_{i}}(t) &= w_{i}.
		\end{aligned}
		\right.
	\end{align}
	For each robot $i$, denote the odometry frame as $\mathcal{O}_{i}$. 
	The $x$-axis of $\mathcal{O}_{i}$ is the initial orientation of robot $i$, as shown in Fig. \ref{rl_figure} B.
	At time $t$, denote the IMU body frame as $\Sigma_{i}(t)$. 
	Similarly, the $x$-axis of $\Sigma_{i}(t)$ is the defined as the real-time orientation of robot $i$. 
	$\textbf{p}^{\mathcal{O}_{i}}_{i}(t)$ is the displacement in its odometry frame $\mathcal{O}_{i}$ from $t_{0}=0$ to $t$, which can be measurement with IO. $v_{i,h}$ and $v_{i,z}$ are the horizontal and vertical velocities of robot $i$.
	Similarly, $\phi_{\Sigma_{i}}^{\mathcal{O}_{i}}(t)$ is the angular displacements of robot $i$ in its odometry frame $\mathcal{O}_{i}$ at time $t$, which is also measurable by IO.
	
	Note that the roll and pitch angles of the robot can be accurately measured with the IMU-based SLAM methods \cite{Xie2023TRO}. We focus on the 4-DoF robot described by Eq.(\ref{model}), which can well describe both 2D and 3D nonholonomic robots.
	
	\subsection{UWB Measurement and Communication Topology}
	Each robot $i$ is equipped with a UWB module to measure the real-time distance and communicate with each other. 
	Define the position of robot $j$ in frame $\mathcal{O}_{i}$ at time $t$ as $\textbf{p}^{\mathcal{O}_{i}}_{j}(t)$. Then the relative position between robot $i$ and robot $j$ in frame $\mathcal{O}_{i}$ can be denoted as $\textbf{p}^{\mathcal{O}_{i}} _{ij}(t) = \textbf{p}^{\mathcal{O}_{i}} _{i}(t) - \textbf{p}^{\mathcal{O}_{i}} _{j}(t)$.
	With the UWB module, each robot $i$ can measure the real-time distance $d_{ij}(t)\triangleq\lVert \textbf{p}^{\mathcal{O}_{i}} _{ij}(t) \rVert$ from itself to its neighbor $j$ and communicate with each other.
	The interaction network among the robots is described by a directed graph $\mathcal{G}=(\mathcal{V},\mathcal{E})$, where $\mathcal{V}=\{0,1,\ldots,N\}$ and  $\mathcal{E}\subseteq \mathcal{V}\times\mathcal{V}$.
	%which is composed of a vertex set $\mathcal{V}=\{0,1,\ldots,N\}$ and an edge set $\mathcal{E}\subseteq \mathcal{V}\times\mathcal{V}$.
	Robot $0$ is the leader robot.
	If $(i,j)\in \mathcal{E}$, robot $i$ and $j$ are neighbors, neighbors can communicate and measure distance with each other. 
	The neighbor set of robot $i$ is denoted as $\mathcal{N}_i=\{j\in \mathcal{V}: (i,j)\in \mathcal{E}\}$.
	
	\subsection{Control Objective}
	As shown in Fig. \ref{rl_figure}, under the condition that there are no global coordinate system and a shared orientation, the initial position and orientation of each robot $i$ is random. 
	For each robot $i$, only the distance $d_{ij}$ to its neighbor $j$, odometer displacements $\textbf{p}^{\mathcal{O}_{i}} _{i}(t)$ and $\phi_{\Sigma_{i}}^{\mathcal{O}_{i}}(t)$ are measurable.
	Define the real-time relative orientation between robot $j$ and robot $i$ as $\theta_{\Sigma_{i}}^{\Sigma_{0}}(t)$. 
	The control objective can be formulated as
	\begin{align}
		\label{control objective}
		\left\{
		\begin{aligned}
			&\lim_{t \to \infty} \theta_{\Sigma_{i}}^{\Sigma_{0}}(t) = 0 \\
			&\lim_{t \to \infty} \textbf{p}^{\mathcal{O}_{0}}_{i0}(t) - \textbf{p}_{i}^{*} = 0,
		\end{aligned}
		\right.
	\end{align}
	where $\textbf{p}_{i}^{*} \triangleq [p_{ix}^{*}, p_{iy}^{*}, p_{iz}^{*}]^T$ denotes the desired relative position for robot $i$ to the leader. Eq. (\ref{control objective}) indicates that each robot $i$ has the same orientation and maintains a desired relative position $\textbf{p}_{i}^{*}$ with the leader in the leader's odometer frame $\mathcal{O}_{0}$. 
	To achieve the control task, the following lemma is required,
	\begin{lemma} \cite{Zhao2023TM}
		\label{a s}
		Consider a perturbed system
		\begin{align}
			\label{system}
			\dot{\textbf{x}} = f(t,\textbf{x}) + g(t,\textbf{x},\bm{\delta}).
		\end{align}
		where $\textbf{x} \in \mathbb{R}^{n_{x}}$ is the state and $\delta \in \mathbb{R}^{n_{\delta}}$ is an exogenous signal. $f(t,\textbf{x})$ and $g(t,\textbf{x},\bm{\delta})$ are piecewise continues in $t$ and locally Lipschitz in $\textbf{x}$ and $(\textbf{x},\bm{\delta})$, respectively. Normal system $\dot{\textbf{x}} = f(t,\textbf{x})$ in (\ref{system}) is uniformly exponentially stable and $g(t,\textbf{x},\bm{\delta})$ satisfies $\lVert g(t,\textbf{x},\bm{\delta}) \rVert \leq a_{1}\lVert \bm{\delta} \rVert \lVert \textbf{x} \rVert + a_{2} \lVert \bm{\delta} \rVert$,
		with two positive constants $a_{1}$ and $a_{2}$. Then system (\ref{system}) uniformly asymptotically stable when $\lVert \bm{\delta} \rVert$ converges to $\textbf{0}$ asymptotically.
	\end{lemma} 
	
	It is worth noting that absolute localization systems, such as GPS or external global anchor points, are not necessary for accomplishing the formation control task.
	The leader robot provides a common reference frame for the swarm using its odometer frame.
	Other robots achieve formation tasks by collaboratively estimating their relative pose to the leader.
	If the current leader is damaged, the swarm can simply select a new leader, and the other robots can continue to perform cooperative localization to the new leader using our method.
	
	%\begin{figure}[!t]\centering
	%	\includegraphics[scale=1.066]{figures/overview.jpg}
	%	\caption{An illustration of the proposed formation control scheme.}\centering
	%	\label{overview}
	%	\vspace{-0.5cm}
	%\end{figure}
	
	\section{Methods}
	
	To achieve the control task, we design a strategy for each robot $i$ including neighbor relative localization, leader cooperative localization and local formation tracking controller. 
	%The overall system architecture is illustrated in Fig. \ref{overview}.
	Firstly, we note that the real-time relative pose estimation problem between robot $i$ and robot $j$ can be formulated as the initial relative pose $\textbf{T}^{\mathcal{O}_{i}} _{\mathcal{O}_{j}}$ estimation problem (see Section \ref{rl model}).
	The second step involves designing concurrent-learning based estimators $\hat{\textbf{p}}^{\mathcal{O}_{i}}_{ij,0}$ and $\hat{\textbf{R}}^{\mathcal{O}_{i}} _{\mathcal{O}_{j}}$ to estimate the initial relative pose between neighboring robots  (see Section \ref{rl neighbor}).
	However, not all the robots can estimate its relative pose to the leader robot directly. 
	To handle this issue, the third step is to design the estimators $\hat{\textbf{q}}^{\mathcal{O}_{i}}_{i0,0}$ and $\hat{\textbf{Q}}^{\mathcal{O}_{i}} _{\mathcal{O}_{0}}$ to estimate the initial relative pose to the leader robot (see Section \ref{rl leader}).
	Finally, based on the relative pose estimators and odometer information. distributed formation controller $v_{i}$ and $w_{i}$ are devised (see Section \ref{controller}).
	
	\subsection{Problem Formulation of Relative Localization}
	\label{rl model}
	
	First, we show that relative localization between robot $i$ and its neighbor $j \in \mathcal{N}_{i}$ is equal to estimate the following relative frame transformation in robot $i$'s odometry frame $\mathcal{O}_{i}$:
	\renewcommand{\arraystretch}{1.5}
	\begin{align}
		\notag
		\textbf{T}^{\mathcal{O}_{i}} _{\mathcal{O}_{j}} := \begin{bmatrix}
			\textbf{R}^{\mathcal{O}_{i}} _{\mathcal{O}_{j}}, & \textbf{p}^{\mathcal{O}_{i}}_{ij}(t_{0}) \\
			\textbf{0}^{T}, & 1
		\end{bmatrix},
	\end{align}
	where $\textbf{p}^{\mathcal{O}_{i}}_{ij}(t_{0})$ is the initial relative position in frame $\mathcal{O}_{i}$. 
	Define $\theta_{\mathcal{O}_{j}}^{\mathcal{O}_{i}}(t_{0})$ as the initial relative orientation between frame $\mathcal{O}_{j}$ and frame $\mathcal{O}_{i}$ at $t_{0}$.
	The rotation matrix from frame $\mathcal{O}_{j}$ to frame $\mathcal{O}_{i}$ can be defined as
	\begin{align}
		\notag
		\textbf{R}^{\mathcal{O}_{i}} _{\mathcal{O}_{j}} :\triangleq \begin{bmatrix}
			\textbf{R}^{\mathcal{O}_{i,h}} _{\mathcal{O}_{j,h}}, & \textbf{0} \\
			\textbf{0}, & \textbf{1}
		\end{bmatrix} =  \begin{bmatrix}
			\cos \theta_{\mathcal{O}_{j}}^{\mathcal{O}_{i}}(t_{0}),&-\sin \theta_{\mathcal{O}_{j}}^{\mathcal{O}_{i}}(t_{0}),&0 \\
			\sin \theta_{\mathcal{O}_{j}}^{\mathcal{O}_{i}}(t_{0}),&\cos \theta_{\mathcal{O}_{j}}^{\mathcal{O}_{i}}(t_{0}),&0 \\
			0,&0,&1
		\end{bmatrix}.
	\end{align}
	
	Define the relative orientation between frame $\Sigma_{j}$ in frame $\Sigma_{i}$ at time $t$ as $\theta_{\Sigma_{j}}^{\Sigma_{i}}(t)$. With the IO measurements, the following equations holds,
	\begin{subequations}
		\label{initial}
		\begin{align}
			\textbf{p}^{\mathcal{O}_{i}}_{ij}(t) &= \textbf{p}^{\mathcal{O}_{i}}_{ij}(t_{0}) + \textbf{p}^{\mathcal{O}_{i}}_{i}(t) - \textbf{R}^{\mathcal{O}_{i}} _{\mathcal{O}_{j}}\textbf{p}^{\mathcal{O}_{j}}_{j}(t) \\
			\theta_{\Sigma_{j}}^{\Sigma_{i}}(t) &= \theta_{\mathcal{O}_{j}}^{\mathcal{O}_{i}}(t_{0}) + \phi_{\Sigma_{i}}^{\mathcal{O}_{i}}(t) - \phi_{\Sigma_{j}}^{\mathcal{O}_{j}}(t),
		\end{align}
	\end{subequations}
	
	One can see that with IO measurements, $\textbf{p}^{\mathcal{O}_{i}}_{i}(t)\triangleq [\textbf{p}^{\mathcal{O}_{i}}_{i,h}(t), p^{\mathcal{O}_{i}}_{i,z}(t)]^{t} \triangleq [p^{\mathcal{O}_{i}}_{i,x}(t), p^{\mathcal{O}_{i}}_{i,y}(t), p^{\mathcal{O}_{i}}_{i,z}(t)]^{T}$ and $\phi_{\Sigma_{i}}^{\mathcal{O}_{i}}(t)$ are available. Furthermore, the neighbor's IO measurements $\textbf{p}^{\mathcal{O}_{j}}_{j}(t)$ and $\phi_{\Sigma_{j}}^{\mathcal{O}_{j}}(t)$ can also be obtainable through UWB module.
	As a result, the relative localization problem between neighboring robots is equal to estimate the relative frame transformation $\textbf{T}^{\mathcal{O}_{i}} _{\mathcal{O}_{j}}$ between odometry frames $\mathcal{O}_{i}$ and $\mathcal{O}_{j}$.
	
	At sampling time $t_{k} \triangleq k\Delta t, k\in \mathbb{N}$, the distance between robot $i$ and robot $j$ can be measured with the UWB module, as shown in Fig. \ref{rl_figure} B. According to the cosine theorem, the measurements obtained from two adjacent samples satisfy
	\begin{align}
		\label{cosine}
		\textbf{u}^{\mathcal{O}_{i}}_{ij}(t_{k})^{T} \textbf{p}^{\mathcal{O}_{i}}_{ij}(t_{k}) = &\frac{1}{2}\Big( d_{ij}(t_{k+1})^2 + d_{ij}(t_{k})^2 - \\
		\notag
		&\textbf{u}^{\mathcal{O}_{i}}_{ij}(t_{k})^{T} \textbf{u}^{\mathcal{O}_{i}}_{ij}(t_{k}) \Big),
	\end{align}
	where $\textbf{u}^{\mathcal{O}_{i}}_{ij}(t_{k})$ is the relative displacement in frame $\mathcal{O}_{i}$ from time $t_{k}$ to $t_{k+1}$ and it is defined as
	\begin{align}
		\label{relative_input}
		\textbf{u}^{\mathcal{O}_{i}}_{ij}(t_{k}) = \textbf{u}^{\mathcal{O}_{i}}_{i}(t_{k}) - \textbf{R}^{\mathcal{O}_{i}} _{\mathcal{O}_{j}}\textbf{u}^{\mathcal{O}_{j}}_{j}(t_{k}),
	\end{align} 
	where $\textbf{u}^{\mathcal{O}_{i}}_{i}(t_{k}) = \textbf{p}^{\mathcal{O}_{i}}_{i}(t_{k+1}) - \textbf{p}^{\mathcal{O}_{i}}_{i}(t_{k}) \triangleq [\textbf{u}^{\mathcal{O}_{i}}_{i,h}(t_{k}), u^{\mathcal{O}_{i}}_{i,z}(t_{k})]^{T} \triangleq [u^{\mathcal{O}_{i}}_{i,x}(t_{k}), u^{\mathcal{O}_{i}}_{i,y}(t_{k}), u^{\mathcal{O}_{i}}_{i,z}(t_{k})]^{T}$ is the displacement of robot $i$ in frame $\mathcal{O}_{i}$ from $t_{k}$ to $t_{k+1}$, similar to $\textbf{u}^{\mathcal{O}_{j}}_{j}(t_{k})$.
	Substituting (\ref{initial}), (\ref{relative_input}) into (\ref{cosine}), yields
	\begin{align}
		\label{full_equ}
		\notag
		&\textbf{u}^{\mathcal{O}_{i}}_{i}(t_{k})^{T} \textbf{p}^{\mathcal{O}_{i}}_{ij}(t_{0})
		+
		\textbf{u}^{\mathcal{O}_{i}}_{i}(t_{k})^{T}\textbf{p}^{\mathcal{O}_{i}}_{i}(t_{k})
		- \textbf{u}^{\mathcal{O}_{i}}_{i}(t_{k})^{T}\textbf{R}^{\mathcal{O}_{i}}_{\mathcal{O}_{j}}\textbf{p}^{\mathcal{O}_{j}}_{j}(t_{k})
		\\
		&-\textbf{p}^{\mathcal{O}_{i}}_{ij}(t_{0})^{T}\textbf{R}^{\mathcal{O}_{i}}_{\mathcal{O}_{j}} \textbf{u}^{\mathcal{O}_{j}}_{j}(t_{k}) 
		-\textbf{p}^{\mathcal{O}_{i}}_{i}(t_{k})^{T}\textbf{R}^{\mathcal{O}_{i}}_{\mathcal{O}_{j}}\textbf{u}^{\mathcal{O}_{j}}_{j}(t_{k})
		\\
		\notag
		&+\textbf{u}^{\mathcal{O}_{j}}_{j}(t_{k})^{T} \textbf{p}^{\mathcal{O}_{j}}_{j}(t_{k}) = \frac{1}{2}\Big( d_{ij}(t_{k})^2 - \textbf{u}^{\mathcal{O}_{i}}_{i}(t_{k})^{T} \textbf{u}^{\mathcal{O}_{i}}_{i}(t_{k}) 
		\\
		\notag
		&- \textbf{u}^{\mathcal{O}_{j}}_{j}(t_{k})^{T} \textbf{u}^{\mathcal{O}_{j}}_{j}(t_{k}) 
		+ d_{ij}(t_{k+1})^2\Big) 
		+ \textbf{u}^{\mathcal{O}_{i}}_{i}(t_{k})^{T} \textbf{R}^{\mathcal{O}_{i}} \textbf{u}^{\mathcal{O}_{j}}_{j}(t_{k}).
	\end{align}
	Second, the obtainable terms and unknown parameters in (\ref{full_equ}) are organized, and one can see,
	\begin{align}
		\label{simple}
		\notag
		\bar{y}_{ij}(t_{k}) = &\textbf{u}^{\mathcal{O}_{i}}_{i}(t_{k})^{T} \textbf{p}^{\mathcal{O}_{i}}_{ij}(t_{0})
		-\textbf{u}^{\mathcal{O}_{j}}_{j}(t_{k})^{T} \left( \textbf{R}^{\mathcal{O}_{i}}_{\mathcal{O}_{j}} \right)^{T}
		\textbf{p}^{\mathcal{O}_{i}}_{ij}(t_{0})
		\\
		\notag
		&-\textbf{u}^{\mathcal{O}_{i}}_{i,h}(t_{k})^{T}\textbf{R}^{\mathcal{O}_{i,h}} _{\mathcal{O}_{j,h}}\textbf{p}^{\mathcal{O}_{j}}_{j,h}(t_{k})
		-\textbf{p}^{\mathcal{O}_{i}}_{i,h}(t_{k})^{T}\textbf{R}^{\mathcal{O}_{i,h}} _{\mathcal{O}_{j,h}}\textbf{u}^{\mathcal{O}_{j}}_{j,h}(t_{k})
		\\
		&-\textbf{u}^{\mathcal{O}_{i}}_{i,h}(t_{k})^{T} \textbf{R}^{\mathcal{O}_{i,h}} _{\mathcal{O}_{j,h}}
		\textbf{u}^{\mathcal{O}_{j}}_{j,h}(t_{k}),
	\end{align}
	where $\bar{y}_{ij}(t_{k})$ is defined as
	\begin{align}
		\notag
		&\bar{y}_{ij}(t_{k}) \triangleq \frac{1}{2}\Big( d_{ij}(t_{k+1})^2 
		+ d_{ij}(t_{k})^2 - \textbf{u}^{\mathcal{O}_{i}}_{i}(t_{k})^{T} \textbf{u}^{\mathcal{O}_{i}}_{i}(t_{k}) 
		\\
		\notag
		&- \textbf{u}^{\mathcal{O}_{j}}_{j}(t_{k})^{T} \textbf{u}^{\mathcal{O}_{j}}_{j}(t_{k}) \Big) - \textbf{u}^{\mathcal{O}_{i}}_{i}(t_{k})^{T}\textbf{p}^{\mathcal{O}_{i}}_{i}(t_{k}) - \textbf{u}^{\mathcal{O}_{j}}_{j}(t_{k})^{T} \textbf{p}^{\mathcal{O}_{j}}_{j}(t_{k})
		\\
		\notag
		&+u^{\mathcal{O}_{i}}_{i,z}(t_{k}) p^{\mathcal{O}_{j}}_{j,z}(t_{k}) + p^{\mathcal{O}_{i}}_{i,z}(t_{k}) p^{\mathcal{O}_{j}}_{j,z}(t_{k}) + u^{\mathcal{O}_{i}}_{i,z}(t_{k}) u^{\mathcal{O}_{j}}_{j,z}(t_{k}).
	\end{align} 
	
	Here, we point out that $\bar{y}_{ij}(t_{k})$ is obtainable at time $t_{k+1}$. Expand the third item in the right hand of Eq. (\ref{simple}) and it has,
	%\begin{align}
	%	\label{equ1}
	%	&\textbf{u}^{\mathcal{O}_{i}}_{i,h}(t_{k})^{T}\textbf{R}^{\mathcal{O}_{i,h}}_{\mathcal{O}_{j,h}}\textbf{p}^{\mathcal{O}_{j}}_{j,h}(t_{k})
	%	\\
	%	\notag
	%	=&\begin{bmatrix}
		%		u^{\mathcal{O}_{i}}_{i,x}(t_{k}), & u^{\mathcal{O}_{i}}_{i,y}(t_{k}) 
		%	\end{bmatrix}  
	%	\begin{bmatrix}
		%		\cos \theta_{\mathcal{O}_{j}}^{\mathcal{O}_{i}}(t_{0}), & -\sin \theta_{\mathcal{O}_{j}}^{\mathcal{O}_{i}}(t_{0}) \\
		%		\sin \theta_{\mathcal{O}_{j}}^{\mathcal{O}_{i}}(t_{0}), & \cos \theta_{\mathcal{O}_{j}}^{\mathcal{O}_{i}}(t_{0})
		%	\end{bmatrix}  
	%	\begin{bmatrix}
		%		p^{\mathcal{O}_{j}}_{j,x}(t_{k}) \\ p^{\mathcal{O}_{j}}_{j,y}(t_{k}) 
		%	\end{bmatrix}
	%	\\
	%	\notag
	%	=& \big( u^{\mathcal{O}_{i}}_{i,x}(t_{k}) p^{\mathcal{O}_{j}}_{j,x}(t_{k}) + u^{\mathcal{O}_{i}}_{i,y}(t_{k}) p^{\mathcal{O}_{j}}_{j,y}(t_{k}) \big)\cos \theta_{\mathcal{O}_{j}}^{\mathcal{O}_{i}}(t_{0})
	%	\\
	%	\notag
	%	&+ \big( u^{\mathcal{O}_{i}}_{i,y}(t_{k}) p^{\mathcal{O}_{j}}_{j,x}(t_{k}) - u^{\mathcal{O}_{i}}_{i,x}(t_{k}) p^{\mathcal{O}_{j}}_{j,y}(t_{k}) \big)\sin \theta_{\mathcal{O}_{j}}^{\mathcal{O}_{i}}(t_{0}) 
	%	\\
	%	\notag
	%	=&\begin{bmatrix}
		%		\textbf{u}^{\mathcal{O}_{i}}_{i,h}(t_{k})^{T} \textbf{p}^{\mathcal{O}_{j}}_{j,h}(t_{k}), & \textbf{u}^{\mathcal{O}_{i}}_{i,h}(t_{k}) \times \textbf{p}^{\mathcal{O}_{j}}_{j,h}(t_{k})
		%	\end{bmatrix}  
	%	\begin{bmatrix}
		%		\cos \theta_{\mathcal{O}_{j}}^{\mathcal{O}_{i}}(t_{0}) \\ \sin \theta_{\mathcal{O}_{j}}^{\mathcal{O}_{i}}(t_{0})
		%	\end{bmatrix}.
	%\end{align}
	One can check the following equations holds:
	\begin{align}
		\label{equ2}
		&\textbf{u}^{\mathcal{O}_{i}}_{i,h}(t_{k})^{T}\textbf{R}^{\mathcal{O}_{i,h}}_{\mathcal{O}_{j,h}}\textbf{p}^{\mathcal{O}_{j}}_{j,h}(t_{k})
		\\
		\notag
		=&\begin{bmatrix}
			\textbf{u}^{\mathcal{O}_{i}}_{i,h}(t_{k})^{T} \textbf{p}^{\mathcal{O}_{j}}_{j,h}(t_{k}), &   \textbf{p}^{\mathcal{O}_{j}}_{j,h}(t_{k}) \times \textbf{u}^{\mathcal{O}_{i}}_{i,h}(t_{k})
		\end{bmatrix}  
		\begin{bmatrix}
			\cos \theta_{\mathcal{O}_{j}}^{\mathcal{O}_{i}}(t_{0}) \\ \sin \theta_{\mathcal{O}_{j}}^{\mathcal{O}_{i}}(t_{0})
		\end{bmatrix}
		\\
		\notag
		&\textbf{p}^{\mathcal{O}_{i}}_{i,h}(t_{k})^{T}\textbf{R}^{\mathcal{O}_{i,h}} _{\mathcal{O}_{j,h}}
		\textbf{u}^{\mathcal{O}_{j}}_{j,h}(t_{k}) \\
		\notag
		= &\begin{bmatrix}
			\textbf{p}^{\mathcal{O}_{i}}_{i,h}(t_{k})^{T} \textbf{u}^{\mathcal{O}_{j}}_{j,h}(t_{k}), &   \textbf{u}^{\mathcal{O}_{j}}_{j,h}(t_{k}) \times \textbf{p}^{\mathcal{O}_{i}}_{i,h}(t_{k})
		\end{bmatrix}  
		\begin{bmatrix}
			\cos \theta_{\mathcal{O}_{j}}^{\mathcal{O}_{i}}(t_{0}) \\ \sin \theta_{\mathcal{O}_{j}}^{\mathcal{O}_{i}}(t_{0})
		\end{bmatrix}.
		\\
		\notag
		&\textbf{u}^{\mathcal{O}_{i}}_{i,h}(t_{k})^{T} \textbf{R}^{\mathcal{O}_{i,h}} _{\mathcal{O}_{j,h}}
		\textbf{u}^{\mathcal{O}_{j}}_{j,h}(t_{k}) \\
		\notag
		= &\begin{bmatrix}
			\textbf{u}^{\mathcal{O}_{i}}_{i,h}(t_{k})^{T} \textbf{u}^{\mathcal{O}_{j}}_{j,h}(t_{k}), &  \textbf{u}^{\mathcal{O}_{j}}_{j,h}(t_{k}) \times\textbf{u}^{\mathcal{O}_{i}}_{i,h}(t_{k}) 
		\end{bmatrix}  
		\begin{bmatrix}
			\cos \theta_{\mathcal{O}_{j}}^{\mathcal{O}_{i}}(t_{0}) \\ \sin \theta_{\mathcal{O}_{j}}^{\mathcal{O}_{i}}(t_{0})
		\end{bmatrix}.
	\end{align}
	According to Eq. (\ref{equ2}), define $\mathbf{\Psi}_{ij}$ and $\mathbf{\Theta}_{ij}$ as
	\begin{align}
		\label{Phi_Theta}
		\mathbf{\Psi}_{ij}(t_{k}) = [&\textbf{u}^{\mathcal{O}_{i}}_{i,h}(t_{k})^{T},  u^{\mathcal{O}_{i}}_{i,z}(t_{k}) - u^{\mathcal{O}_{j}}_{j,z}(t_{k}),  -\textbf{u}^{\mathcal{O}_{j}}_{j,h}(t_{k})^{T}, 
		\\
		\notag
		&-a_{ij}(t_{k}), -b_{ij}(t_{k})]^{T}
		\\
		\notag
		\mathbf{\Theta}_{ij} = [&\textbf{p}^{\mathcal{O}_{i}}_{ij}(t_{0})^{T}, (\textbf{R}^{\mathcal{O}_{i}}_{\mathcal{O}_{j}})^{T} (\textbf{p}^{\mathcal{O}_{i}}_{i,h}(t_{0}) - \textbf{p}^{\mathcal{O}_{j}}_{j,h}(t_{0})) ,
		\\
		\notag
		& \cos \theta_{\mathcal{O}_{j}}^{\mathcal{O}_{i}}(t_{0}), \sin \theta_{\mathcal{O}_{j}}^{\mathcal{O}_{i}}(t_{0})
		]^{T}.
	\end{align}
	where $a_{ij}(t_{k})$ and $b_{ij}(t_{k})$ are defined as
	\begin{subequations}
		\begin{align}
			\notag
			a_{ij}(t_{k}) \triangleq & \textbf{u}^{\mathcal{O}_{i}}_{i,h}(t_{k})^{T} \textbf{p}^{\mathcal{O}_{j}}_{j,h}(t_{k}) + \textbf{p}^{\mathcal{O}_{i}}_{i,h}(t_{k})^{T} \textbf{u}^{\mathcal{O}_{j}}_{j,h}(t_{k}) \\
			\notag
			&+ \textbf{u}^{\mathcal{O}_{i}}_{i,h}(t_{k})^{T} \textbf{u}^{\mathcal{O}_{j}}_{j,h}(t_{k}) \\
			\notag
			b_{ij}(t_{k}) \triangleq &
			\textbf{p}^{\mathcal{O}_{j}}_{j,h}(t_{k}) \times \textbf{u}^{\mathcal{O}_{i}}_{i,h}(t_{k})+   \textbf{u}^{\mathcal{O}_{j}}_{j,h}(t_{k}) \times \textbf{p}^{\mathcal{O}_{i}}_{i,h}(t_{k}) \\
			\notag
			&+ \textbf{u}^{\mathcal{O}_{j}}_{j,h}(t_{k}) \times \textbf{u}^{\mathcal{O}_{i}}_{i,h}(t_{k}) .
		\end{align}
	\end{subequations}
	Note that both $a_{ij}(t_{k})$ and $b_{ij}(t_{k})$ are available at time $t_{k}$. According to Eq. (\ref{Phi_Theta}), (\ref{simple}) can be simplified as:
	\begin{align}
		\notag
		y_{ij}(t_{k}) \triangleq \frac{\bar{y}_{ij}(t_{k})}{\lVert \mathbf{\Psi}_{ij}(t_{k}) \rVert} = \mathbf{\Theta}_{ij}^{T} \frac{\mathbf{\Psi}_{ij}(t_{k})}{\lVert \mathbf{\Psi}_{ij}(t_{k}) \rVert} \triangleq \mathbf{\Theta}_{ij}^{T}\mathbf{\Phi}_{ij}(t_{k}).
	\end{align}
	Denote $\hat{\mathbf{\Theta}}_{ij}$ as an estimation of $\mathbf{\Theta}_{ij}$ and $\tilde{\mathbf{\Theta}}_{ij} = \hat{\mathbf{\Theta}}_{ij} - \mathbf{\Theta}_{ij}$ as the estimation error. The innovation $\epsilon_{ij}(t_{k})$ is defined as
	\begin{align}
		\label{innovation}
		\epsilon_{ij}(t_{k}) = \hat{\mathbf{\Theta}}_{ij}^{T}(t_{k})\mathbf{\Phi}_{ij}(t_{k}) - y_{ij}(t_{k}) = \tilde{\mathbf{\Theta}}_{ij}^{T}(t_{k})\mathbf{\Phi}_{ij}(t_{k}).
	\end{align}
	The innovation $\epsilon_{ij}(t_{k})$ will be used later to design the concurrent-learning based estimator in Section \ref{rl neighbor}.
	
	\subsection{Relative Localization to Neighboring Robot}
	\label{rl neighbor}
	Based on the innovation $\epsilon_{ij}(t_{k})$ defined in (\ref{innovation}), now we can design the concurrent-learning based relative pose estimator to update $\hat{\mathbf{\Theta}}_{ij}(t)$ as
	\begin{align}
		\label{relative estimator}
		\left\{
		\begin{aligned}
			&\hat{\mathbf{\Theta}}_{ij}(t) = \hat{\mathbf{\Theta}}_{ij}(t_{k}), \quad t_{k} < t \leq t_{k+1} \\
			&\hat{\mathbf{\Theta}}_{ij}(t_{k+1}) = \hat{\mathbf{\Theta}}_{ij}(t_{k}) - \eta \sum_{m=1}^{\varsigma} \mathbf{\Phi}_{ij}(t_{m}) \epsilon_{ij}(t_{m}) \\
			&- \eta \mathbf{\Phi}_{ij}(t_{k}) \epsilon_{ij}(t_{k}). 
		\end{aligned}
		\right.
	\end{align}
	
	with the learning rate $\eta=\frac{\lambda_{min}(\textbf{S}_{ij})}{(\lambda_{max}(\textbf{U}_{ij}(t_{k})) + \lambda_{max}(\textbf{S}_{ij})^2)}$ 
	%\begin{align}
	%	\label{learning_rate}
	%	\eta = \frac{\lambda_{min}(\textbf{S}_{ij})}{(\lambda_{max}(\textbf{U}_{ij}(t_{k})) + \lambda_{max}(\textbf{S}_{ij})^2)},
	%\end{align}
	where matrices $\textbf{U}_{ij}(t_{k})$ and $\textbf{S}_{ij}$ are defined as $\textbf{U}_{ij}(t_{k}) = \mathbf{\Phi}_{ij}(t_{k}) \mathbf{\Phi}_{ij}(t_{k})^{T}\in \mathbb{R}^{7 \times 7}$ and $\textbf{S}_{ij}(t_{k}) = \mathbf{R}_{ij} \mathbf{R}_{ij}^{T} \in \mathbb{R}^{7 \times 7}$, respectively. Here, $\mathbf{R}_{ij} = [ \mathbf{\Phi}_{ij}(t_{1}),...,\mathbf{\Phi}_{ij}(t_{\varsigma})]$.
	For brief, define $\hat{\textbf{p}}^{\mathcal{O}_{i}}_{ij,0} \triangleq \hat{\mathbf{\Theta}}_{ij}[1:3]$ as the estimate of the initial relative position $\textbf{p}^{\mathcal{O}_{i}}_{ij}(t_{0})$ in frame $\mathcal{O}_{i}$. 
	Similarly, the sine $\sin \theta_{\mathcal{O}_{j}}^{\mathcal{O}_{i}}(t_{0})$ and cosine $\cos \theta_{\mathcal{O}_{j}}^{\mathcal{O}_{i}}(t_{0})$ value of the initial relative frame angle can be estimated by $\hat{\mathbf{\Theta}}_{ij}[6]$ and $\hat{\mathbf{\Theta}}_{ij}[7]$.
	Based on estimator (\ref{relative estimator}), the initial rotation matrix $\textbf{R}^{\mathcal{O}_{i}} _{\mathcal{O}_{j}}(t)$ between robot $i$ and robot $j$ can be estimated by 
	%\begin{align}
	%	\hat{\textbf{p}}^{\mathcal{O}_{i}}_{ij}(t) \triangleq \hat{\textbf{p}}^{\mathcal{O}_{i}}_{ij,0}(t) + \textbf{z}^{\mathcal{O}_{i}}_{i}(t) - \hat{\textbf{R}}^{\mathcal{O}_{i}} _{\mathcal{O}_{j}}(t)\textbf{z}^{\mathcal{O}_{j}}_{j}(t)
	%\end{align}
	\begin{align}
		\label{initial inter-robot R}
		\hat{\textbf{R}}^{\mathcal{O}_{i}} _{\mathcal{O}_{j}}(t) := \mathrm{Norm} \left( \begin{bmatrix}
			\hat{\mathbf{\Theta}}_{ij}[6], & -\hat{\mathbf{\Theta}}_{ij}[7], &0 \\
			\hat{\mathbf{\Theta}}_{ij}[7], & \hat{\mathbf{\Theta}}_{ij}[6], &0 \\
			0, &0, &1
		\end{bmatrix} \right),
	\end{align}
	where $R=\mathrm{Norm}(A)$ is a function that normalizes an antisymmetric matrix $A$ into a rotation matrix $R$. 
	
	It is worth mentioning that for 2D robots, there is no need to estimate the relative position in $z$-axis, the variables in $z$ dimension in Eq. (\ref{Phi_Theta}) can be set to zero. The estimator for the two-dimensional situation can be easily obtained based on Eq. (\ref{relative estimator}), and we omit it for brief.
	In the next theorem, we will show that the exponential convergence of the proposed estimator (\ref{relative estimator}) can be guaranteed if the recorded data matrix $S_{ij}$ satisfies the following assumption.
	\begin{assumption}
		\label{record data}
		The recorded data matrix $S_{ij}$ is full rank.
	\end{assumption}
	
	Assumption \ref{record data} indicates that the data matrix $S_{ij}$ records sufficient information, which is also known as SE condition. 
	Compared to the PE condition, SE condition is much weaker and can also be found in many data-driven estimation results like \cite{Chowdhary2010CDC} and \cite{Djaneye2019TAC}. 
	We also analyze the observability of relative pose when the relative motion between neighboring robots is different, which can be found in Appendix \ref{Theoretical analysis}-D.
	The main results regarding the convergence of the estimator (\ref{relative estimator}) are presented in the following theorem. 
	
	\begin{theorem}
		\label{relative localization}
		Under Assumption \ref{record data}, estimation error $\tilde{\mathbf{\Theta}}_{ij}$ is globally exponentially stable with relative pose estimator (\ref{relative estimator}).
	\end{theorem}
	\begin{proof}
		See Appendix \ref{Theoretical analysis}-A.
	\end{proof}
	
	\begin{remark}
		Different from the latest result \cite{Zhao2023TM} in which the relative position should be available to estimate the unknown relative orientation, the estimator (\ref{relative estimator}) proposed in this article can achieve inter-robot relative pose with only UWB distance and IO displacement measurements.
		Theorem 1 indicates that when the historical measurement data records sufficient information, i.e. data matrix $S_{ij}$ has full rank, the localization error converges exponentially.
	\end{remark}
	
	\blue{In practice, both UWB and odometer measurements are subject to errors. After outlier rejection, these errors can be reasonably assumed to be bounded. According to Theorem 1 in \cite{muhlegg2012concurrent}, the resulting estimation error is bounded and linearly related to the measurement-error bound. Moreover, moderate odometry drift can be addressed by reinitializing the initial state and recollecting informative data when necessary.}
	
	\subsection{Relative Localization to Leader Robot}
	\label{rl leader}
\begin{figure}[!t]\centering
		\includegraphics[scale=1]{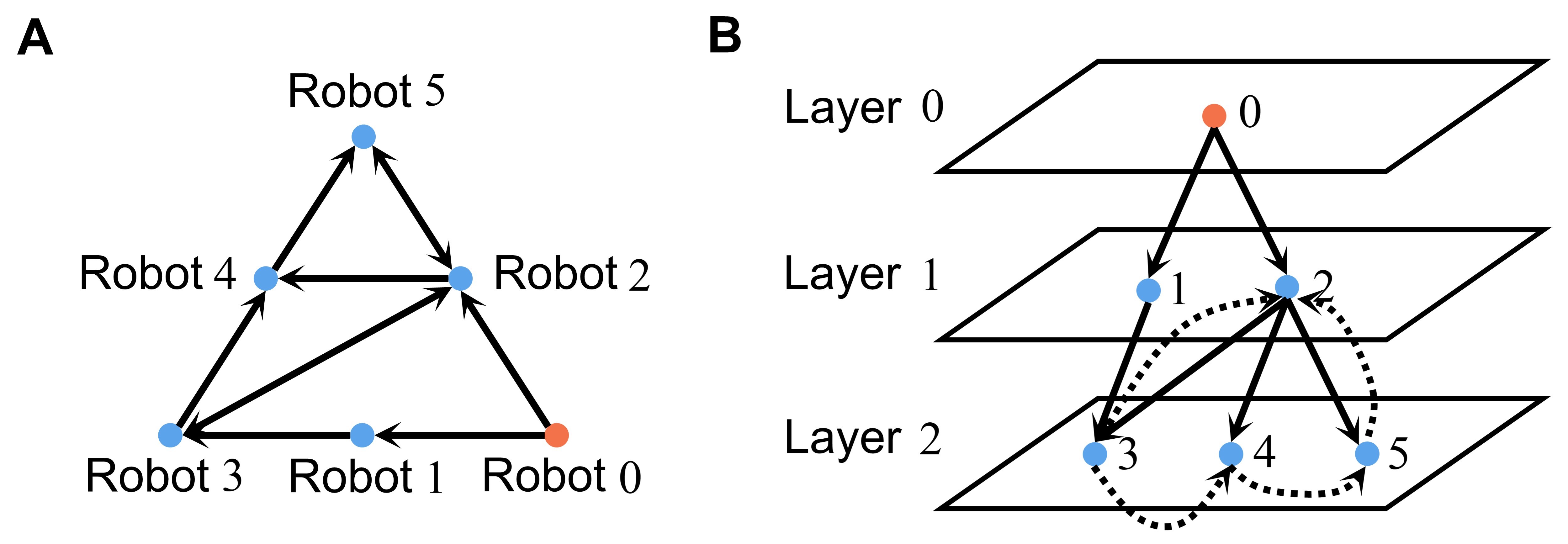}
		\caption{An example directed acyclic graph, the leader robot and follower robots are denoted in the orange and blue dots.}\centering
		\label{topology}
		\vspace{-0.5cm}
	\end{figure}

	To achieve the formation control, the relative pose between each robot $i$ and the leader robot are required. However, due to the directed communication topology, not all the robots has the ability to estimate the relative pose to the leader robot directly. 
	To deal with this issue, each robot need to utilize the information of its neighboring robots to estimate its own relative pose to the leader robot.
	To achieve this, we assume that the communication topology graph $\mathcal{G}$ is directed acyclic graph (DAG), such as the graph shown in Fig. \ref{topology} B. 
	This condition is mild and can be found in many formation control results, such as \cite{Xie2019TCNS} and \cite{Cao2023TC}. 
	For large-scale robot swarms, the hierarchical layering structure of swarm has been proven to be an effective approach to achieve cooperative task \cite{zhu2024self}.
	
	%\begin{algorithm}[!t]
	%	\small
	%	\caption{Neighbor robot selection}
	%	\KwIn{$\lambda_{0}$ $\leftarrow$ Data matrix threshold \; \qquad \quad $\delta_{0}$ $\leftarrow$ Threshold convergence rate \; \qquad \quad $n$ $\leftarrow$ Swarm size \;}
	%	\KwIn{Robot ID $i$, Robot $i$'s neighbor set $\bar{\mathcal{N}_{i}}$}
	%	\KwOut{Robot $i$'s valuable neighbor set $\mathcal{N}_{i}$}
	%	\BlankLine    
	%	$\mathcal{N}_{i}$ $\leftarrow$ set as empty set $\phi$ \;
	%	hop count $h_{i}$ $\leftarrow$ set as $0$ \;
	%	\While{ture}{
		%		\If{i == 0}{
			%			\textcolor{blue}{\scriptsize{\tcp*[h]{Hop count of leader robot is set as 0}\;}}
			%			continue
			%		}
		%		\Else{
			%			$h_{i}$ $\leftarrow$ $\min_{j \in \bar{\mathcal{N}}_{i}}\{ h_{j} \} + 1$ \;
			%			$\mathcal{N}_{i}$ $\leftarrow$ $\{ j | h_{j} = h_{i} -1\}$;
			%		}
		%	}
	%\end{algorithm}
	
	In our approach, we employ a DAG hierarchical topology.
	%to avoid robots using neighbor information that is too distant from the leader within the topology layer, as this can introduce additional localization errors.
	For instance, consider the measurement topology depicted in Fig. \ref{topology}A. 
	The measurement topology can be reorganized into the DAG form shown in Fig. \ref{topology}B through a hop based distributed neighbor selection algorithm\cite{cao2023similar}. 
	In the DAG graph, Robot 2 directly locates Leader 0 without relying on the estimation information of robot 3 and robot 5, thereby avoiding the introduction of additional localization errors.  
	\blue{Moreover, with the distributed neighbor selection algorithm\cite{cao2023similar}, the localization task can still be achieved even if some edges fail, i.e., due to communication lost or sensor damage, as long as the entire topology remains directionally connected.}
	
	The estimators $\hat{\textbf{q}}^{\mathcal{O}_{i}}_{i0}$ and $\hat{\textbf{Q}}^{\mathcal{O}_{i}}_{\mathcal{O}_{0}}$ are designed to estimate the initial relative position and frame rotation matrix between robot $i$ and the leader robot.
	Define the estimation error as $\tilde{\textbf{q}}^{\mathcal{O}_{i}}_{i0} = \hat{\textbf{q}}^{\mathcal{O}_{i}}_{i0} - \textbf{p}^{\mathcal{O}_{i}}_{i0}$ and $\tilde{\textbf{Q}}^{\mathcal{O}_{i}}_{\mathcal{O}_{0}} = \hat{\textbf{Q}}^{\mathcal{O}_{i}}_{\mathcal{O}_{0}} - \textbf{R}^{\mathcal{O}_{i}}_{\mathcal{O}_{0}}$.
	For robot $i$ in Layer $1$, i.e. $0 \in \mathcal{N}_{i}$, $\hat{\textbf{q}}^{\mathcal{O}_{i}}_{i0,0}$ and $\hat{\textbf{Q}}^{\mathcal{O}_{i}}_{\mathcal{O}_{0}}$ are designed as,
	\begin{align}
		\label{leader estimator initial1}
		\hat{\textbf{q}}^{\mathcal{O}_{i}}_{i0,0}(t) = \hat{\textbf{p}}^{\mathcal{O}_{i}}_{i0,0}(t), \quad
		\hat{\textbf{Q}}^{\mathcal{O}_{i}}_{\mathcal{O}_{0}}(t) = \hat{\textbf{R}}^{\mathcal{O}_{i}}_{\mathcal{O}_{0}}(t).
	\end{align}
	For robot $i$ in other Layers, i.e. $0 \notin \mathcal{N}_{i}$, $\hat{\textbf{q}}^{\mathcal{O}_{i}}_{i0,0}$ and $\hat{\textbf{Q}}^{\mathcal{O}_{i}}_{\mathcal{O}_{0}}$ are designed as,
	\begin{align}
		\label{leader estimator initial2}
		\hat{\textbf{q}}^{\mathcal{O}_{i}}_{i0,0}(t) &= \sum_{j \in \mathcal{N}_{i}} \frac{1}{|\mathcal{N}_{i}|} \left( \hat{\textbf{p}}^{\mathcal{O}_{i}}_{ij,0}(t) + \hat{\textbf{R}}^{\mathcal{O}_{i}} _{\mathcal{O}_{j}}(t) \hat{\textbf{q}}^{\mathcal{O}_{j}}_{j0,0}(t) \right) \\
		\notag
		\hat{\textbf{Q}}^{\mathcal{O}_{i}}_{\mathcal{O}_{0}}(t) &= \mathrm{Norm} \left( \sum_{j \in \mathcal{N}_{i}} \frac{1}{|\mathcal{N}_{i}|} \left( \hat{\textbf{R}}^{\mathcal{O}_{i}} _{\mathcal{O}_{j}}(t) \hat{\textbf{Q}}^{\mathcal{O}_{j}}_{\mathcal{O}_{0}}(t) \right) \right).
		\\
		\notag   
		&\triangleq \begin{bmatrix}
			\widehat{\cos \theta_{\mathcal{O}_{0}}^{\mathcal{O}_{i}}}(t), & -\widehat{\sin \theta_{\mathcal{O}_{0}}^{\mathcal{O}_{i}}}(t),&0 \\
			\widehat{\sin \theta_{\mathcal{O}_{0}}^{\mathcal{O}_{i}}}(t), & \widehat{\cos \theta_{\mathcal{O}_{0}}^{\mathcal{O}_{i}}}(t),&0 \\
			0, & 0, & 1
		\end{bmatrix}.
	\end{align}
	With estimators $\hat{\textbf{q}}^{\mathcal{O}_{i}}_{i0,0}(t)$ and $\hat{\textbf{Q}}^{\mathcal{O}_{i}}_{\mathcal{O}_{0}}(t)$ in (\ref{leader estimator initial1}) and (\ref{leader estimator initial2}), relative pose estimators $\hat{\textbf{q}}^{\mathcal{O}_{i}}_{i0}(t)$ and $\hat{\textbf{Q}}^{\Sigma_{i}}_{\Sigma_{0}}(t)$ are defined as
	\begin{align}
		\label{leader estimator}
		\hat{\textbf{q}}^{\mathcal{O}_{i}}_{i0}(t) \triangleq \hat{\textbf{q}}^{\mathcal{O}_{i}}_{i0,0}(t) + \textbf{z}^{\mathcal{O}_{i}}_{i}(t) - \hat{\textbf{Q}}^{\mathcal{O}_{i}} _{\mathcal{O}_{0}}\textbf{z}^{\mathcal{O}_{0}}_{0}(t) \\
		\notag
		\hat{\textbf{Q}}^{\Sigma_{0}}_{\Sigma_{i}}(t)\triangleq \begin{bmatrix}
			\widehat{\cos \theta_{\Sigma_{i}}^{\Sigma_{0}}}(t), & -\widehat{\sin \theta_{\Sigma_{i}}^{\Sigma_{0}}}(t), &0 \\
			\widehat{\sin \theta_{\Sigma_{i}}^{\Sigma_{0}}}(t), & \widehat{\cos \theta_{\Sigma_{i}}^{\Sigma_{0}}}(t), &0 \\
			0, &0, &1
		\end{bmatrix},
	\end{align}
	where $\widehat{\cos \theta_{\Sigma_{i}}^{\Sigma_{0}}}(t)$ and $\widehat{\sin \theta_{\Sigma_{i}}^{\Sigma_{0}}}(t)$ are defined as 
	\begin{subequations}
		\begin{align}
			\notag
			\widehat{\cos \theta_{\Sigma_{i}}^{\Sigma_{0}}}(t) 
			\triangleq& \widehat{\cos \theta_{\mathcal{O}_{0}}^{\mathcal{O}_{i}}}(t) \cos (\phi_{\Sigma_{i}}^{\mathcal{O}_{i}}(t)- \phi_{\Sigma_{0}}^{\mathcal{O}_{0}}(t)) \\
			\notag
			&+ \widehat{\sin \theta_{\mathcal{O}_{0}}^{\mathcal{O}_{i}}}(t) \sin (\phi_{\Sigma_{i}}^{\mathcal{O}_{i}(t)}- \phi_{\Sigma_{0}}^{\mathcal{O}_{0}}(t)) \\
			\notag
			\widehat{\sin \theta_{\Sigma_{i}}^{\Sigma_{0}}}(t) 
			\triangleq& \widehat{\cos \theta_{\mathcal{O}_{0}}^{\mathcal{O}_{i}}}(t) \sin (\phi_{\Sigma_{i}}^{\mathcal{O}_{i}(t)}- \phi_{\Sigma_{0}}^{\mathcal{O}_{0}}(t))
			\\
			\notag
			&- \widehat{\sin \theta_{\mathcal{O}_{0}}^{\mathcal{O}_{i}}}(t) \cos (\phi_{\Sigma_{i}}^{\mathcal{O}_{i}(t)}- \phi_{\Sigma_{0}}^{\mathcal{O}_{0}}(t)).
		\end{align}
	\end{subequations}
	
	Note that $\textbf{z}^{\mathcal{O}_{i}}_{i}(t)$ in (\ref{leader estimator}) can be obtained by integrating $v_{0,h}$, $v_{0,z}$ and $w_{0}$ with the following Assumption $\ref{leader vw}$. 
	
	\begin{assumption}
		\label{leader vw}
		The velocity $v_{0,h}$, $v_{0,v}$ and angle velocity $w_{0}$ of the leader robot are bounded and available to all followers. 
	\end{assumption}
	\blue{Assumption $\ref{leader vw}$ indicates that each robot is available to the leader's odometry information. In practical applications, that information can be achieved through hop count propagation \cite{Chen2024TCNS,Zhao2023TM}. 
		We provide the pseudocode of the hop count propagation algorithm and simulation results to prove that the communication delay is acceptable in Appendix \ref{Technical details}-B. 
		Besides, that information can also be estimated by designing distributed signal observers, as detailed in \cite{Ze2023TIE}.}
	The convergence of the estimator (\ref{leader estimator initial1}) and (\ref{leader estimator initial2}) are presented in the following theorem.
	\begin{theorem}
		\label{leader localization}
		Under Assumption \ref{leader vw}, given $(i,j) \in \mathcal{G}$, the record data matrix $S_{ij}$ satisfies Assumption \ref{record data}. 
		Then for robot $i$ in each layer, the cooperative localization errors $\tilde{\textbf{q}}^{\mathcal{O}_{i}}_{i0,0}$ and $\tilde{\textbf{Q}}^{\mathcal{O}_{i}}_{\mathcal{O}_{0}}$ converge asymptotically.
	\end{theorem}
	\begin{proof}
		See Appendix \ref{Theoretical analysis}-B.
	\end{proof}
	
	\subsection{Distributed Formation tracking Controller}
	\label{controller}
	The real-time relative orientation for robot $i$ in the leader's odometer frame $\mathcal{O}_{0}$ can be defined as 
	\begin{align}
		\label{theta_sigma_o}
		\theta_{\Sigma_{i}}^{\mathcal{O}_{0}}(t) =  \phi_{\Sigma_{i}}^{\mathcal{O}_{i}}(t) - \theta_{\mathcal{O}_{0}}^{\mathcal{O}_{i}}(t_{0}),
	\end{align} 
	where $\theta_{\mathcal{O}_{0}}^{\mathcal{O}_{i}}(t_{0})$ is the initial relative orientation to the leader robot. 
	%Furthermore the real-time relative orientation to the leader robot can be defined as
	%\begin{align}
	%	\theta_{\Sigma_{i}}^{\Sigma_{0}}(t) =  \phi_{\Sigma_{i}}^{\mathcal{O}_{i}}(t)- \phi_{\Sigma_{0}}^{\mathcal{O}_{0}}(t) - \theta_{\mathcal{O}_{0}}^{\mathcal{O}_{i}}(t_{0}). 
	%\end{align}
	Define the formation error $\textbf{e}_{i} = [\textbf{e}_{i,p}^T, e_{i,c}, e_{i,s}]^{T}$ as
	\begin{align}
		\label{tracking_error}
		\textbf{e}_{i,p} &=  \textbf{R}_{\mathcal{O}_{0}}^{\Sigma_{i}} \left( \textbf{p}^{\mathcal{O}_{0}}_{i0} - \textbf{p}_{i}^{*} \right)
		\\
		\notag
		e_{i,c} &= 1 - \cos\theta_{\Sigma_{i}}^{\Sigma_{0}}, \quad
		e_{i,s} = \sin\theta_{\Sigma_{i}}^{\Sigma_{0}},
	\end{align}
	where $\textbf{e}_{i,p}\triangleq [e_{i,x}, e_{i,y}, e_{i,z}]^{T}$. We note that although the value of $\textbf{p}^{\mathcal{O}_{0}}_{i0}$, $\sin \theta_{\mathcal{O}_{0}}^{\mathcal{O}_{i}}(t_{0})$ and $\cos \theta_{\mathcal{O}_{0}}^{\mathcal{O}_{i}}(t_{0})$ in (\ref{tracking_error}) can not be measured directly, they can be estimated asymptotically with the estimator (\ref{leader estimator}). 
	Define the estimated formation tracking error for robot $i$ as
	\begin{align}
		\label{estimated_tracking_error}
		\notag
		\hat{\textbf{e}}_{i,p} &=  \hat{\textbf{R}}_{\mathcal{O}_{0}}^{\Sigma_{i}} \left( \left(\hat{\textbf{Q}}^{\mathcal{O}_{i}}_{\mathcal{O}_{0}}(t)\right)^{T} \hat{\textbf{q}}^{\mathcal{O}_{i}}_{i0}(t) - \textbf{p}_{i}^{*} \right)
		\\ 
		\hat{e}_{i,c} &= 1 - \widehat{\cos \theta_{\Sigma_{i}}^{\Sigma_{0}}}(t), \quad
		\hat{e}_{i,s} = \widehat{\sin \theta_{\Sigma_{i}}^{\Sigma_{0}}}(t)
		\\ \notag
		\hat{\textbf{R}}_{\mathcal{O}_{0}}^{\Sigma_{i}} &\triangleq \begin{bmatrix}
			\widehat{\cos \theta_{\Sigma_{i}}^{\mathcal{O}_{0}}}(t), & \widehat{\sin \theta_{\Sigma_{i}}^{\mathcal{O}_{0}}}(t), &0 \\ 
			-\widehat{\sin \theta_{\Sigma_{i}}^{\mathcal{O}_{0}}}(t), & \widehat{\cos \theta_{\Sigma_{i}}^{\mathcal{O}_{0}}}(t), &0 \\
			0, &0, &1 
		\end{bmatrix},
	\end{align}
	where $\widehat{\cos \theta_{\Sigma_{i}}^{\mathcal{O}_{0}}}(t)$ and $\widehat{\sin \theta_{\Sigma_{i}}^{\mathcal{O}_{0}}}(t)$ are defined as
	\begin{align}
		\notag
		\widehat{\cos \theta_{\Sigma_{i}}^{\mathcal{O}_{0}}}(t) &= \widehat{\cos \theta_{\mathcal{O}_{0}}^{\mathcal{O}_{i}}}(t) \cos \phi_{\Sigma_{i}}^{\mathcal{O}_{i}}(t) + \widehat{\sin \theta_{\mathcal{O}_{0}}^{\mathcal{O}_{i}}}(t) \sin \phi_{\Sigma_{i}}^{\mathcal{O}_{i}}(t)\\
		\notag
		\widehat{\sin \theta_{\Sigma_{i}}^{\mathcal{O}_{0}}}(t) &= \widehat{\cos \theta_{\mathcal{O}_{0}}^{\mathcal{O}_{i}}}(t) \sin \phi_{\Sigma_{i}}^{\mathcal{O}_{i}}(t) - \widehat{\sin \theta_{\mathcal{O}_{0}}^{\mathcal{O}_{i}}}(t) \cos \phi_{\Sigma_{i}}^{\mathcal{O}_{i}}(t).
	\end{align}
	Define $\tilde{e}_{i,x} = \hat{e}_{i,x} - e_{i,x}$, $\tilde{e}_{i,y} = \hat{e}_{i,y} - e_{i,y}$, $\tilde{e}_{i,z} = \hat{e}_{i,z} - e_{i,z}$, $\tilde{e}_{i,c} = \hat{e}_{i,c} - (1-\cos \theta_{\Sigma_{i}}^{\Sigma_{0}})$ and $\tilde{e}_{i,s} = \hat{e}_{i,s} - \sin \theta_{\Sigma_{i}}^{\Sigma_{0}}$, define estimation error as,
	\begin{align}
		\label{estimate_error}
		\tilde{\textbf{e}}_{i} = [\tilde{e}_{i,x}, \tilde{e}_{i,y}, \tilde{e}_{i,c}, \tilde{e}_{i,s}].
	\end{align}
	
	The proposed control scheme consists of two stage including localization stage and sharp formation stage.
	Initially, each robot $i$ performs specified circular actions \cite{Zhao2022TRO} to achieve localization enhancement in the localization stage, i.e. collects  measurement data matrix $\textbf{S}_{ij}$ with its neighbor $j$.
	In this stage, for the follower robot $i$, the control input is designed as
	\begin{align}
		\label{local controller1}
		\left\{
		\begin{aligned}
			v_{i,h} &=r_{i}w_{i}, \quad v_{i,z} = c_{i,v} \sin(c_{i,v}t) \\
			w_{i} &= c_{i,w},
		\end{aligned}
		\right.
	\end{align}
	where $r_{i}$ is the circular radius, $c_{i,v}$ and $c_{i,w}$are positive constant.
	%In this article, the robot localization enhancement motion is designed as a circular motion. 
	%In fact, the fastest convergence of the relative localization algorithm can be achieved by designing the localization enhancement motion such that $\frac{\lambda_{\min}(\textbf{S}_{ij})}{\lambda_{\max}(\textbf{S}_{ij})}$ is maximum. 
	%But this is beyond the scope of this article and will be discussed in future work.
	For robot $i$, the first control stage ends when all $S_{ij}, j \in \mathbb{N}_{i}$ data matrices are full rank. 
	The second control stage begin when all the robots end their first stages. 
	When the second stage control start can be determined with methods such as hop-count propagation \cite{Ze2024Arxiv}.
	In the second control stage, for the follower robot $i$, the control input is designed as
	\begin{align}
		\label{local controller}
		\left\{
		\begin{aligned}
			v_{i,h} &= v_{0,h} - k_{1}\hat{e}_{i,x} + k_{2}w_{0}\hat{e}_{i,y}\\
			v_{i,z} &= v_{0,z} - k_{4}\hat{e}_{i,z}, \quad w_{i} = w_{0} - k_{3}\hat{e}_{i,s}.
		\end{aligned}
		\right.
	\end{align}
	where $k_1>0$, $k_2>0$, $k_3>0$ and $k_4>0$ are the control gain.
	Note that only the estimated formation tracking errors (\ref{estimated_tracking_error}) are adopoted in controller (\ref{local controller}).
	To achieve the formation control task, the following assumption should be satisfied,
	\begin{assumption}
		\label{leader vw1}
		The angle velocity $w_{0}$ is persistently exciting for all $t \geq t_{0}$.
	\end{assumption}
	\begin{figure}[!t]\centering
		\includegraphics[scale=1]{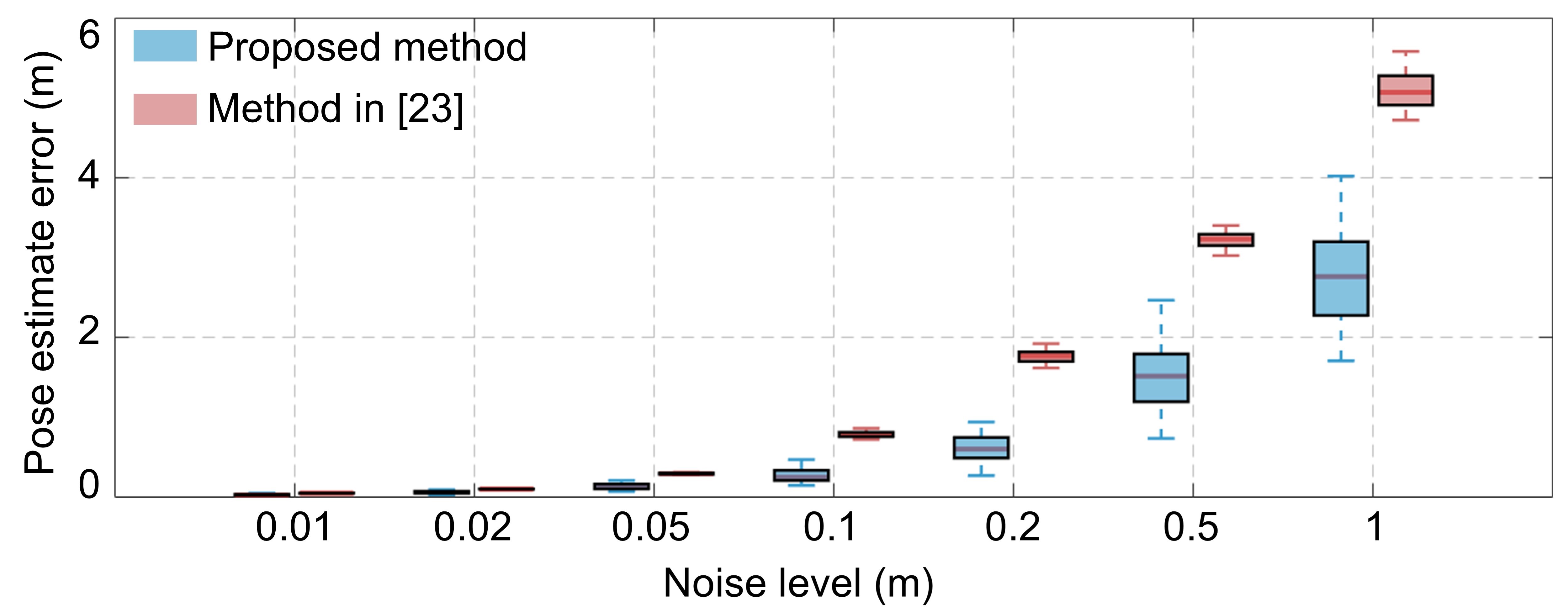}
		\caption{Pose estimation error defined in (\ref{estimate_error}) under different noise level.}\centering
		\label{S3_1}
		\vspace{-0.5cm}
	\end{figure}
	
	\begin{figure}[!t]\centering
		\includegraphics[scale=1]{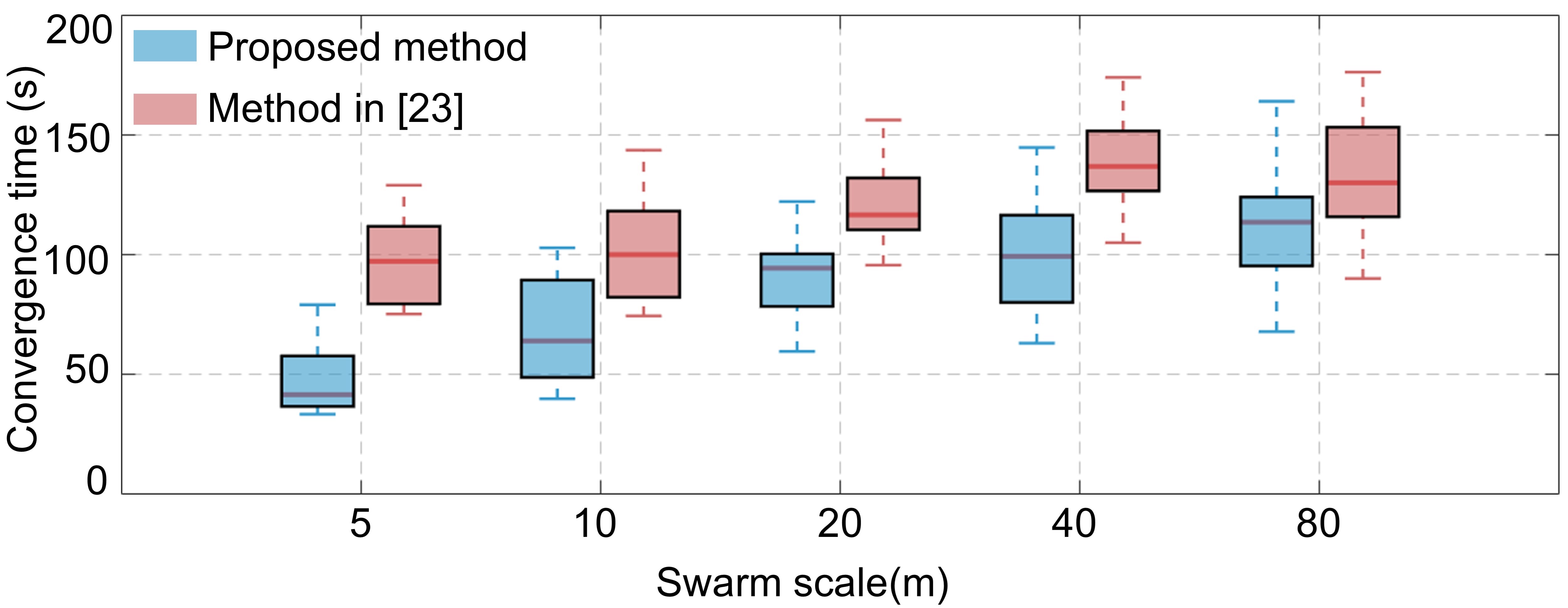}
		\caption{Convergence time (time required for the relative error of the estimated value to be less than 5\%) of the two algorithm under different swarm scales.}\centering
		\label{S3_2}
		\vspace{-0.5cm}
	\end{figure}
	
	\begin{figure}[!t]\centering
		\includegraphics[scale=1]{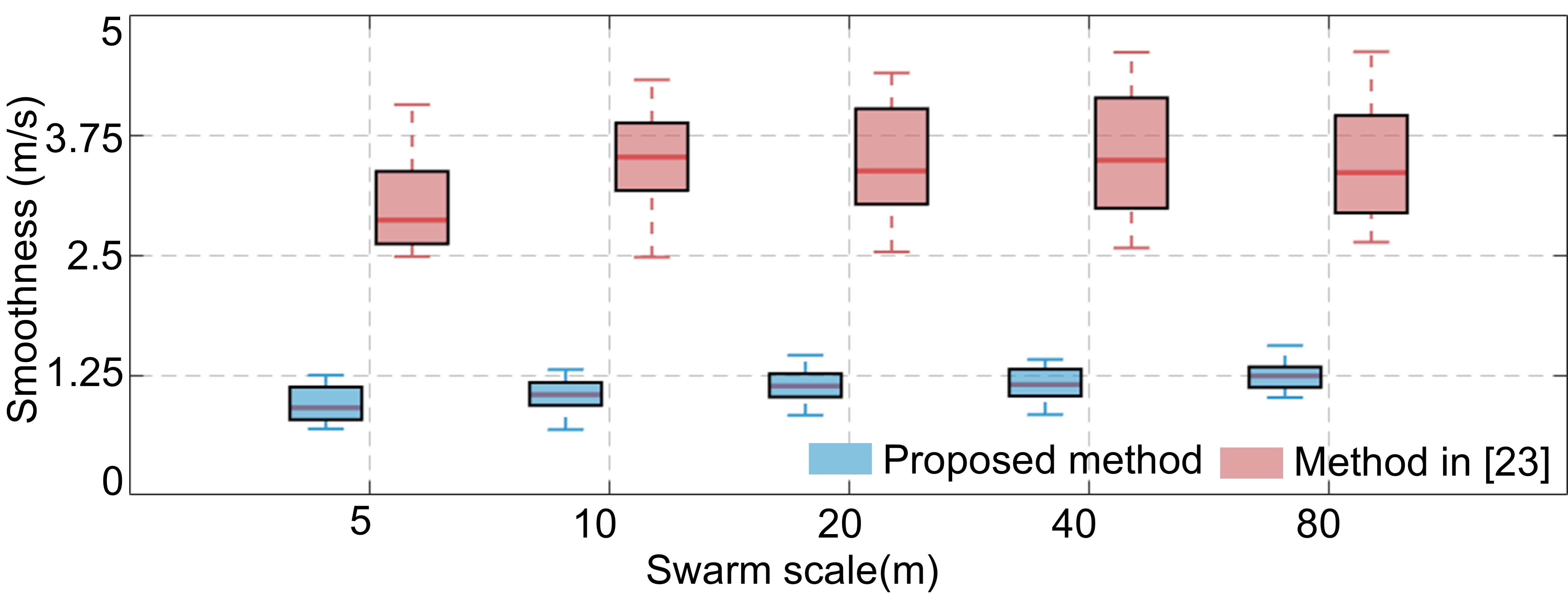}
		\caption{Smoothness indicator defined in Eq.(\ref{smoothness}) under different swarm scale. }\centering
		\label{S3_3}
		\vspace{-0.5cm}
	\end{figure}
	
	\blue{It is worth to point out that the persistently exciting angle velocity $w_{0}$ is a common condition to achieve asymptotically tracking control for nonholonomic robots and can be found in related results \cite{Zhu2023TCNS}\cite{Zhao2023TM}. 
		Furthermore, the requirement for $w_{0}$ to meet the PE condition is only for the asymptotically tracking control of nonholonomic robot systems, whereas the convergence of the proposed estimators (\ref{relative estimator}) or (\ref{leader estimator}) does not require this condition.}
	To better illustrate the calculation process of algorithm data collection, cooperative localization, and control instructions, we have added pseudocode for the proposed method in Appendix \ref{Technical details}-B.
	The stability of the proposed controller can be given in the following theorem.
	\begin{theorem}
		Under Assumption \ref{leader vw} and \ref{leader vw1}, given $(i,j) \in \mathcal{G}$, the recorded data matrix $S_{ij}$ satisfies Assumption \ref{record data}. consider the closed-loop system consisting of $N+1$ nonholonomic robots (\ref{model}), the relative pose estimator (\ref{leader estimator}) and local controller (\ref{local controller}), then the tracking error $\textbf{e}_{i}$ asymptotically converge to zero. 
	\end{theorem}
	\begin{proof}
		See Appendix \ref{Theoretical analysis}-C.
	\end{proof}
	\begin{figure}[!t]\centering
		\includegraphics[scale=1]{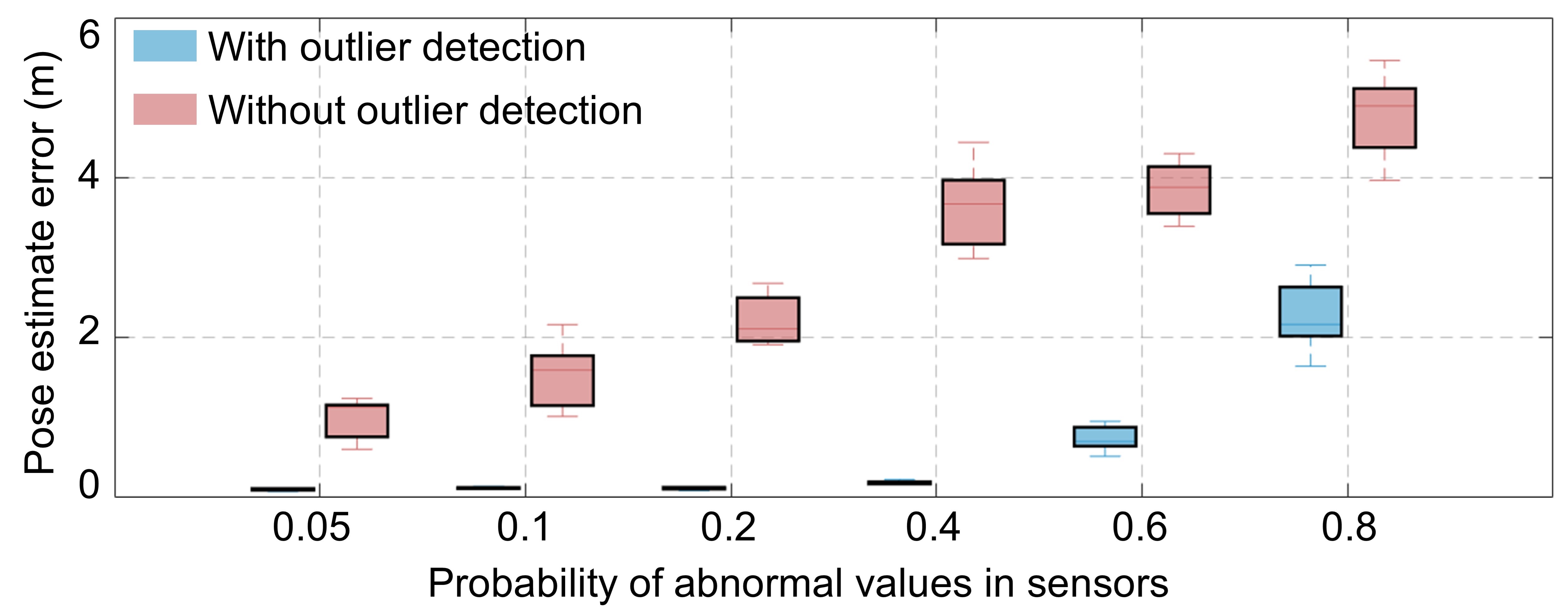}
		\caption{Estimation accuracy of localization method under different sensor outlier probabilities.}\centering
		\label{S5_1}
		\vspace{-0.5cm}
	\end{figure}
	
	\begin{figure}[!t]\centering
		\includegraphics[scale=1]{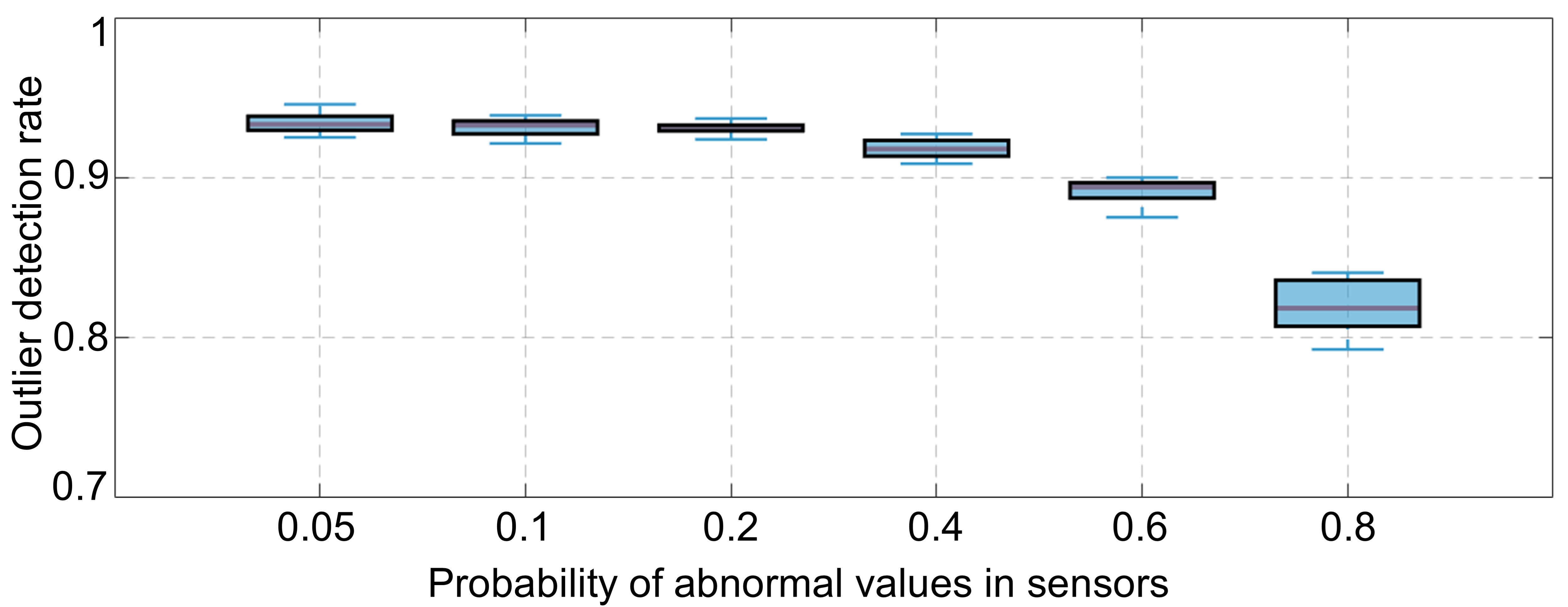}
		\caption{Detection success rate of outlier detection algorithm under different sensor outlier probabilities.}\centering
		\label{S5_2}
		\vspace{-0.5cm}
	\end{figure}
	
	\section{Simulation Results}
	In this section, we present the simulation results to demonstrate the effectiveness and robustness of the proposed method.
	\subsection{Performance of the PE-free relative localization method}
	\label{Performance of the PE-free relative localization method}
	
	\blue{To evaluate the robustness of the proposed relative localization method to measurement noise, we carried out comparative simulations in 3D scenarios against the recent method reported in \cite{Liu2023Auto}.}
	In scenarios involving two robots, we systematically introduced standard deviations ranging from 0.01m to 1m into the measurements of distance and odometer displacement.
	For each noise case, we established 100 different simulation scenarios by varying initial conditions.
	The parameters used in these simulations are $\Delta t=0.05$, $r_{0}=2$, $r_{0}=3$, $c_{0,w}=0.1$, $c_{1,w}=0.3$, $c_{0,v}=0.2$, $c_{1,v}=0.3$, $k_1=1$, $k_2=0.5$, $k_3=0.4$, $k_4=0.2$.
	For further details on parameter selection and parameter sensitivity analysis, please refer to Appendix \ref{supplemental}-A.
	The parameters employed in the comparative simulations can be referenced from \cite{Liu2023Auto}.
	%The simulation results are shown in Fig. \ref{S3_1}, which reveals that as the measurement noise level increases, the localization accuracy of both algorithms declines. 
	%But the performance of the PE based method in \cite{Liu2023Auto} is still inferior to that of the proposed data-driven algorithm. 
	%That is because the data-driven method can tolerate abnormal measurement values to a certain extent.
	Fig. \ref{S3_1} shows that the localization accuracy of both methods degrades as the measurement noise increases, while the proposed data-driven method still outperforms the PE-based method in \cite{Liu2023Auto}, owing to its stronger tolerance to abnormal measurements.
	
	Next, we examine the convergence speed and control performance of both our proposed method and the latest method \cite{Liu2023Auto} in the 3D scenarios.
	Across various swarm scales, we set up 100 formation simulation scenarios with different initial conditions. 
	The parameters for these simulations are $\Delta t=0.05$, $r_{i}=2 + 2\mathrm{rand}(-0.5,0.5)$, $c_{i,w}=4\mathrm{rand}(-0.5,0.5)$, $\mathrm{rand}(0,1)$, $k_1=1$, $k_2=0.5$, $k_3=0.4$, $k_4=0.2$.
	The function $\mathrm{rand}(a,b)$ represents s random value between $a$ and $b$.
	In the comparative simulation, we maintained identical conditions except for the localization and control algorithms under analysis. \blue{The control performance is evaluated based on the smoothness of the motion trajectory, defined by the standard root-mean-square (RMS) velocity tracking error:}
	\begin{align}
		\label{smoothness}
		S_{i} = \sqrt{\frac{1}{t_{\mathrm{end} - t_{0}}}\int_{t_{0}}^{t_{\mathrm{end}}}  \lVert \bm{v}_{i} - \bm{v}_{\mathrm{leader}} \rVert^2  \mathrm{d}t}
	\end{align}
	where $t_{0}$ and $t_{\mathrm{end}}$ denote the start and end time of the simulation, $\bm{v}_{i}$ and $\bm{v}_{\mathrm{leader}}$ are the velocity of robot $i$ and the leader robot. 
	\begin{figure}[!t]\centering
		\includegraphics[scale=1.07]{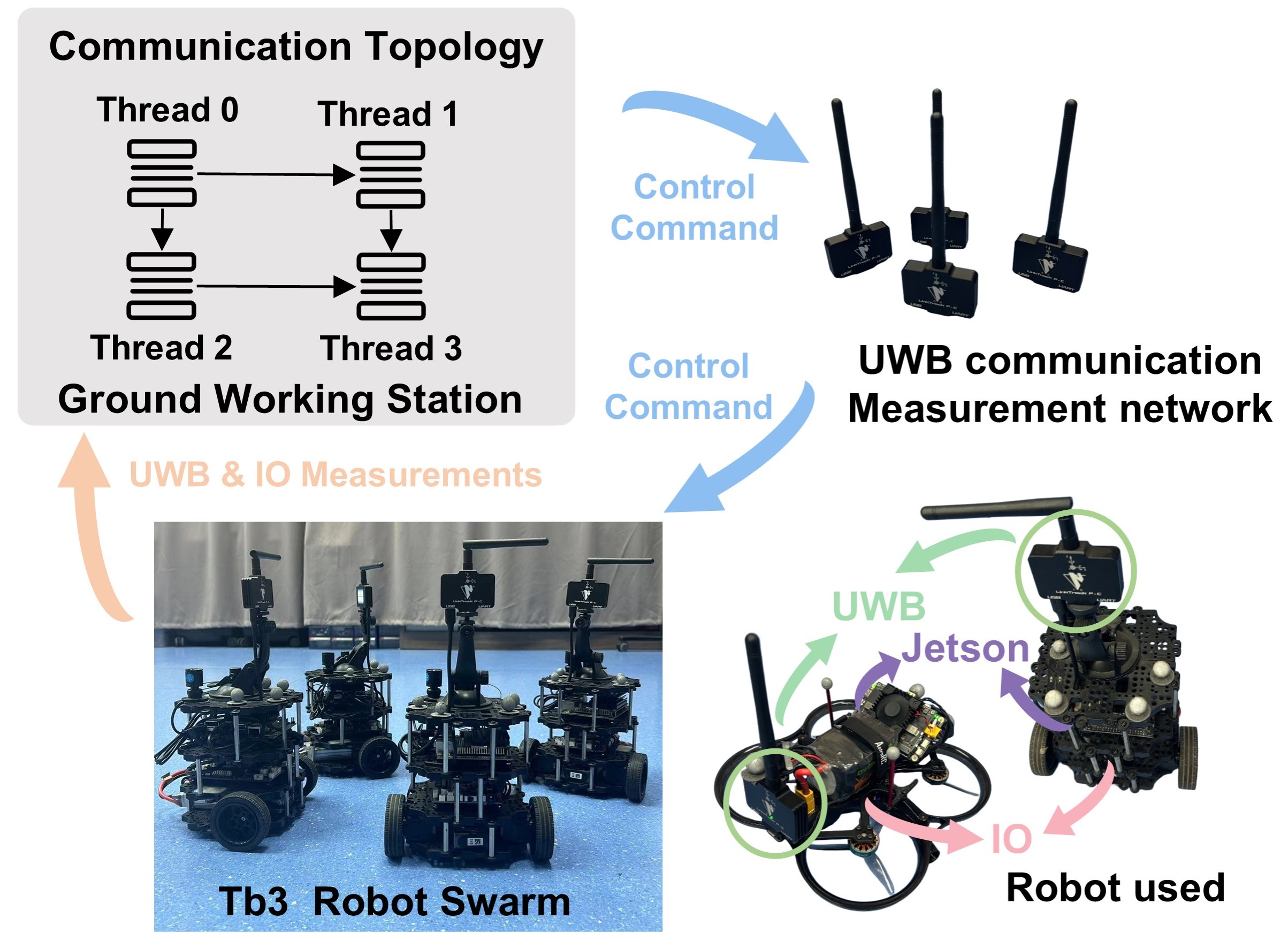}
		\caption{Diagram of the experimental system.}\centering
		\label{setup}
		\vspace{-0.5cm}
	\end{figure}
	For first-order nonholonomic system, the smoothness indicator (\ref{smoothness}) quantifies how much the speed of robot $i$ deviates from the expected speed (i.e. leader speed).
	The convergence time and smoothness indicator (\ref{smoothness}) of the two algorithms are depicted in Fig. \ref{S3_2} and Fig. \ref{S3_3}. 
	%Our results demonstrate that our algorithm consistently exhibits superior convergence speed across different swarm sizes, attributable to its incorporation of both historical and current information, thereby enhancing convergence efficiency. 
	Our results show that the proposed method consistently converges faster across different swarm sizes by exploiting both historical and current information.
	We also analyze the communication bandwidth required for algorithm deployment and discuss the time required for the algorithm to run on a laptop and an onboard computer (NVIDIA Jetson Orin). 
	Please refer to the Appendix \ref{Technical details}-C for details.
	\blue{The proposed method provides smoother trajectories than representative PE-based methods such as \cite{Liu2023Auto}. 
		From a methodological viewpoint, the difference comes from the excitation requirement: PE-based methods such as \cite{Xie2023TRO} and \cite{Liu2023Auto} rely on sufficiently rich relative motion and therefore typically require additional excitation-oriented maneuvers to support localization, while the proposed method relaxes this requirement by using both current and recorded data. As a result, the control input is less aggressive and the resulting trajectories are smoother.
		For a 5-robot swarm, Appendix \ref{supplemental}-B shows that the proposed method yields better control performance than the PE-based method.
	}
	%When the swarm size is 5, in a simulation scenarios, the robot motion trajectories of two algorithms are shown in Appendix \ref{supplemental}-B. It can be clearly seen that our algorithm outperforms the PE based algorithm in terms of control performance.
	
	\subsection{Outlier measurement value detection}
	\label{Outlier measurement value detection}
	\begin{figure}[!t]\centering
		\includegraphics[scale=1.07]{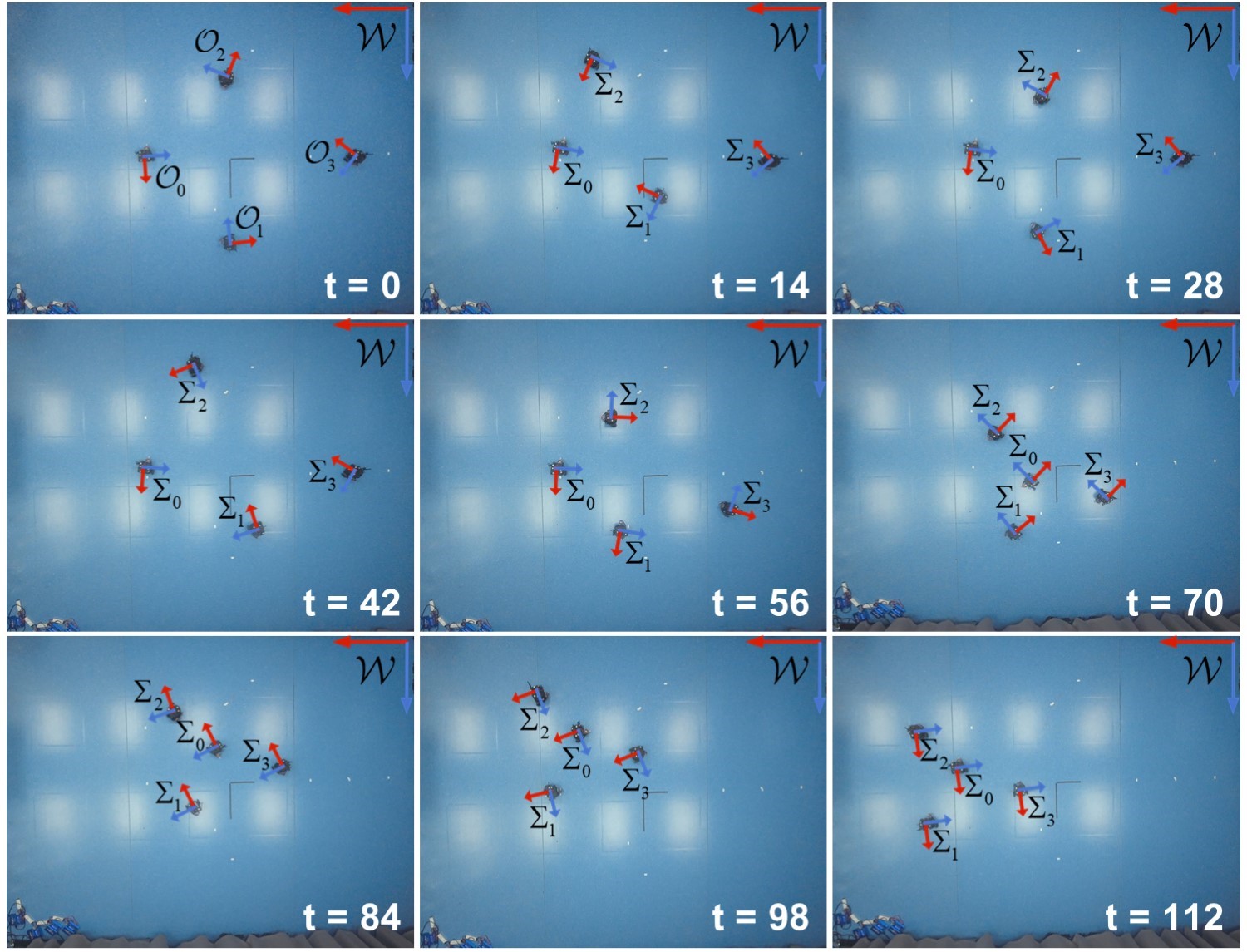}
		\caption{Snapshots of the grounded robot formation control.}\centering
		\label{snapshut2}
		\vspace{-0.5cm}
	\end{figure}
	\begin{figure}[!t]\centering
		\includegraphics[scale=1.07]{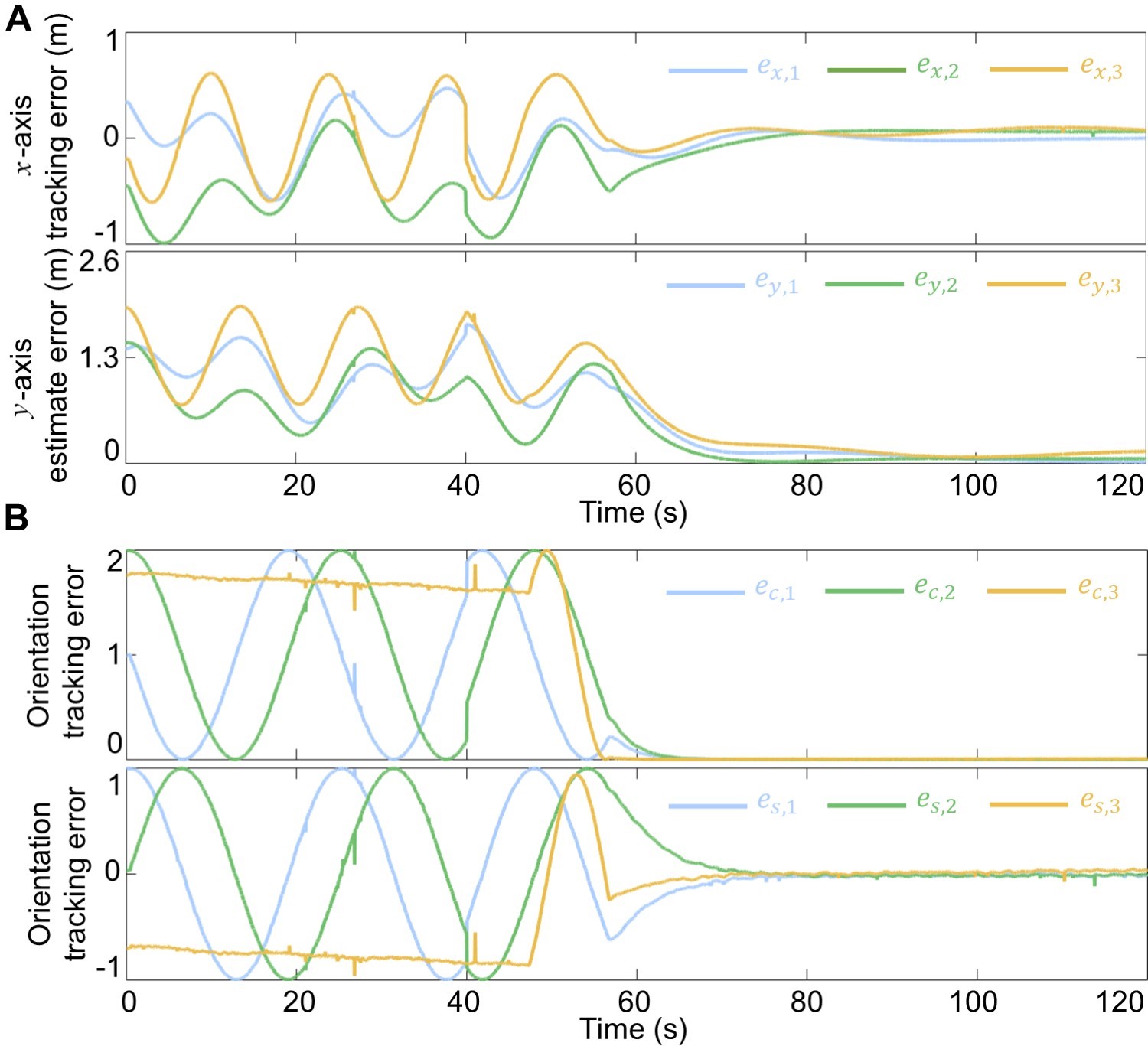}
		\caption{Tracking error of each robot. A. Position tracking error $e_{x,i}$ and $e_{y,i}$ defined in (\ref{tracking_error}) of each robot. B. Orientation tracking error $e_{c,i}$ and $e_{s,i}$ defined in (\ref{tracking_error}) of each robot.}\centering
		\label{tracking error}
		\vspace{-0.5cm}
	\end{figure}
	
	To mitigate the impact of measurement outliers, an outlier detection method is proposed in Appendix \ref{Technical details}-B.
	The core idea of the detection method is based on the following principle: for any nominal distance measurement $d_{ij}[k]$, the absolute value of the difference between $d_{ij}[k]$ and any historical distance $d_{ij}[m]$ should be less than the sum of the robot's displacements between the two measurements, i.e. $\lVert z_{i}^{\mathcal{O}_{i}}[k] - z_{i}^{\mathcal{O}_{i}}[m] \rVert + \lVert z_{j}^{\mathcal{O}_{j}}[k] - z_{j}^{\mathcal{O}_{j}}[m] \rVert$, which is actually a triangular inequality.	
	To validate the outlier detection technique, we conduct 100 sets of relative localization experiments with varying initial conditions. 
	To simulate the realistic situation, we introduce measurement noise with a standard deviation of 0.05 to the UWB and IO measurements. 
	Additionally, the UWB sensor encounters outlier values with a measurement noise having a standard deviation of 3. The probabilities of detecting outlier values are set at 0.05, 0.1, 0.2, 0.4, 0.6, and 0.8, respectively.
	The parameters adopted are the same as those in the two-robot case presented in Section \ref{Performance of the PE-free relative localization method}.  
	Fig \ref{S5_1} and\ref{S5_2} illustrate the pose estimation error with and without outlier detection algorithms, as well as the accuracy of outlier detection. 
	The simulation results demonstrated that the outlier detection algorithms can effectively detect measurement outliers, thus enhancing the accuracy of localization algorithms. More simulation results are provided in Appendix \ref{supplemental}-(C-F), including validation of the effectiveness of the algorithm with dynamic topology robustness, a large scale of swarm, noise robustness, and sensor failures.
	
	\section{Experiment results}
	Both 2-D and 3-D localization and control experiments are provided to verify the effectiveness of the proposed method.
	\subsection{Experiment Setup}
	\begin{figure}[!t]\centering
		\includegraphics[scale=1.07]{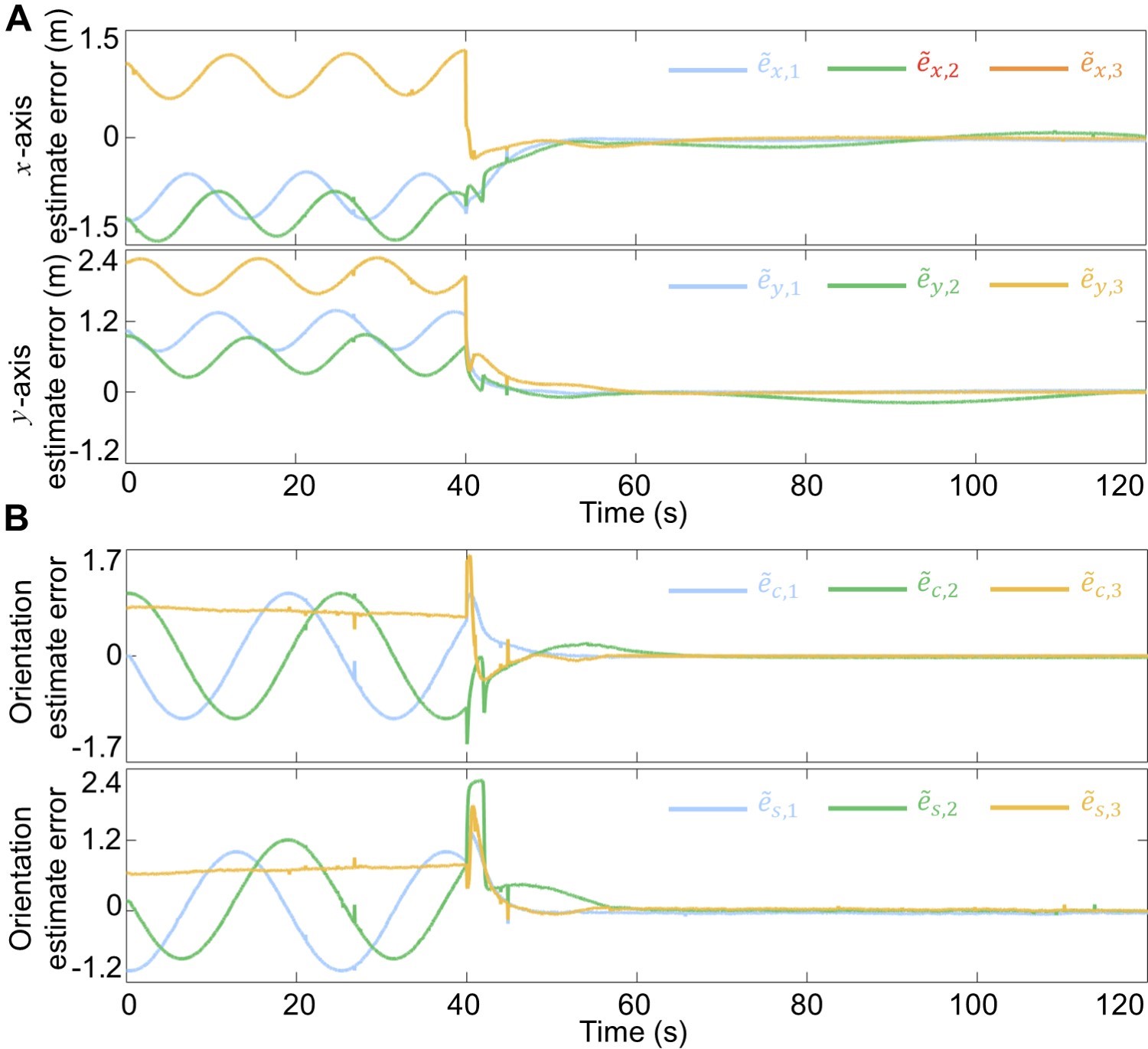}
		\caption{Estimation error of each robot. A. Position estimate error $\tilde{e}_{x,i}$ and $\tilde{e}_{y,i}$ defined in (\ref{estimate_error}) of each robot. B. Orientation estimate error $\tilde{e}_{c,i}$ and $\tilde{e}_{s,i}$ defined in (\ref{estimate_error}) of each robot.}\centering
		\label{estimation error}
		\vspace{-0.5cm}
	\end{figure}
	\begin{figure}[!t]\centering
		\includegraphics[scale=1]{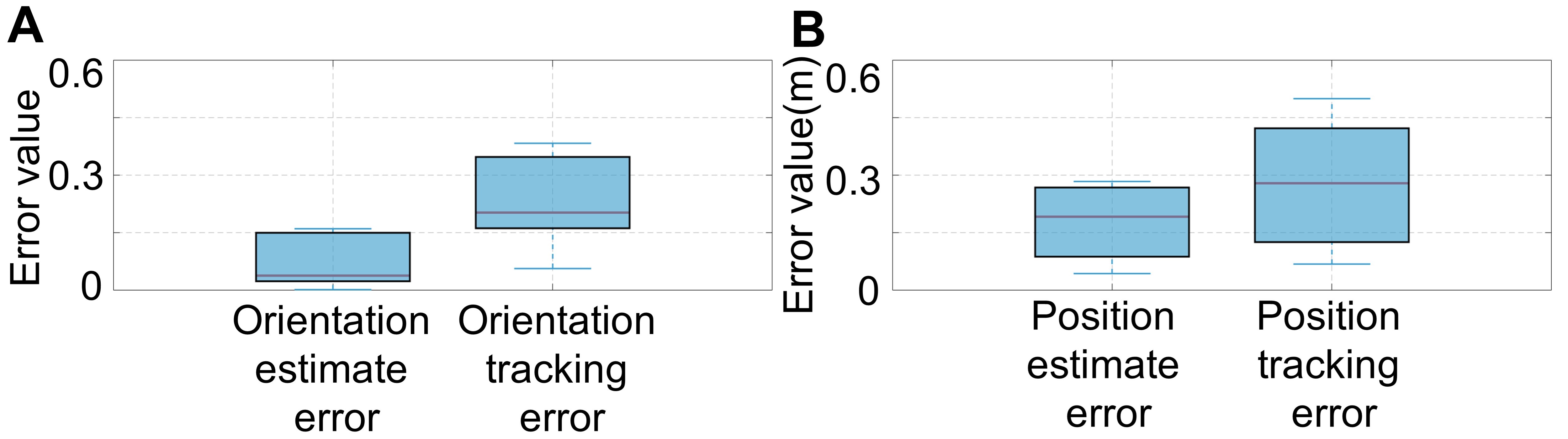}
		\caption{Pose estimation and tracking error in ten experiments with different initial position conditions.}\centering
		\label{E3_1}
		\vspace{-0.5cm}
	\end{figure}
	
	\begin{figure*}[!t]\centering
		\includegraphics[scale=1.1]{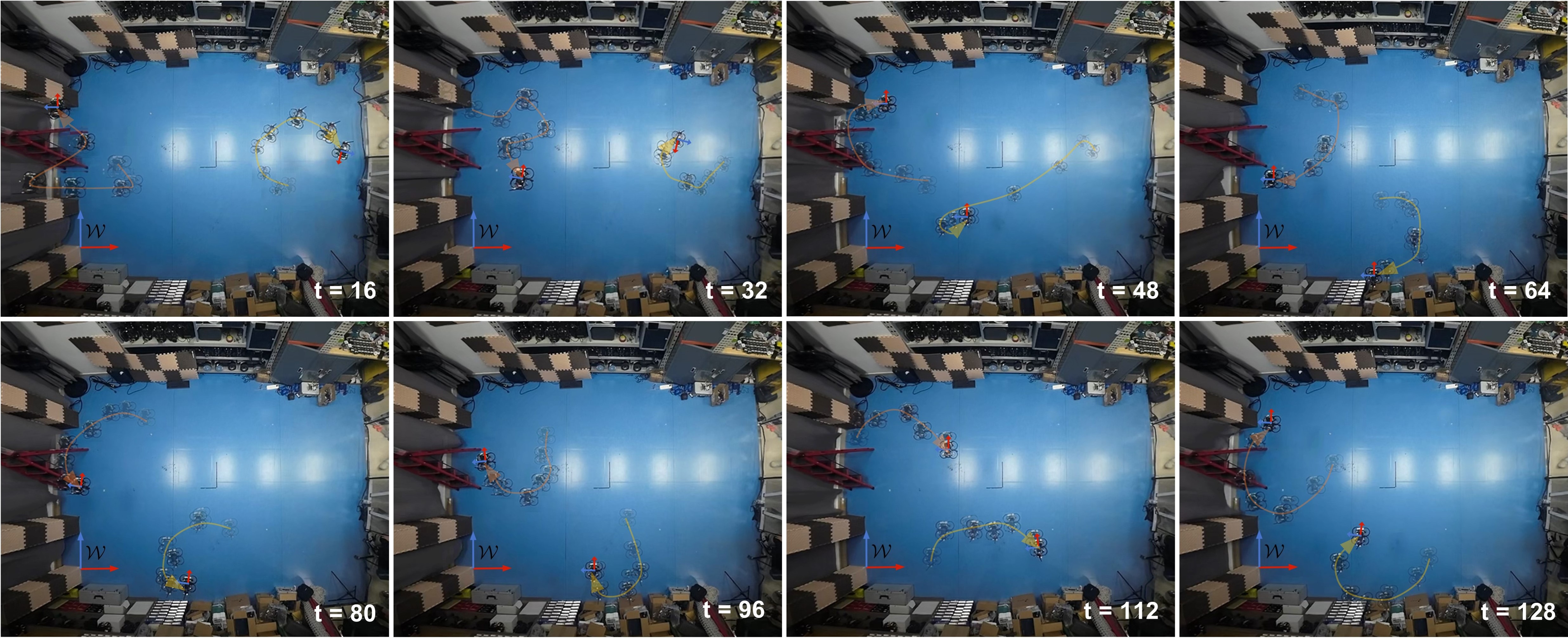}
		\caption{Snapshots of the aerial robot relative tracking experiments.}\centering
		\label{snapshut1}
		\vspace{-0.5cm}
	\end{figure*}
	
	\begin{figure}[!t]\centering
		\includegraphics[scale=1.07]{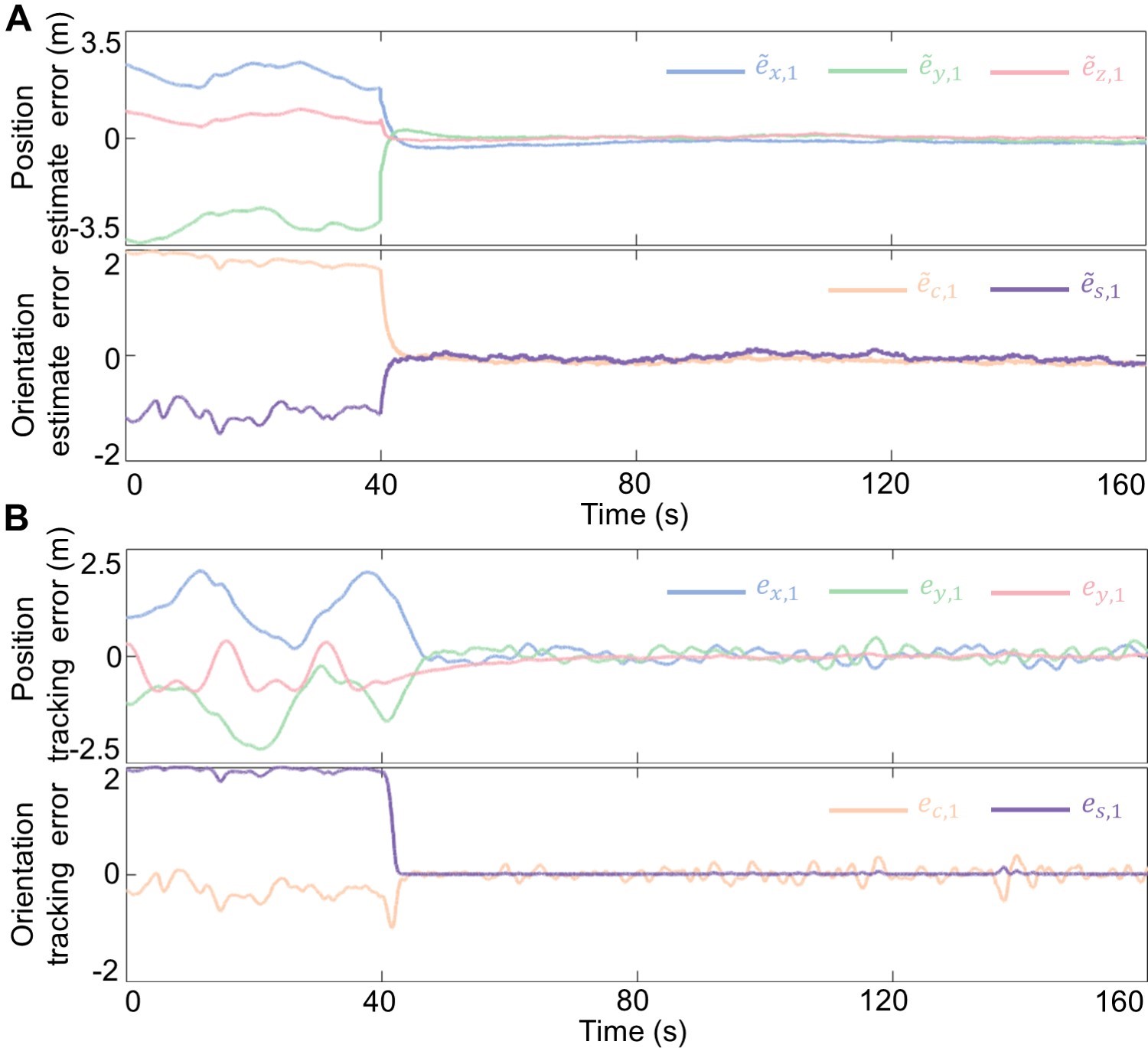}
		\caption{Estimation error and robot tracking error of robot $1$. A. Estimated error $\tilde{\textbf{e}}_{1}$ defined in (\ref{estimate_error}) of robot $1$. B. Tracking error $\textbf{e}_{1}$ defined in (\ref{tracking_error}) of robot $1$.}\centering
		\label{error1}
		\vspace{-0.5cm}
	\end{figure}
	
	In Fig. \ref{setup}, the diagram of the experimental system is illustrated. 
	For the 3D experiments, we utilized an unmanned aerial vehicle based on a PX4 flight controller.
	For the 2D experiments, we employed a nonholonomic robot Turtlebot3 as the experimental platform.
	Each robot is equipped with an IO to measure its displacement and a UWB sensor to measure the real-time distance to its neighbors. 
	For the aerial robot, a Visual inertial odometer (VIO) is adopted to improve the accuracy of IO. 
	The VIO is a technically mature odometer also used in related distance and odometer based relative localization work \cite{Xie2020TRO, Xie2023TRO}.
	Several threads operate on a ground working station, with each thread $i$ receiving the real-time measurement information sent by robot $i$.
	Distributed communication between threads are imitated
	with robot operating system (ROS), and the communication topology is also defined in Fig. \ref{setup}. 
	The proposed distributed relative localization and distributed formation controller for robot $i$ are also executed by thread $i$.
	It is important to note that despite the ground station's involvement in the experiments, the algorithm still operates in a distributed manner.
	
	\subsection{Grounded robot Formation Control Experiment}
	Multi-robot formation results are provided to verify the effectiveness of the localization and control method in 2D situation.
	In this experiment, $3$ follower robots and a leader robot are assigned to achieve the formation task with the prescribed relative position $\textbf{p}_{1}^{*} = [0.5, -0.5, 0]^T$, $\textbf{p}_{2}^{*} = [-0.5, -0.5, 0]^T$ and $\textbf{p}_{3}^{*} = [0, 0.5, 0]^T$ in leader odometer frame $\mathcal{O}_{0}$. 
	Parameter used in the experiment are $\Delta t=0.1$, $r_{i}=0.3,i=0,...,3$, $c_{0,w}=0.5$, $c_{1,w}=0.2$, $c_{2,w}=0.2$, $c_{3,w}=0.5$, $k_1=1$, $k_2=1$, $k_3=0.5$, $k_4=0$.
	After the data collection stage, the leader robot maintain a constant velocity $v_{0}$ and angle velocity $w_{0} $ to achieve a circular formation.
	
	In this experiment, NOKOV motion capture system is used to record the ground truth pose of the leader and follower robots. The snapshuts of the experiment process is shown in Fig. \ref{snapshut2}. 
	The estimated error $\tilde{\textbf{e}}_{i}$ defined in (\ref{estimate_error}) are shown in Fig. \ref{estimation error}. Fig. \ref{estimation error} A shows the position estimate error $\tilde{e}_{x,i}$ and $\tilde{e}_{y,i}$, Fig. \ref{estimation error} B shows the orientation estimate error $\tilde{e}_{c,i}$ and $\tilde{e}_{s,i}$.
	The tracking error $\textbf{e}_{i}$ defined in (\ref{tracking error}) are shown in Fig. \ref{tracking error}.
	Fig. \ref{estimation error} A shows the position tracking error $e_{x,i}$ and $e_{y,i}$, Fig. \ref{estimation error} B shows the orientation estimation error $e_{c,i}$ and $e_{s,i}$.
	One can see that when the measurement data is collected sufficiently, the estimation error and tracking error fast converge. 
	The formation control task is achieved.
	This experiment verifies the effectiveness of the proposed localization algorithm.
	
	To verify the consistency of the algorithm's results under various initial conditions, we conducted Tb3 swarm formation experiments under 10 different initial conditions, with each experiment lasting 150 seconds. 
	We recorded the average estimation and control errors throughout the experiment, as well as the relative pose estimation and tracking control error results, as shown in Fig. \ref{E3_1}. 
	The results indicate that the experimental results at different initial positions exhibit strong consistency. 
	A snapshot of experiments, initiated from three different initial conditions in ten sets of experiments, can be found in Appendix \ref{supplemental}-G.
	
	\subsection{Aerial robot Tracking Control Experiment}
	Both 2-D and 3-D localization and formation control experiments are provided to verify the effectiveness of the method.
	
	In this experiment, the follower robot $1$ is assigned with tracking the leader robot, maintaining a prescribed relative position $\textbf{p}_{1}^{*} \triangleq [1.2, 1.2, 0]^T$ in leader odometer frame $\mathcal{O}_{i}$. 
	The experimental parameters are $\Delta t=0.1$, $r_{0}=0.3$, $r_{1}=0.5$, $c_{0,w}=-0.5$, $c_{1,w}=0.4$, $k_1=0.4$, $k_2=0.8$, $k_3=0.5$, $k_4=0.5$.
	The data collection stage ends when the ratio of the minimum to maximum eigenvalues of the data collection matrix $S_{ij}$ exceeds a  preset threshold of 0.1, which can typically be estimated via simulation.
	
	\begin{figure}[!t]\centering
		\includegraphics[scale=1.07]{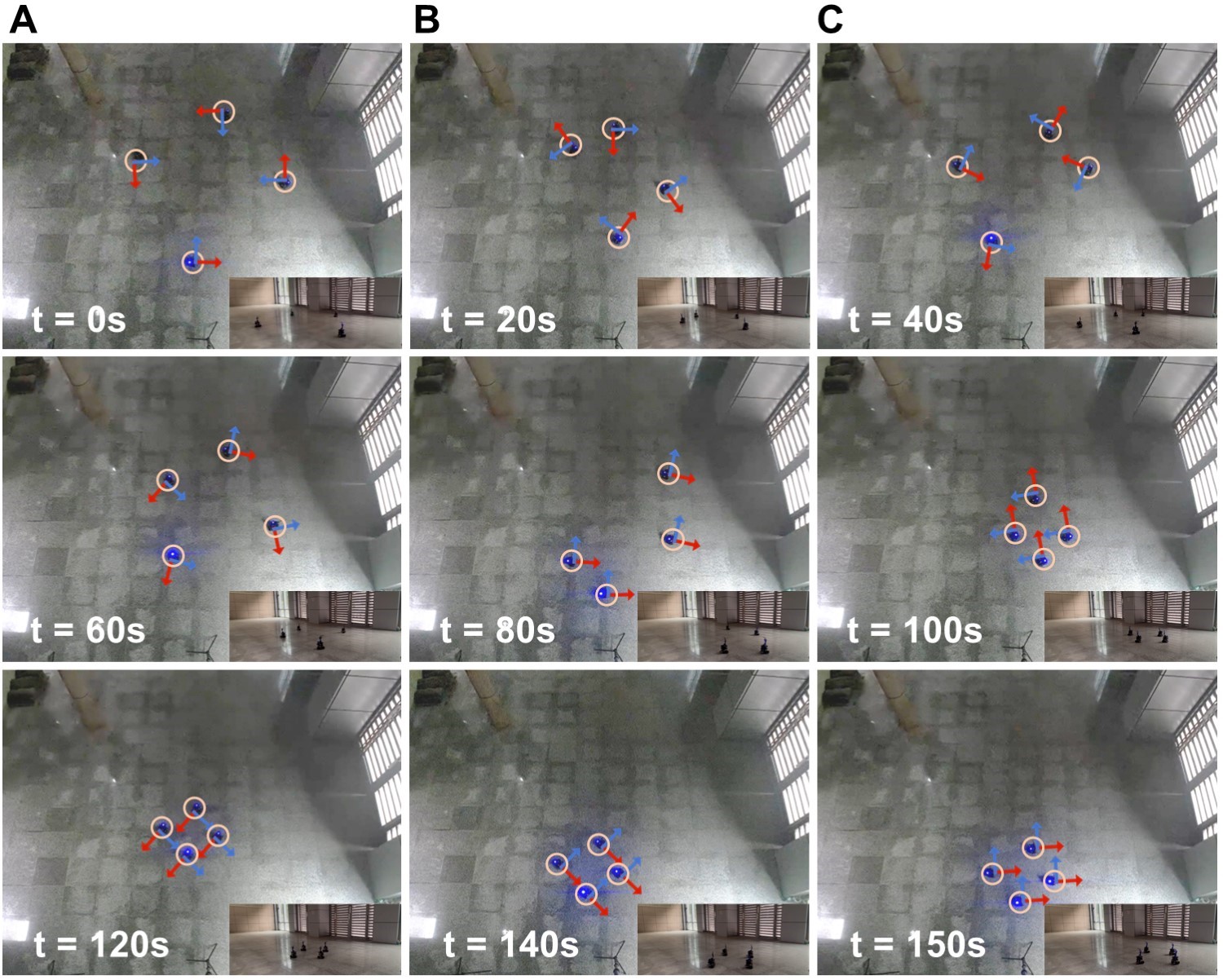}
		\caption{Snapshots of the localization and formation control in the courtyard corridor environment.}\centering
		\label{E4_1}
		\vspace{-0.5cm}
	\end{figure}
	
	Results from the dual UAV tracking experiment demonstrate the effectiveness of the relative localization method in a 3D context.
	In this experiment, a NOKOV motion capture system is used to record the ground truth pose of the leader and follower robots. 
	Snapshots of the experiment process are shown in Fig. \ref{snapshut1} where the trajectories of the leader robot and the follower robot are presented with orange and yellow curves.
	The estimated error $\tilde{\textbf{e}}_{1}$ defined in (\ref{estimate_error}) and tracking error $\textbf{e}_{1}$ defined in (\ref{tracking_error}) are shown in Fig. \ref{error1} A and Fig. \ref{error1} B, respectively. 
	The results indicate that with sufficient data collection, both the estimated error and tracking error converge rapidly. This outcome affirms the effectiveness of the proposed localization algorithm in the 3D tracking scenario. 
	
	%\begin{figure}[!t]\centering
	%	\includegraphics[scale=1.07]{figures/E2_3.jpg}
	%	\caption{Trajectory, velocity and angle velocity of each robot. A. Trajectory of each robot. B. Velocity $v_{i}$ and angle velocity $w_{i}$ of each robot.}\centering
	%	\label{trajectory}
	%	\vspace{-0.5cm}
	%\end{figure}

	\subsection{Outdoor Formation Control Experiment}
	
	To further demonstrate the effectiveness of our algorithm in other environments, we tested the relative localization and formation control algorithm in a building courtyard corridor setting.
	The parameters adopted are consistent with those used in the indoor experiment. 
	A snapshot of the experimental process is shown in Fig. \ref{E4_1}.
	As depicted, the robot swarm effectively execute the formation control task based on UWB and IO measurements, verifying the effectiveness of our method.
	For more complex outdoor environments (such as foggy environments, rainy environments, etc.), UWB-IO fusion based localization systems can be studied in the future.
	
	\section{Conclusion}
	This article has addressed the relative localization based distributed formation control for multi nonholonomic robots.
	The motivation arises from applications such as object transport in indoor and underground environments.
	\blue{First, a data-driven RL estimator is proposed to achieve inter-robot RL task within local frame through only UWB and IO measurements.
		Second, to determine the relative pose between each robot and the leader robot, a cooperative localization strategy is introduced.
		Third, with the proposed data-driven localization method, distributed formation control for nonholonomic robots is achieved with only UWB distance and IO measurements.
		Both theoretical analysis and experimental results are provided to validate the effectiveness and adaptability of the proposed methods.
		Future work includes relative localization under asynchronous measurement conditions, as well as localization and obstacle avoidance in cluttered environments.}
	
	\appendices
	\section{Theoretical analysis}
	\label{Theoretical analysis}
	\label{theorem}
	\subsection{Proof of Theorem 1}
		\label{proof1}
		The dynamics of the relative localization error is
		\begin{align}
			\label{T1_equ1}
			\tilde{\mathbf{\Theta}}_{ij}(t_{k+1}) = \left(I - \eta U_{ij}(t_{k}) - \eta S_{ij} \right) \tilde{\mathbf{\Theta}}_{ij}(t_{k}).
		\end{align}
		Choose a Lyapunov candidate as
		\begin{align}
			\label{V}
			V_{ij}(t_{k}) = \tilde{\mathbf{\Theta}}_{ij}(t_{k})^T\tilde{\mathbf{\Theta}}_{ij}(t_{k}).
		\end{align}
		Define $\Delta V_{ij} = V_{ij}(t_{k+1}) - V_{ij}(t_{k})$, substituting (\ref{T1_equ1}) into (\ref{V}), it has
		\begin{align}
			\notag
			\Delta V_{ij} = &\tilde{\mathbf{\Theta}}_{ij}(t_{k})^{T} \left(-2\eta S_{ij} + 2\eta^2S_{ij}U_{ij}(t_{k}) + \eta^2S_{ij}^2 \right) \tilde{\mathbf{\Theta}}_{ij}(t_{k}) \\
			\notag
			& - \tilde{\mathbf{\Theta}}_{ij}(t_{k})^T \eta U_{ij}(t_{k}) (2I - \eta U_{ij}(t_{k}) ) \tilde{\mathbf{\Theta}}_{ij}(t_{k}).
		\end{align}
		It can be checked that
		\begin{align}
			\notag
			\Delta V_{ij} \leq &\left(-2 \eta \lambda_{\min}(S_{ij}) + 2\eta^2 \lambda_{\max}(U_{ij}(t_{k})) \lambda_{\max}(S_{ij}) \right. \\
			\notag
			& \left. + \eta^2 \lambda_{\max}(S_{ij})^2 \right) \lVert \tilde{\mathbf{\Theta}}_{ij}(t_{k}) \rVert^2 \\
			\notag
			& + \eta^2 \lambda_{\max}(U_{ij}(t_{k}))^2 \lVert \tilde{\mathbf{\Theta}}_{ij}(t_{k}) \rVert^2.
		\end{align}
		With the learning rate $\eta=\frac{\lambda_{min}(\textbf{S}_{ij})}{(\lambda_{max}(\textbf{U}_{ij}(t_{k})) + \lambda_{max}(\textbf{S}_{ij})^2)}$, one can get,
		\begin{align}
			\label{T1_equ2}
			\Delta V_{ij} \leq & \frac{-\lambda_{\min}(S_{ij})^2}{(\lambda_{\max}(U_{ij}(t_{k}))(t_{k}) + \lambda_{\max}(S_{ij})) ^2} \lVert \tilde{\mathbf{\Theta}}_{ij}(t_{k}) \rVert^2.
		\end{align}	
		One can see that for matrix $U_{ij}(t_{k}) = u_{ij}(t_{k})u_{ij}(t_{k})^T \in \mathbb{R}^3$.
		According to Assumption \ref{record data}, it can be check that the maximum eigenvalue satisfies
		\begin{align}
			\label{T1_equ3}
			\lambda_{\max}(U_{ij}(t_{k})) =  \mathbf{\Phi}_{ij}(t_{k})^{T} \mathbf{\Phi}_{ij}(t_{k}) \leq 1.
		\end{align}
		Substituting (\ref{T1_equ2}) into (\ref{T1_equ3}), the relative localization error $\tilde{\mathbf{\Theta}}_{ij}$ satisfies 
		\begin{align}
			\notag
			\lVert \tilde{\mathbf{\Theta}}_{ij}(t_{k}) \rVert \leq \left( \sqrt{\left( 1-\frac{\lambda_{\min}(S_{ij})^2}{1 + \lambda_{\max}(S_{ij})) ^2} \right)} \right)^{k} \lVert \tilde{\mathbf{\Theta}}_{ij}(t_{0}) \rVert.\\
		\end{align} 
		which means that $\tilde{\mathbf{\Theta}}_{ij}$ is global exponentially stable.
	
	\subsection{Proof of Theorem 2}
		\label{proof2}
		Under the DAG topology, Each robot $i$ in layer $l+1$ has at least one neighbor $j$ in layer $l$.
		For the robots in layer $1$, the leader state estimators $\hat{\textbf{q}}^{\mathcal{O}_{i}}_{i0,0}$ and $\hat{\textbf{Q}}^{\mathcal{O}_{i}}_{\mathcal{O}_{0}}$ are directly set as the relative pose estimators $\hat{\textbf{p}}^{\mathcal{O}_{i}}_{i0,0}$ and $\hat{\textbf{R}}^{\mathcal{O}_{i}}_{\mathcal{O}_{0}}$. According to Theorem \ref{relative localization}, the estimation error $\lVert \tilde{\textbf{q}}^{\mathcal{O}_{i}}_{i0,0} \rVert$ and $\lVert \tilde{\textbf{Q}}^{\mathcal{O}_{i}}_{\mathcal{O}_{0}} \rVert$ converge to zero asymptotically. 
		Next, we will use mathematical induction to prove the convergence of $\lVert \tilde{\textbf{q}}^{\mathcal{O}_{i}}_{i0,0} \rVert$ and $\lVert \tilde{\textbf{Q}}^{\mathcal{O}_{i}}_{\mathcal{O}_{0}} \rVert$. 
		Assuming that for robot $i$ in layer $l$, the estimation error $\lVert \tilde{\textbf{q}}^{\mathcal{O}_{i}}_{i0,0} \rVert$ converges to zero asymptotically. For robot $i$ in layer $l+1$, it has
		\begin{align}
			\label{T2_equ1}
			\notag
			\tilde{\textbf{q}}^{\mathcal{O}_{i}}_{i0,0}(t) =& 
			%\sum_{j \in \mathcal{N}_{i}} \frac{1}{|\mathcal{N}_{i}|} \left( \textbf{p}^{\mathcal{O}_{i}}_{ij,0} + \textbf{R}^{\mathcal{O}_{i}} _{\mathcal{O}_{j}} \textbf{p}^{\mathcal{O}_{j}}_{j0} \right) \\
			%\notag
			%&- \sum_{j \in \mathcal{N}_{i}} \frac{1}{|\mathcal{N}_{i}|} \left( \hat{\textbf{p}}^{\mathcal{O}_{i}}_{ij,0}(t) + \hat{\textbf{R}}^{\mathcal{O}_{i}} _{\mathcal{O}_{j}}(t) \hat{\textbf{q}}^{\mathcal{O}_{j}}_{j0}(t) \right) \\
			%\notag
			\sum_{j \in \mathcal{N}_{i}} \frac{1}{|\mathcal{N}_{i}|} \left( \tilde{\textbf{p}}^{\mathcal{O}_{i}}_{ij,0}(t) - \tilde{\textbf{R}}^{\mathcal{O}_{i}} _{\mathcal{O}_{j}}(t) \textbf{p}^{\mathcal{O}_{j}}_{j0} - \hat{\textbf{R}}^{\mathcal{O}_{i}} _{\mathcal{O}_{j}}(t) \tilde{\textbf{q}}^{\mathcal{O}_{j}}_{j0}(t)\right) \\
			\notag
			\lVert \tilde{\textbf{q}}^{\mathcal{O}_{i}}_{i0,0}(t) \rVert \leq& \sum_{j \in \mathcal{N}_{i}} \frac{1}{|\mathcal{N}_{i}|} \left( \lVert\tilde{ \textbf{p}}^{\mathcal{O}_{i}}_{ij,0}(t) \rVert + \lVert \tilde{\textbf{R}}^{\mathcal{O}_{i}} _{\mathcal{O}_{j}}(t) \rVert \lVert \textbf{p}^{\mathcal{O}_{j}}_{j0} \rVert \right. \\
			&\left. + \lVert \hat{\textbf{R}}^{\mathcal{O}_{i}} _{\mathcal{O}_{j}}(t) \rVert \lVert \tilde{\textbf{q}}^{\mathcal{O}_{j}}_{j0}(t) \rVert \right).
		\end{align}  
		One can see that the norm of the normalized rotation matrix $\lVert \hat{\textbf{R}}^{\mathcal{O}_{i}} _{\mathcal{O}_{j}}(t) \rVert$ in (\ref{T2_equ1}) is bounded. Under Assumption \ref{record data} and Theorem \ref{relative localization}, $\lVert\tilde{ \textbf{p}}^{\mathcal{O}_{i}}_{ij,0}(t) \rVert$ converges to zero asymptotically. 
		Under the assuming convergence of $\lVert \tilde{\textbf{q}}^{\mathcal{O}_{j}}_{j0,0} \rVert$ for $j$ in layer $l$, $\lVert \tilde{\textbf{q}}^{\mathcal{O}_{i}}_{i0,0}(t) \rVert$ converges asymptotically.
		Similarly, assuming that for robot $i$ in layer $l$, the estimation $\lVert \tilde{\textbf{Q}}^{\mathcal{O}_{i}}_{\mathcal{O}_{0}} \rVert$ converges to zero asymptotically. Then for robot $i$ in layer $l+1$, it has
		\begin{align}
			\label{T2_equ2}
			\tilde{\textbf{Q}}^{\mathcal{O}_{i}}_{\mathcal{O}_{0}}(t) 
			=& \sum_{j \in \mathcal{N}_{i}} \frac{1}{|\mathcal{N}_{i}|} \left( \textbf{R}^{\mathcal{O}_{i}}_{\mathcal{O}_{0}} - \hat{\textbf{R}}^{\mathcal{O}_{i}} _{\mathcal{O}_{j}}(t) \hat{\textbf{Q}}^{\mathcal{O}_{j}}_{\mathcal{O}_{0}}(t) \right) \\
			\notag
			=& -\sum_{j \in \mathcal{N}_{i}} \frac{1}{|\mathcal{N}_{i}|} \left( \tilde{\textbf{R}}^{\mathcal{O}_{i}}_{\mathcal{O}_{j}}(t) \textbf{R}^{\mathcal{O}_{j}}_{\mathcal{O}_{0}}(t) + \hat{\textbf{R}}^{\mathcal{O}_{i}}_{\mathcal{O}_{j}}(t) \tilde{\textbf{Q}}^{\mathcal{O}_{j}}_{\mathcal{O}_{0}}(t) \right), \\
			\notag
			\lVert \tilde{\textbf{Q}}^{\mathcal{O}_{i}}_{\mathcal{O}_{0}}(t) \rVert \leq& \sum_{j \in \mathcal{N}_{i}} \frac{1}{|\mathcal{N}_{i}|} \Big( \lVert \tilde{\textbf{R}}^{\mathcal{O}_{i}}_{\mathcal{O}_{j}}(t) \rVert \lVert \textbf{R}^{\mathcal{O}_{j}}_{\mathcal{O}_{0}}(t) \rVert \\
			&+ \lVert \hat{\textbf{R}}^{\mathcal{O}_{i}}_{\mathcal{O}_{j}}(t)\rVert \lVert \tilde{\textbf{Q}}^{\mathcal{O}_{j}}_{\mathcal{O}_{0}}(t) \rVert \Big).
		\end{align}
		Under Assumption \ref{record data} and Theorem \ref{relative localization}, $\lVert \tilde{\textbf{R}}^{\mathcal{O}_{i}}_{\mathcal{O}_{j}}(t) \rVert$ in (\ref{T2_equ2}) converges asymptotically. Under the assuming convergence of $\lVert \tilde{\textbf{Q}}^{\mathcal{O}_{j}}_{\mathcal{O}_{0}}(t) \rVert$ for robot $j$ in layer $l$, $\lVert \tilde{\textbf{Q}}^{\mathcal{O}_{i}}_{\mathcal{O}_{0}}(t) \rVert$ converges asymptotically for robot $i$ in layer $l+1$.
	
	\subsection{Proof of Theorem 3}
		\label{proof3}
		The derivative of the tracking error $\textbf{e}_{i}$ is
		\begin{align}
			\label{dot e}
			\dot{\textbf{e}}_{i} = \begin{bmatrix}
				-v_{0,h} + v_{0,h} e_{i,c} + v_{i,h} + w_{i} e_{i,y} \\
				v_{0,h} e_{i,s} - w_{i} e_{i,x} \\
				-v_{0,z} + v_{i,z}\\
				w_{i} - w_{0} - e_{i,c}w_{i} + e_{i,c}w_{0} \\
				e_{i,s}w_{i} - e_{i,s}w_{0}
			\end{bmatrix}.
		\end{align}
		Note that in Eq.(\ref{dot e}), we use a fact that Eq.(\ref{model1}) hold
		\begin{align}
			\label{model1}
			\left\{
			\begin{aligned}
				\dot{\textbf{p}}^{\mathcal{O}_{0}}_{i}(t) &=  \begin{bmatrix}
					\cos \phi_{\Sigma_{i}}^{\mathcal{O}_{0}}(t),& 0\\
					\sin \phi_{\Sigma_{i}}^{\mathcal{O}_{0}}(t),& 0\\
					0,& 1
				\end{bmatrix} 
				\begin{bmatrix}
					v_{i,h} \\
					v_{i,z}
				\end{bmatrix}\\
				\dot{\phi}_{\Sigma_{i}}^{\mathcal{O}_{0}}(t) &= w_{i}.
			\end{aligned}
			\right.
		\end{align}
		With the control law (\ref{local controller}), the derivative of $\textbf{e}_{i}$ is,
		\begin{align}
			\label{T3_equ1}
			\dot{\textbf{e}}_{i} &= F(\textbf{e}_{i}) + G(\textbf{e}_{i}, \tilde{\textbf{e}}_{i}),
		\end{align}
		where $F(\textbf{e}_{i})$ and $G(\textbf{e}_{i}, \tilde{\textbf{e}}_{i})$ in (\ref{T3_equ1}) are defined as
		\begin{align}
			\notag
			F(\textbf{e}_{i}) &= \begin{bmatrix}
				-k_{1}e_{i,x} + (1+k_{2})w_{0}e_{i,y} - k_{3}e_{i,s} + v_{0}e_{i,c} \\
				v_{0} e_{i,s} - w_{0}e_{i,x} - k_{3}e_{i,s}e_{i,x} \\
				-k_{4}e_{i,z} \\
				-k_{3} e_{i,s} + k_{3}e_{i,c}e_{i,s} \\
				-k_{3} e_{i,s}^2
			\end{bmatrix}
		\end{align}
		\begin{align}
			\notag
			G(\textbf{e}_{i}, \tilde{\textbf{e}}_{i}) &= \begin{bmatrix}
				-k_{1}\tilde{e}_{i,x} - k_{3}\tilde{e}_{i,s}e_{i,y} \\
				k_{3}\tilde{e}_{i,s}e_{i,x} \\
				-k_{4}\tilde{e}_{i,z} \\
				-k_{3}(1-e_{i,c})\tilde{e}_{i,s} \\
				-k_{3}e_{i,s}\tilde{e}_{i,s}
			\end{bmatrix}.
		\end{align}
		where $\tilde{e}_{i,x} = \hat{e}_{i,x} - e_{i,x}$, $\tilde{e}_{i,y} = \hat{e}_{i,y} - e_{i,y}$, $\tilde{e}_{i,z} = \hat{e}_{i,z} - e_{i,z}$, $\tilde{e}_{i,c} = \hat{e}_{i,c} - (1-\cos \theta_{\Sigma_{i}}^{\Sigma_{0}})$ and $\tilde{e}_{i,s} = \hat{e}_{i,s} - \sin \theta_{\Sigma_{i}}^{\Sigma_{0}}$ and the estimation error is defined as,
		\begin{align}
			\tilde{\textbf{e}}_{i} = [\tilde{e}_{i,x}, \tilde{e}_{i,y}, \tilde{e}_{i,c}, \tilde{e}_{i,s}].
	\end{align}
	
	According to Theorem \ref{leader localization}, under Assumption \ref{record data} and \ref{leader vw}, 
		it is easy to know that the estimated error $\tilde{\textbf{e}}_{i}$ converge to zero asymptotically.
		According to Proposition 1 in \cite{Zhao2023TM}, under Assumption \ref{leader vw} and \ref{leader vw1}, the nominal tracking error system $\dot{\textbf{e}}_{i} = F(\textbf{e}_{i})$ converge to zero exponentially uniformly.
		Note that the boundness assumption on the leader velocity $v_{0}$ and  $w_{0}>0$ are to ensure the uniformly exponential convergence of the nominal tracking error system $\dot{\textbf{e}}_{i} = F(\textbf{e}_{i})$.
		One can see $G(\textbf{e}_{i}, \tilde{\textbf{e}}_{i})$ satisfies
		\begin{align}
			\lVert G(\textbf{e}_{i}, \tilde{\textbf{e}}_{i}) \rVert \leq \max\{ k_{1},k_{3} \}\lVert \textbf{e}_{i}\rVert \lVert \tilde{\textbf{e}}_{i}\rVert + k_{3} \lVert \textbf{e}_{i}\rVert.
		\end{align}
		According to Lemma \ref{a s}, the formation tracking error $\textbf{e}_{i}$ converge asymptotically.
	
	\subsection{Complete observability analysis of relative localization}
	\label{complete observability}
	
	Taking robot $i$ and its neighbor robot $j$ as an example, we discuss several cases where matrix $S_{ij}$ is not full rank and analyze which states cannot be observed.
	
	When robot $i$ remains stationary all the time, i.e. $v_{i}^{\mathcal{O}_{i}}\equiv0$, both the relative position and the relative yaw is unobservable. If robot $i$ remains stationary all the time, then $\textbf{u}^{\mathcal{O}_{i}}_{i,h}(t_{k}) = \textbf{0}$ and $\textbf{p}^{\mathcal{O}_{i}}_{i,h}(t_{k}) = \textbf{0}$, then $\mathbf{\Psi}_{ij}(t_{k}) = [\textbf{0}_{3\times1}, *_{2\times1}, \textbf{0}_{2\times1}]^T$ and $S_{ij} = \begin{bmatrix} \textbf{0}_{3\times3}, &\textbf{0}_{2\times2}, &\textbf{0}_{2\times2} \\
			\textbf{0}_{3\times3}, &\textbf{*}_{2\times2}, &\textbf{0}_{2\times2} \\
			\textbf{0}_{3\times3}, &\textbf{0}_{2\times2}, &\textbf{0}_{2\times2}
		\end{bmatrix}$. The data matrix block corresponding to the relative position and the sine and cosine of and relative yaw angle is always $\textbf{0}$, indicating that they are unobservable.
	
	When the neighbor robot $j$ remains stationary all the time, i.e. $v_{j}^{\mathcal{O}_{j}}\equiv0$, at least the  relative yaw is unobservable. One can see that if robot $j$ remains stationary all the time, $\textbf{u}^{\mathcal{O}_{j}}_{j,h}(t_{k}) = \textbf{0}$ and $\textbf{p}^{\mathcal{O}_{j}}_{j,h}(t_{k}) = \textbf{0}$, then $S_{ij} = \begin{bmatrix} \textbf{*}_{5\times5}, &\textbf{*}_{5\times2} \\
			\textbf{*}_{2\times5}, &\textbf{0}_{2\times2} 
		\end{bmatrix}$. The data matrix block corresponding to the sine and cosine of and relative yaw angle is always $\textbf{0}$, indicating that they are unobservable.
	\begin{table}[!t]
		\caption{List of main symbols used in this article}
			\centering
			\begin{tabular}{ll}
				\hline\hline \\[-3mm]
				\multicolumn{1}{c}{Symbols} & \multicolumn{1}{c}{Description} \\ \hline \\[-2mm]
				$ t_{k} $  & \pbox{10cm}{UWB \& IO measurement time instant for robot $i$. } \\ [1.3ex] 
				$ \mathcal{O}_{i} $  & \pbox{10cm}{The odometry frame of robot $i$. } \\ [1.3ex] 
				$ \Sigma_{i} $  & \pbox{10cm}{Real-time IMU body frame of robot $i$.} \\ [1.3ex] 
				$ \textbf{p}^{\mathcal{O}_{i}}_{i} $  & \pbox{10cm}{Displacement of robot $i$ in frame
					$\mathcal{O}_{i}$ from $t_{0}$ to $t_{k}$. } \\ [1.3ex] 
				$ \phi_{\Sigma_{i}}^{\mathcal{O}_{i}}(t) $  & \pbox{10cm}{Angular displacement of robot $i$ in frame
					$\mathcal{O}_{i}$ from $t_{0}$ to $t_{k}$. } \\ [1.3ex] 
				$ \textbf{u}^{\mathcal{O}_{i}}_{i}(t_{k}) $  & \pbox{10cm}{Displacement of robot $i$ in frame $\mathcal{O}_{i}$ from $t_{k}$ to $t_{k+1}$ } \\ [1.3ex]
				$ d_{ij}(t_{k}) $ & \pbox{10cm}{Distance between robots $i$ and $j$. } \\ [1.3ex]
				$ y_{ij} $ & \pbox{10cm}{Auxiliary signal which records innovation. } \\ [1.3ex]
				$ \hat{\textbf{p}}^{\mathcal{O}_{i}}_{ij,0} $ & \pbox{10cm}{ Estimate of initial relative position between robot $i$ and its \\ neighbor robot $j$ in frame $\mathcal{O}_{i}$. } \\ [1.3ex]
				$ \hat{\textbf{R}}^{\mathcal{O}_{i}} _{\mathcal{O}_{j}} $ & \pbox{10cm}{ Estimate of initial rotation matrix between frame $\mathcal{O}_{i}$ and $\mathcal{O}_{j}$.} \\ [1.3ex]
				$ \hat{\textbf{q}}^{\mathcal{O}_{i}}_{i0} $ & \pbox{10cm}{ Estimate of initial relative position between robot $i$ and leader \\ robot $0$ in frame $\mathcal{O}_{i}$. } \\ [1.3ex]
				$ \hat{\textbf{Q}}^{\mathcal{O}_{i}}_{\mathcal{O}_{0}} $ & \pbox{10cm}{ Estimate of rotation matrix between Frame $\mathcal{O}_{i}$ and $\mathcal{O}_{0}$. } \\ [1.3ex]
				$ v_{i} $ & \pbox{10cm}{Velocity control command for robot $i$. } \\ [1.3ex]
				\hline\hline \\[-4mm]
			\end{tabular} \label{Table1}
	\end{table}
	
	When robot $i$ and its neighbor robot $j$ remain relatively stationary all the time, i.e. $v_{i}^{\mathcal{O}_{i}}\equiv v_{j}^{\mathcal{O}_{j}}$, at least the z-axis relative position is unobservable. 
		One can see that if robot $i$ and its neighbor robot $j$ remain relatively stationary all the time, then $u^{\mathcal{O}_{i}}_{i,z}(t_{k}) = u^{\mathcal{O}_{j}}_{j,z}(t_{k})$, then $S_{ij} = \begin{bmatrix} \textbf{\textsc{*}}_{2\times2}, &0, &\textbf{*\textsc{0}}_{2\times4} \\
			\textbf{0}_{1\times2}, &0, &\textbf{0}_{1\times4} \\
			\textbf{*}_{4\times2}, &0, &\textbf{0}_{4\times4}
		\end{bmatrix}$. The data matrix block corresponding to the z-axis relative position is always $\textbf{0}$, indicating that it is unobservable.
	
	When the velocity of robot $i$ and its neighbor robot $j$ remain constant, i.e. $v_{i}^{\mathcal{O}_{i}} \equiv v_{i,0}$ and $ v_{j}^{\mathcal{O}_{j}} \equiv v_{j,0}$, then the relative position is unobservable. When the velocity of robot $i$ and its neighbor robot $j$ remain constant, one can see that $\textbf{u}^{\mathcal{O}_{i}}_{i,h}(t_{k})^{T}$ and $u^{\mathcal{O}_{i}}_{i,z}(t_{k}) - u^{\mathcal{O}_{j}}_{j,z}(t_{k})$ are constant. As a result $S_{ij} = \begin{bmatrix} \textbf{A}_{3\times3}, &\textbf{*}_{3\times4} \\
			\textbf{*}_{4\times3}, &\textbf{*}_{4\times4} 
		\end{bmatrix}$, where $rank(A) = 1$, then the relative position is unobservable.
	
	\section{Technical details}
	\label{Technical details}
	\subsection{Notation list of the proposed method}
	For ease of reading, in this section, we provide a table of the main symbols that appear in this article in the following Table \ref{Table1}.
	\subsection{Pseudocode of the proposed method}
	\label{Pseudocode}
	
	The pseudocode of the proposed data-driven localization and formation control method is shown in Algorithm \ref{UWB-IO method}. 
	
	\begin{algorithm}[!t]
			\label{UWB-IO method}
				\small
				\caption{Localization and control algorithm}
				\KwIn{UWB and IO measurement. Neighbor IO measurement. Neighbor estimator $\hat{\textbf{q}}^{\mathcal{O}_{i}}_{i0}$ and $\hat{\textbf{Q}}^{\mathcal{O}_{i}}_{\mathcal{O}_{0}}$.}
				\KwOut{Control command $v_{i,h}$, $v_{i,z}$ and $w_{i}$}
				\BlankLine    	
				\While{$k+1$-th measurement time instant}{
					Localization enhancement flag = false \;
					\For{neighbor $j$ of robot $i$}{
						Outiler measurment data detetion according to Algorithm \ref{outlier detection} \;
						\If{measurement data is not outiler}{
							Collect data according to (\ref{Phi_Theta}) \;
							Update relative pose estimator with (\ref{relative estimator}) \;
							\If{data matrix $S_{ij}$ is not full rank}{
								Localization enhancement flag = True;
							}
						}
					}
					Update cooperative pose estimator with (\ref{leader estimator initial1}) and (\ref{leader estimator initial2}) \;
					\If{localization enhancement flag = True}{
						Calculate control command with (\ref{local controller1}) 
					}
					\Else{
						Calculate control command with (\ref{local controller}
					}	
			}
	\end{algorithm}

	\begin{algorithm}[!t]
		\blue{
		\label{hop count}
			\label{hop}
			\small
			\caption{Multi hop propagation algorithm for leader information}
			\KwIn{Robot ID $i$, Neighbor set $\mathcal{N}_{i}$, Neighbor hop $h_{j}$, Leader information estimated by neighbor $\hat{\sigma}_{j}$ }
			\KwOut{Estimatin of leader information $\hat{\sigma}_{i}$}
			\BlankLine   
			$h_{i}$ $\leftarrow$ $\infty$\;
			\While{$k$-th measurement time instant}{
				\If{ID $i$ == leader ID 0}{ 
					$\hat{\sigma}_{i}$ $\leftarrow$ Leader information\;
					$h_{i}$ $\leftarrow$ $0$\;
				} 	
				\Else{
					find minimum hop value $h_{j}$ in neighbor set$\mathcal{N}_{i}$\;
					$h_{i}$ $\leftarrow$ $h_{j}+1$\;
					$\hat{\sigma}_{i}$ $\leftarrow$ $\hat{\sigma}_{j}$\;
				}
			}}
	\end{algorithm}
	In engineering, in addition to the multi hop propagation method, consensus estimation can also be used to achieve consistent estimation of the leader odometer information for each robot. Specifically, the leader odometer information is used as a reference signal, and a consensus observer is designed by constructing a consensus error to achieve asymptotic estimation of the time-varying signal of the leader odometer information.
	
	In order to reduce the impact of measurement outliers on the localization method, an outlier detection algorithm is proposed in Algorithm \ref{outlier detection}.
		In Algorithm \ref{outlier detection}, we maintain a real-time updated data queue $Q_{ij,\mathrm{judge}}$ (lines 15-17 of the algorithm), with a maximum length of $n_{\mathrm{judge},\mathrm{max}}$. 
		The core idea of the judgment is that the absolute value of the difference between the current measured distance $d_{ij}[k]$ and any distance $d_{ij}[m]$ in the queue should be less than the sum of the robot's displacement between the two measurements, i.e. $\lVert z_{i}^{\mathcal{O}_{i}}[k] - z_{i}^{\mathcal{O}_{i}}[m] \rVert + \lVert z_{j}^{\mathcal{O}_{j}}[k] - z_{j}^{\mathcal{O}_{j}}[m] \rVert$. 
		This is actually a triangular inequality, as stated in lines 5-9 of the algorithm. 
		If this condition is not met, then there is a problem with the current measurement and it should not be used. 
		To avoid randomness, we set a threshold $\eta_{\mathrm{judge}}$. 
		When the percentage of measurements in the queue that exceeds the threshold does not meet this condition i.e. $n_{\mathrm{outlier}} / n_{\mathrm{judge}} > \eta_{\mathrm{judge}}$, it is considered that the current measurement is incorrect, that is an outlier (lines 10-14 of the algorithm).
		To reduce the impact of odometer drift, we update the queue with the latest measurement data (lines 15-17 in the algorithm).  $n_{\mathrm{judge}}=20$ and $\eta_{\mathrm{judge}}=0.5$. 
		\begin{algorithm}[!t]
			\label{outlier detection}
			\small
			\caption{UWB outlier detection algorithm}
				\KwIn{current data triplet $\mathcal{D}_{ij}[k]=\{ d_{ij}[k], z_{i}^{\mathcal{O}_{i}}[k], z_{j}^{\mathcal{O}_{j}}[k] \}$. maximum data queue size $n_{\mathrm{judge}, \mathrm{max}}$. Judgment threshold $\eta_{\mathrm{judge}}$.}
				\KwOut{Outlier detection flag $f_{ij,\mathrm{outlier}}$}
				\BlankLine    	
				Judgment data queue $Q_{ij,\mathrm{judge}}$ $\leftarrow$ empty set $\phi$\;
				\While{$k$-th measurement time instant}{
					Current data triplet $\mathcal{D}_{ij}[k]$ $\leftarrow$ UWB \& IO measurement\;
					Outlier vote number $n_{\mathrm{outlier}}$ $\leftarrow$ 0\;
					\For{data triplet $\mathcal{D}_{ij}[m] \in Q_{\mathrm{judge}}$}{
						\If{$|d_{ij}[k] - d_{ij}[m]| \geq \lVert z_{i}^{\mathcal{O}_{i}}[k] - z_{i}^{\mathcal{O}_{i}}[m] \rVert + \lVert z_{j}^{\mathcal{O}_{j}}[k] - z_{j}^{\mathcal{O}_{j}}[m] \rVert$}{
							$n_{\mathrm{outlier}}$ $\leftarrow$  $n_{\mathrm{outlier}}+1$\;
						}
					}
					Data queue $n_{\mathrm{judge}}$ $\leftarrow$ data triplets number in $Q_{ij,\mathrm{judge}}$\;
					Outlier detection flag initialize $f_{ij, \mathrm{outlier}}$ $\leftarrow$ 1\;
					\If{$n_{\mathrm{outlier}} / n_{\mathrm{judge}} > \eta_{\mathrm{judge}}$}{
						$f_{ij, \mathrm{outlier}}$ $\leftarrow$ 0
					}	
					\If{$f_{ij, \mathrm{outlier}} = 1$}{
						Update $Q_{ij,\mathrm{judge}}$ with the current Measurement data triplet $\mathcal{D}_{ij}[k]$
					}
			}
		\end{algorithm}
	
	\begin{figure}[!t]\centering
		\includegraphics[scale=1]{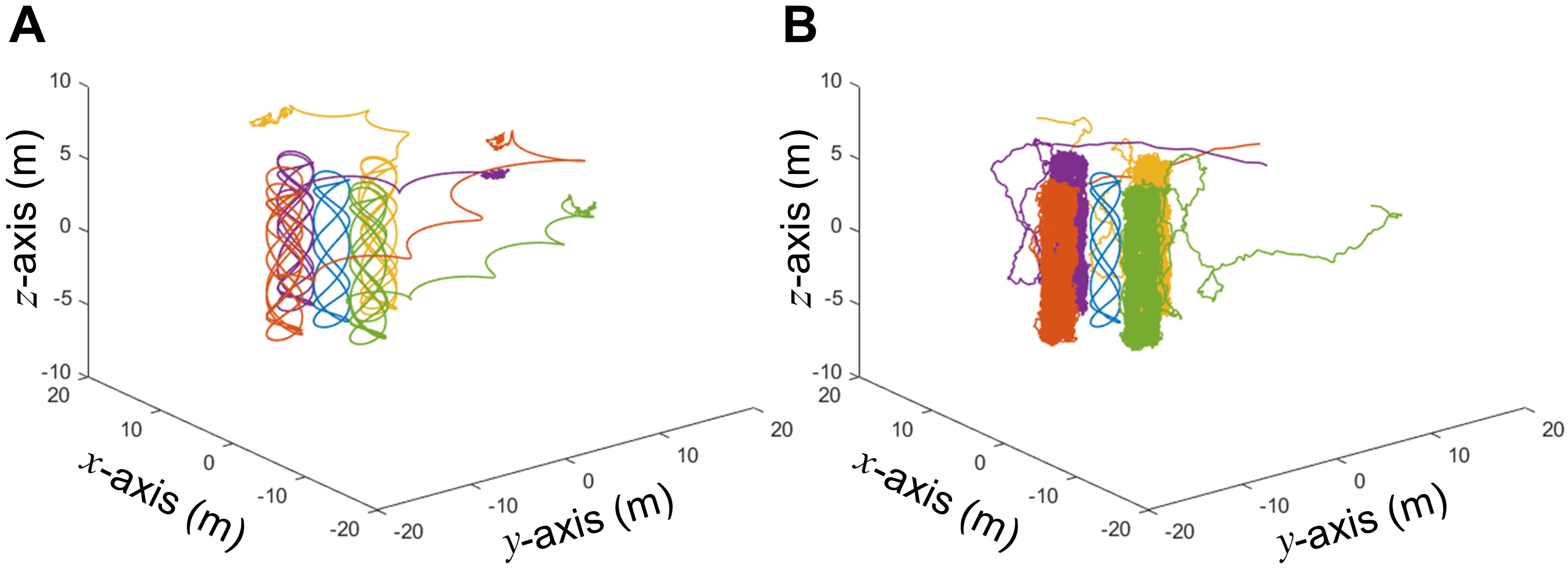}
		\caption{Motion trajectory of the proposed method and method in \cite{Liu2023Auto}. A: Proposed method. B: method in \cite{Liu2023Auto}.}\centering
		\label{S3_4}
		\vspace{-0.5cm}
	\end{figure}
	\begin{figure}[!t]\centering
		\includegraphics[scale=1]{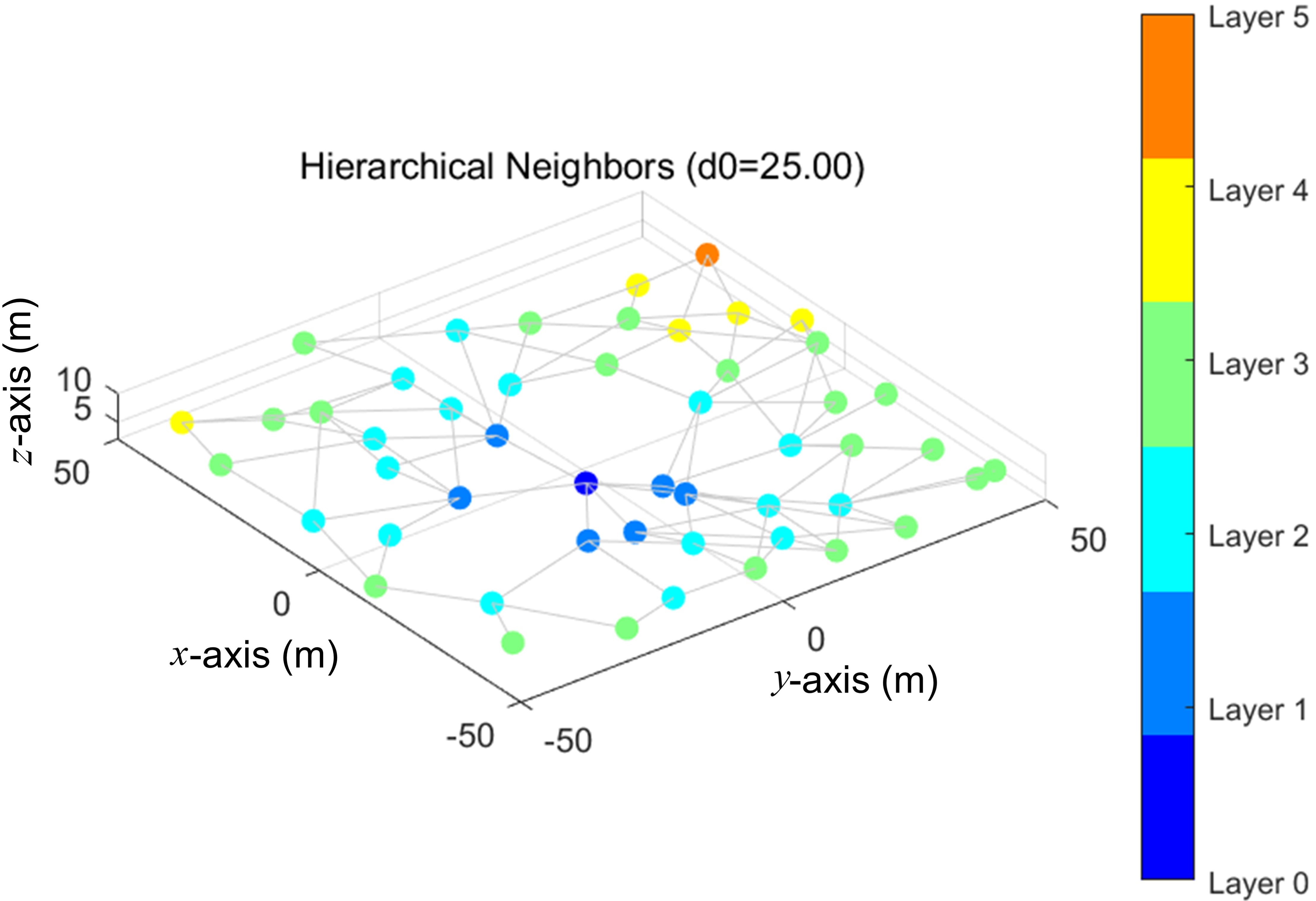}
		\caption{Measurement and communication topology with loop}\centering
		\label{S1_1}
		\vspace{-0.5cm}
	\end{figure}
	
	\subsection{Computation time and communication bandwidth}
	\begin{table}[h]
		\centering
			\caption{Performance Comparison on Different Hardware Platforms}
			\label{hardware performance}
			\begin{tabular}{c c c}
				\toprule
				\textbf{Metric} & \textbf{Laptop (X86-64)} & \textbf{Jetson Orin (ARM)} \\
				& \textbf{8 Cores, 2.7 GHz} & \textbf{8 Cores, 2.0 GHz} \\
				\midrule
				\begin{tabular}{c} Minimum program \\ loop time (s)\end{tabular} & 0.000144 & 0.000459 \\
				\midrule
				\begin{tabular}{c}Maximum program \\ loop time (s)\end{tabular} & 0.002904 & 0.008829 \\
				\midrule
				\begin{tabular}{c}Average program \\ loop time (s)\end{tabular} & 0.000160 & 0.000481 \\
				\midrule
				\begin{tabular}{c}Standard deviation \\ of program loop time (s)\end{tabular} & 0.000035 & 0.000089 \\
				\bottomrule
		\end{tabular}
	\end{table}
	A discussion on the algorithm's communication bandwidth requirements and computation time is presented in this section. 
	Taking robot $i$ as an example, in each algorithm iteration, the variable information of neighbor $j$ needs to be used. According to the methods mentioned in the paper, they roughly include: inter-robot distance $d_{ij}(t_{k})$ (1 float); odometer displacement of robot $j$ $\textbf{p}^{\mathcal{O}_{j}} _{j}(t_{k})$ (3 float); odometer angle displacement of robot $j$ $\phi_{\Sigma_{j}}^{\mathcal{O}_{j}}(t_{k})$ (1 float); hop count of robot $j$ (1 float); leader estimator $\hat{\textbf{q}}^{\mathcal{O}_{j}}_{j0}$ (3 float) and $\hat{\textbf{Q}}^{\mathcal{O}_{j}}_{\mathcal{O}_{0}}$ (2 float) of robot $j$; leader odometer information for multi hop propagation $v_{0,h}(t_{k})$ (1 float); $v_{0,z}(t_{k})$ (1 float); $w_{0}(t_{k})$ (1 float); $\textbf{p}^{\mathcal{O}_{0}} _{0}(t_{k})$ (3 float); $\phi_{\Sigma_{0}}^{\mathcal{O}_{0}}(t_{k})$ (1 float). 
	At time instant $t_{k}$, for robot $i$ and robot $j$, communication and interaction of $18 \times 4=64$ Byte of information is required. 
	Assuming that robot $i$ has 10 neighbors and the algorithm iteration frame rate is 20 frames per second (i.e., $\Delta t = 0.05$s, consistent with our simulation and experiment), the communication bandwidth required by robot i is $64 \times 10 \times 20/1024=12.5$kb/s. 
	According to the calculation results, it can be known that increasing the iteration frequency of the algorithm will require higher communication bandwidth. However, it should be pointed out that the algorithm execution frequency of 20 frames is generally sufficient to meet the localization and control requirements. 
	In practical use, the algorithm calculation frame rate can be appropriately increased under the condition of meeting the communication bandwidth.
	Besides, even if the swarm size is large, due to the distributed algorithm, it only needs to communicate and interact with some neighboring robots. The existing UWB module can fully meet this bandwidth.
	In addition, to verify the computational speed of the algorithm on the onboard measurement platform (jetson orin), we ran the same simulation experiment code simultaneously on a laptop and jetson platform, recording the time required for each iteration of the algorithm (including data collection, data filtering, auxiliary variable calculation, relative localization algorithm iteration, cooperative localization calculation, etc.). The specific results are shown in the table \ref{hardware performance}.
	
	\section{Supplemental simulation and experiment results}
	\label{supplemental}
	\subsection{Discussion on parameter setting}
	\blue{In the following, we present a detailed description of all parameters involved in the proposed method and clarify their physical meanings. Moreover, for all simulation and hardware experimental scenarios, we discuss the parameter selection strategy and analyze the sensitivity of the proposed method to these parameters.}
	
	For the parameters adopted in the cooperative localization and formation control:
	
	The parameters $w_{i}$ and $r_{i}$ denote the angular velocity and the radius of the horizontal circular motion to achieve localization enhancement for each robot. Considering that the accuracy of the UWB module used in the experiment is 0.1m, in order to reflect the significant changes in the relative distance during the motion, it is recommended that the sum of the circular motion diameters of the two neighboring robots be more than 10 times the accuracy error. 
		In our experiment, the radius $r_{i}$ is set to 0.3m for all robots, which can reduce the relative measurement error to a certain extent.
		In our response to your comment 4, we analyzed some situations where the state cannot be fully observed, and the conclusion can be summarized as follows: in order to achieve effective relative pose estimation, there needs to be a certain relative motion between neighboring robots, that is, there must be differences in the robot's motion speed and angular velocity. Taking this as a guide,  it is recommended that the speed and angular velocity of the neighbor localization enhancement motion differ by 0.5 to 2 times.
		Considering that the maximum linear velocity of the Turtlebot3 robot we used is 0.22m/s and the maximum angular velocity is 2.5rad/s, with a circular motion radius of 0.3m, we set the angular velocities of the neighboring robots to 0.6rad/s and 0.2rad/s respectively, which can effectively achieve relative motion between neighboring agents without exceeding the maximum linear velocity of the agents.
	
	Similarly, the parameter $c_{i,v}$ represents the amplitude and phase of longitudinal motion. 
		To achieve the relative motion of neighboring drones, the selection of this parameter is related to the actual flight altitude. 
		If the flight altitude is low, $c_{i,v}$ cannot be too large to prevent the drone from touching the ground during descent. 
		Specifically, the maximum downward motion amplitude of the drone can be calculated. It is recommended to widen the gradient of the $c_{i,v}$ parameter of neighboring drones while ensuring that the aircraft does not touch the ground. 
		For example, in our drone experiment, we set it to 0.1 and 0.3, respectively.
	
	\begin{figure}[!t]\centering
		\includegraphics[scale=1]{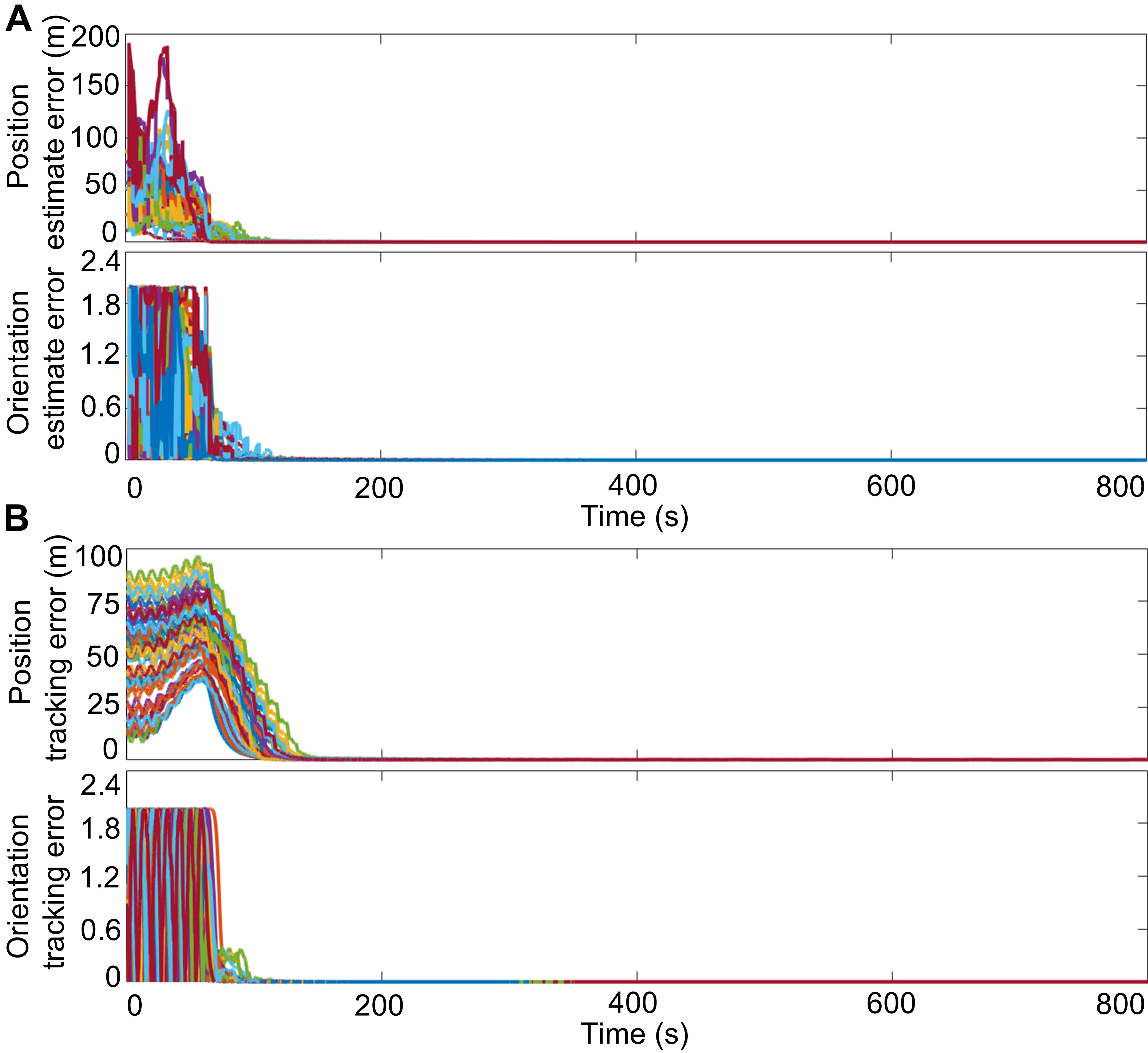}
		\caption{Pose estimation error and formation control error for each robot in the swarm. A: Estimation error. B: Formation control error. }\centering
		\label{S1_2}
		\vspace{-0.5cm}
	\end{figure}
	
	\begin{figure}[!t]\centering
		\includegraphics[scale=1]{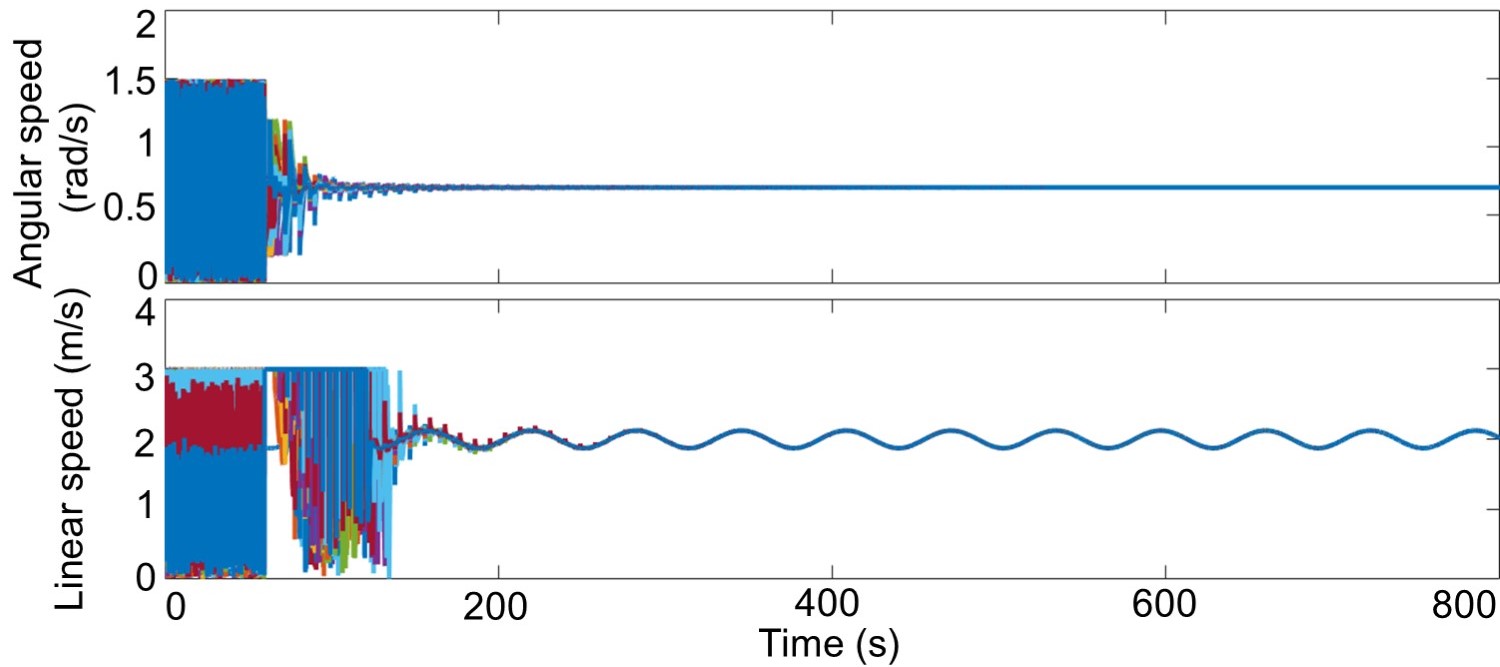}
		\caption{Speed and angle velocity for each robot in the swarm.}\centering
		\label{S1_3}
		\vspace{-0.5cm}
	\end{figure}
	
	The control gain $k_{1}$, $k_{2}$, $k_{3}$ and $k_{4}$ only needs to be greater than 0 to ensure asymptotic convergence of the control error. 
		Generally speaking, when the gain value is larger, the algorithm converges faster. 
		However, in computer control systems, due to the discrete control instructions (such as UAV control instructions generally ranging from 20 to 50HZ), divergence or numerical instability may occur when the gain is too large. 
		After multiple experimental verifications, we recommend that the gain $k_{1}$, $k_{2}$, $k_{3}$ and $k_{4}$ be between 1 and 2.
	\subsection{Motion trajectory of comparative simulation}
	When there are 5 robots in the swarm, based on our proposed method and the existing PE condition based method \cite{Liu2023Auto}, the motion trajectory of the robots is shown in Fig. \ref{S3_4}, which clearly demonstrates the advantages of the proposed data-driven method in improving control performance.
	
	\subsection{Formation control for large scale swarm}
	
	50 robots formed a random topology based on their initial positions, which included loops. The topology diagram is shown in Figure. \ref{S1_1}.
	The parameters adopted are the same with those in the multi-robot case in Section \ref{Performance of the PE-free relative localization method}. 
	The relative pose estimation error and formation control tracking error of all individuals in the swarm for themselves and the leader are shown in Fig. \ref{S1_2}(A) and Fig. \ref{S1_2}(B), and the velocity and angular velocity of all individuals in the swarm are shown in Fig. \ref{S1_3}. 
	It can be seen that the relative pose estimation error and tracking control error of the final individual converge, and the velocity and angular velocity tend to be a consistent value. (i.e., the velocity and angular velocity of the leader).
	
	\begin{figure}[!t]\centering
		\includegraphics[scale=1]{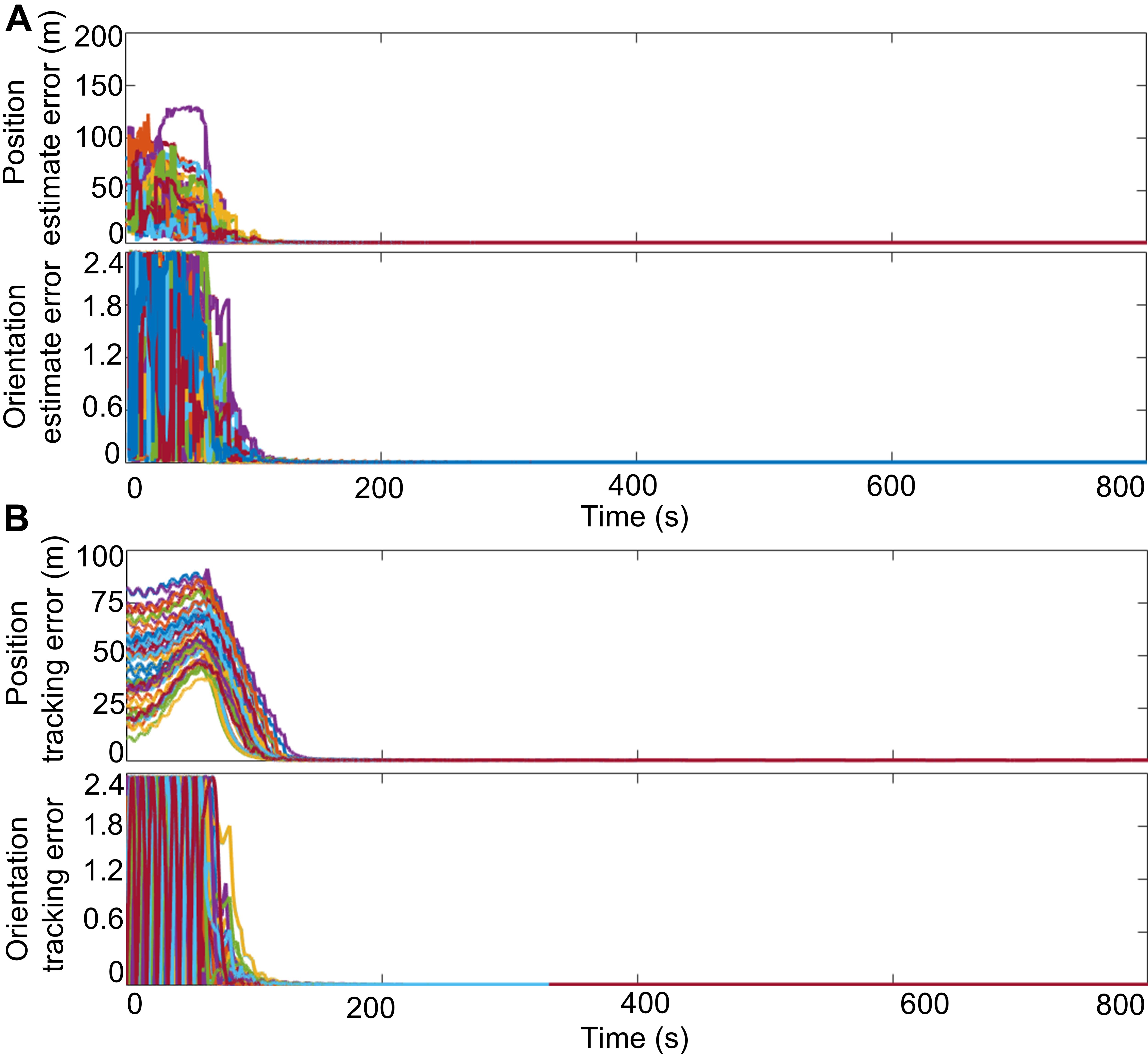}
		\caption{Pose estimation error and formation control error for each robot in the swarm with multi hop propagation of leader odometer measurement information. A: Estimation error. B: Formation control error. }\centering
		\label{S1_4}
		\vspace{-0.5cm}
	\end{figure}
	
	\begin{figure}[!t]\centering
		\includegraphics[scale=1]{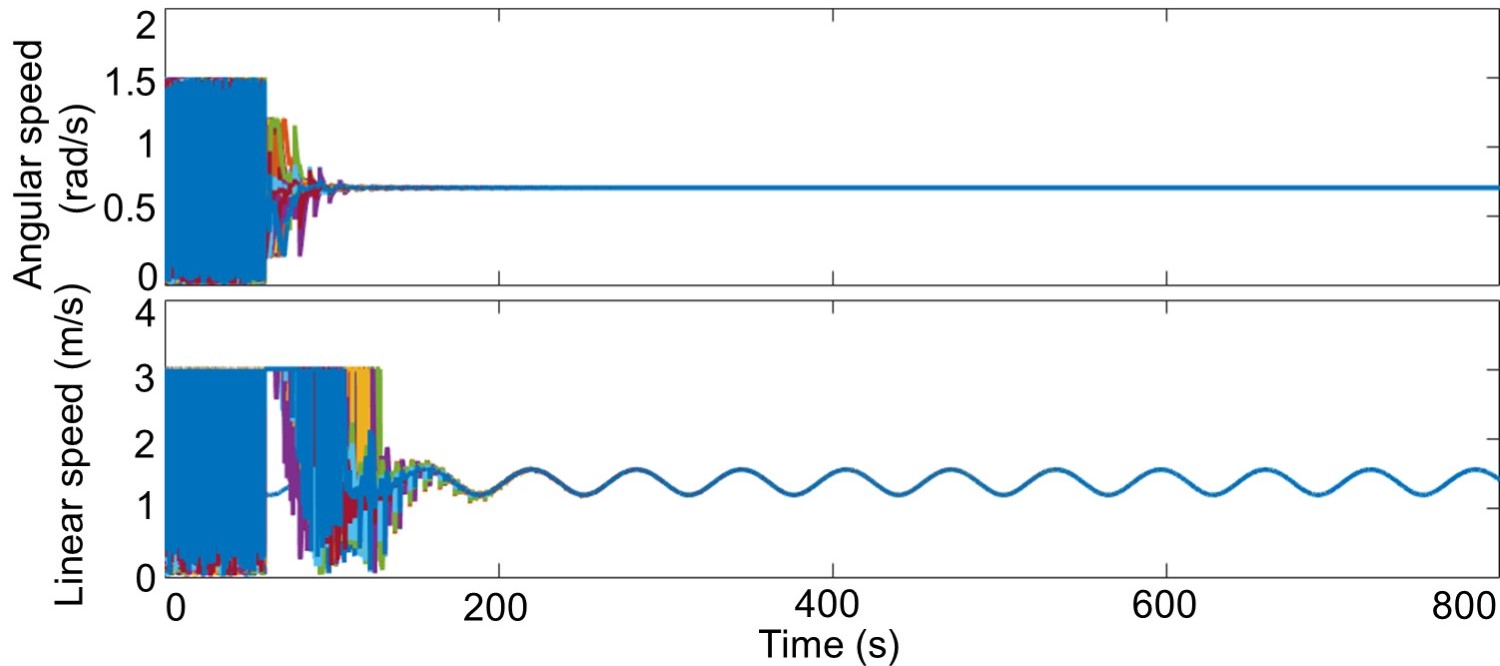}
		\caption{Speed and angle velocity for each robot with multi-hop propagation of leader odometer measurement information.}\centering
		\label{S1_5}
		\vspace{-0.5cm}
	\end{figure}
	
	In order to verify the method of multi hop propagation of leader odometer measurement information, we conducted simulation experiments in a random topology consisting of 50 robots. The experimental results are shown in Fig. \ref{S1_4} and Fig. \ref{S1_5}, which is basically consistent with directly using leader information (Fig. \ref{S1_2} and Fig. \ref{S1_3}). 
	This is because in engineering, a neighbor communication frame rate of 20 frames is sufficient for each robot to observe leader odometer information in a very short time (robot's own layers $\times$ 0.05 seconds).

	\subsection{Noise robustness for large scale swarm}
	
	\begin{figure}[!t]\centering
		\includegraphics[scale=1]{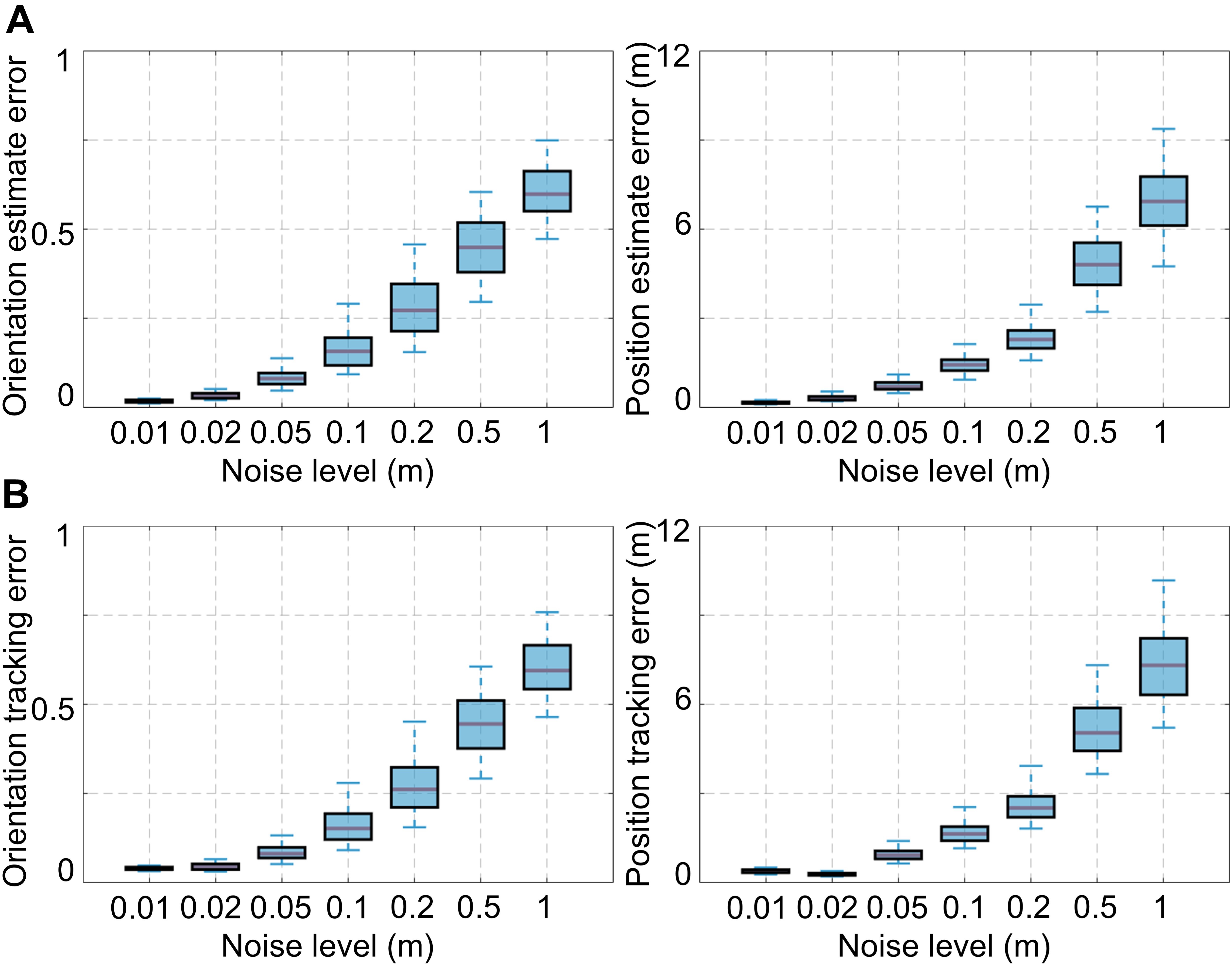}
		\caption{Pose estimation and tracking control error under different noise level. A: Estimation error. B: Tracking control error.}\centering
		\label{S2_1}
		\vspace{-0.5cm}
	\end{figure}
	
	\blue{We evaluate the cooperative localization error and formation control error of a 50-robot swarm under different noise conditions. }
		The parameters adopted are the same as those in the multi-robot case in Section \ref{Performance of the PE-free relative localization method}. 
		Specifically, under each noise condition, the algorithm is executed under 100 different initial conditions, and the statistical results of all robot estimation and control errors obtained in each simulation experiment are recorded as shown in Figure \ref{S2_1}. It can be seen that under normal noise levels (0.1m), the algorithm has good consistency with noise, but its performance decreases as the noise level increases. 
		In addition, due to the large size of the swarm, the error of cooperative localization may propagate with the increase of topology layers, but this can be solved through topology layering, adding leaders, and other methods, which can be discussed in future work.
	
	\subsection{Sensor highly inaccurate case}
	\begin{figure}[!t]\centering
		\includegraphics[scale=1]{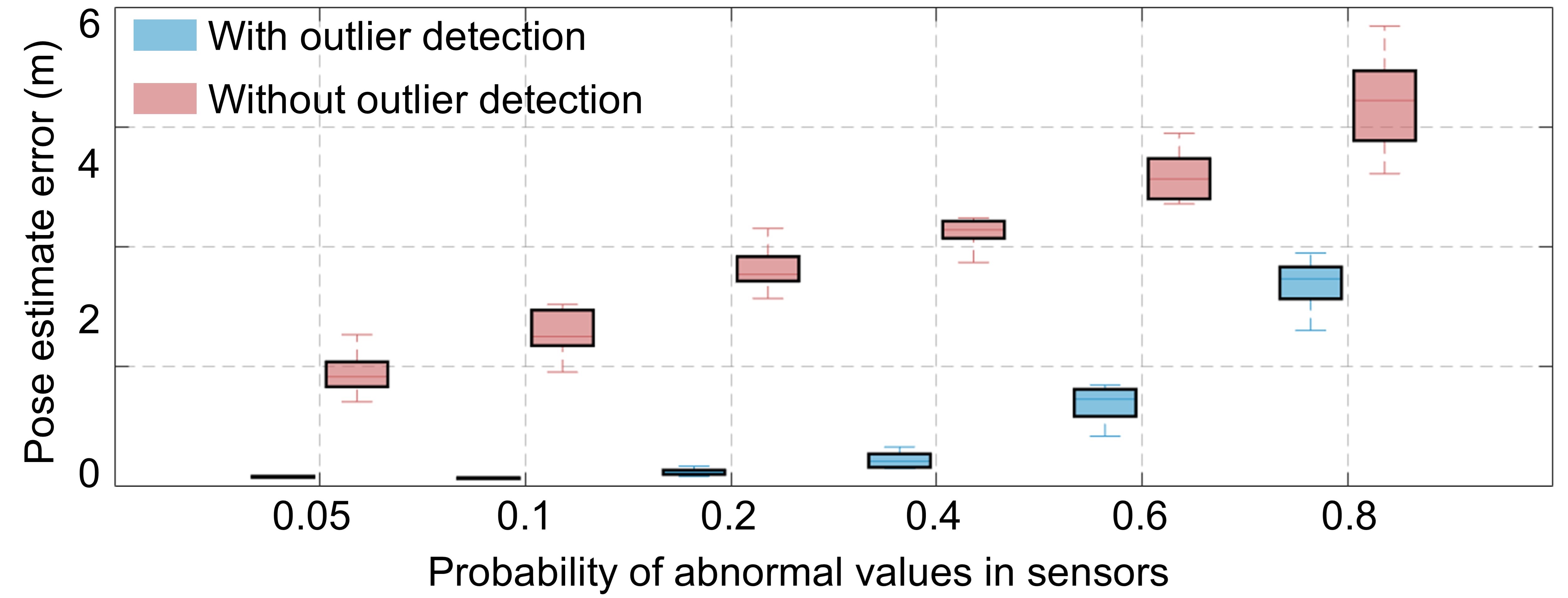}
		\caption{Accuracy of localization algorithm with and without outlier detection algorithm for sensors at different sensor failure rates.}\centering
		\label{S4_1}
		\vspace{-0.5cm}
	\end{figure}
	
	\blue{Based on the simulation setting in Section \ref{Outlier measurement value detection}, we further considered sensor faults in 100 Monte Carlo runs. Specifically, after a random time instant in each simulation, the sensor was assumed to become faulty and, for the remainder of the simulation, produced severely corrupted measurements with probability 0.95, modeled as noise with variance 5.}
		The parameters adopted are the same as those in the two-robot case in Section \ref{Performance of the PE-free relative localization method}. 
		Under such unfavorable conditions, the accuracy of the localization method is shown in Figure \ref{S4_1}. 
		It can be seen that the outlier detection algorithm can improve the accuracy of localization performance to a certain extent.

	\subsection{Formation control with dynamic topology}
\begin{figure}[!t]\centering
		\includegraphics[scale=1]{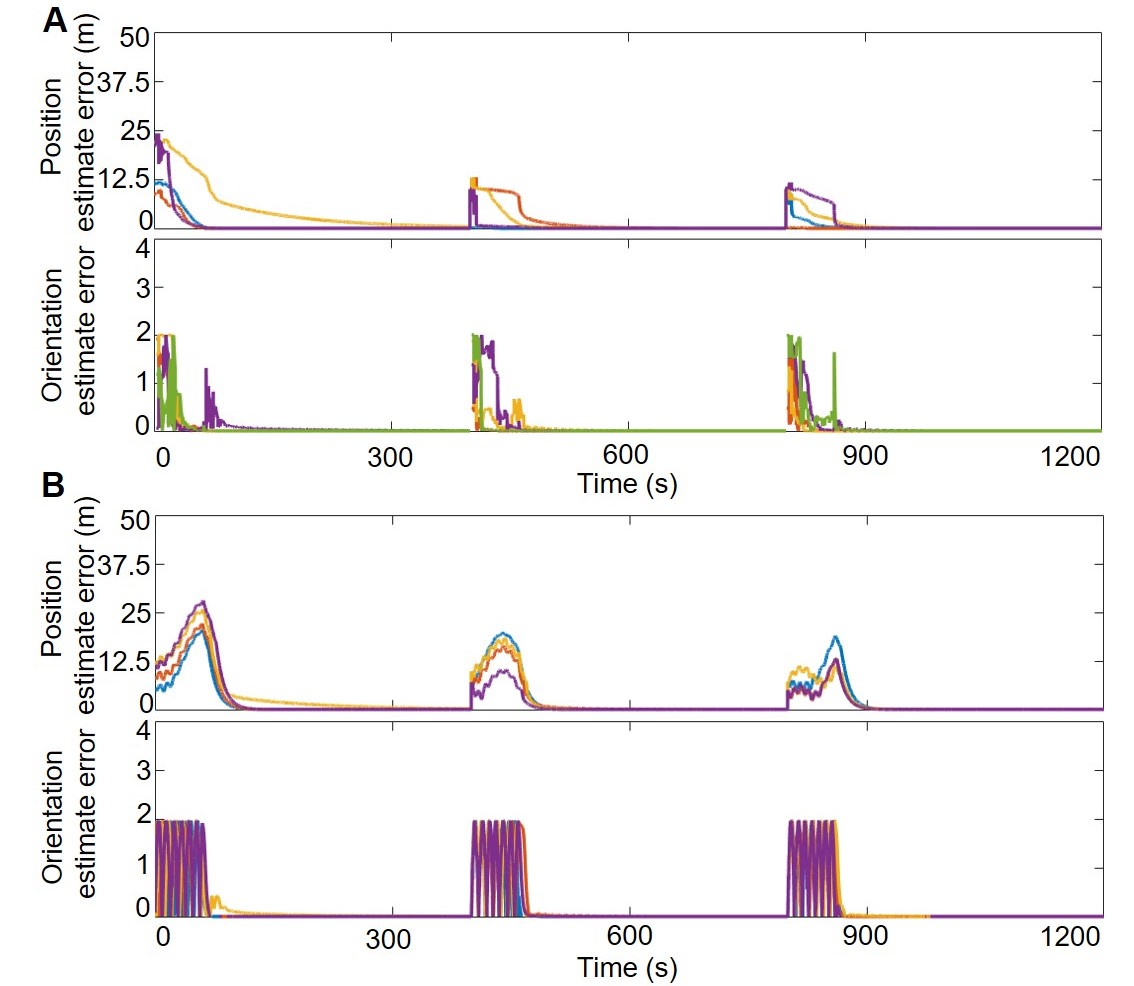}
		\caption{Pose estimation error and formation control error for each robot in the swarm under switched measurement topologies. A: Estimation error. B: Formation control error.}\centering
		\label{S6_1}
		\vspace{-0.5cm}
	\end{figure}

\begin{figure}[!t]\centering
		\includegraphics[scale=1]{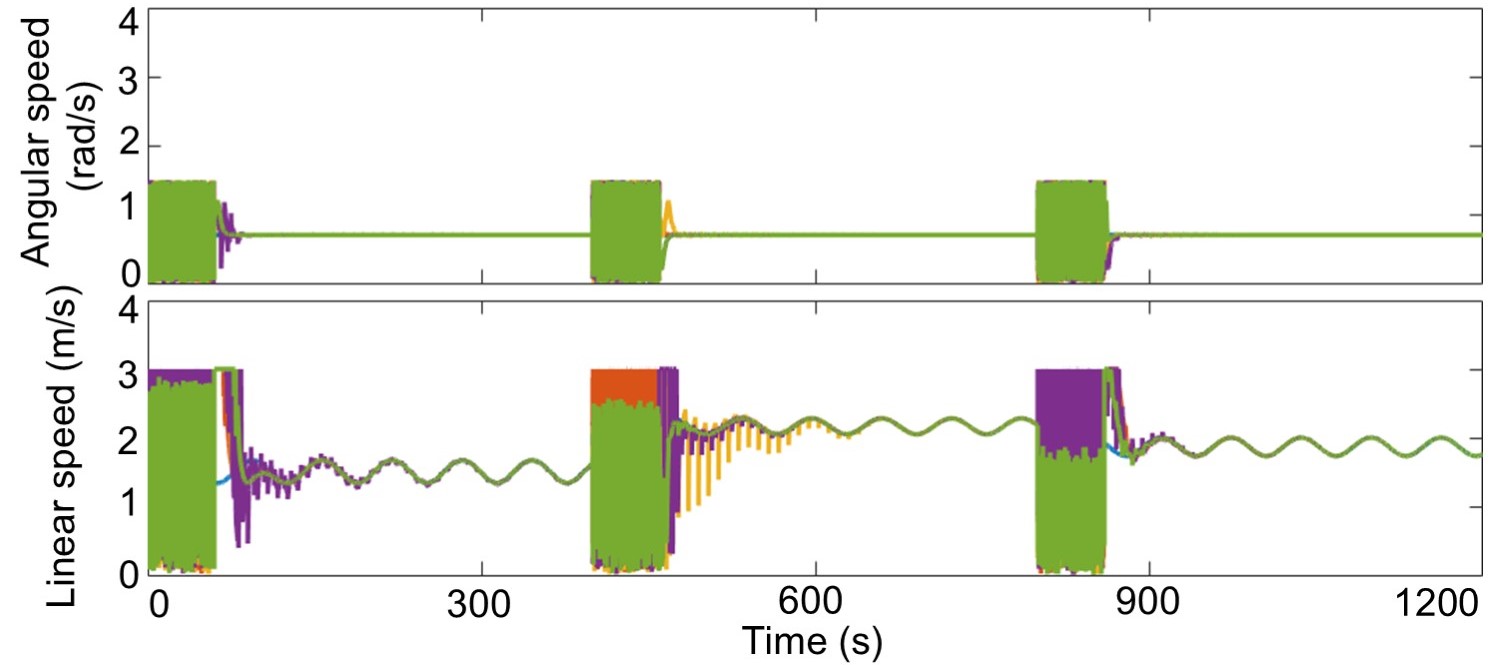}
		\caption{Velocity and angle velocity for each robot under switched measurement topologies.}\centering
		\label{S6_2}
		\vspace{-0.5cm}
	\end{figure}

	\blue{To address the dynamic topology issue, we conducted simulation studies under switching communication topologies. Specifically, 50 robots had random initial poses and switched measurement topologies at 400 seconds and 800 seconds, respectively. The experimental results are shown in Fig. \ref{S6_1} and Fig. \ref{S6_2}. It can be seen that after switching topologies, the robots repositioned their neighbors and carried out cooperative localization and distributed formation of leader robots based on this. The experimental results show that the algorithm converges quickly after topology dynamic switching.}
	
	\subsection{Formation control with different initial state}
\begin{figure}[!t]\centering
		\includegraphics[scale=1]{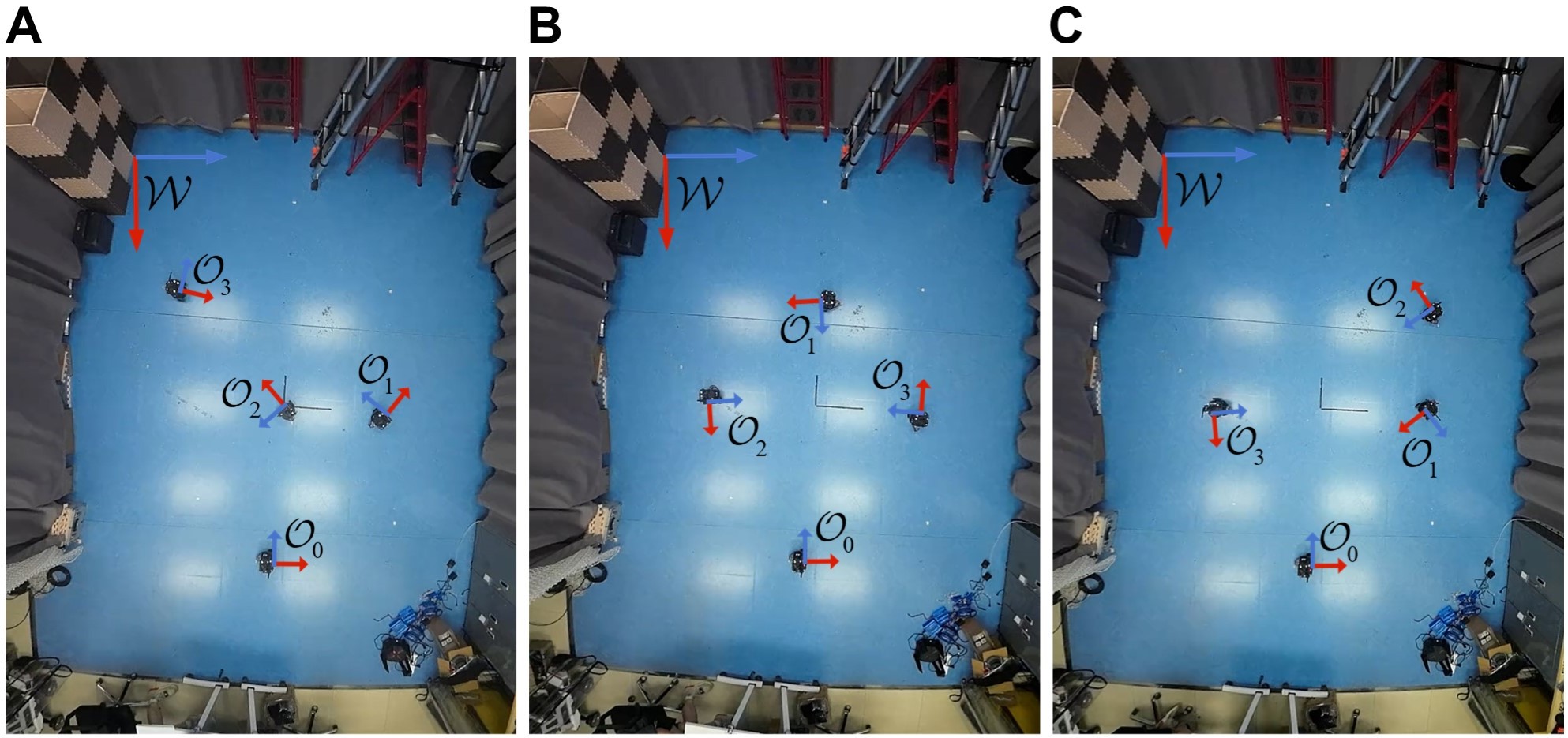}
		\caption{Snapshot of experiments starting from three different initial conditions in ten sets of experiments.}\centering
		\label{E3_2}
		\vspace{-0.5cm}
	\end{figure}

	The snapshot of experiments starting from three different initial conditions in ten sets of experiments is shown in Figure. \ref{E3_2}.

	\footnotesize
	\bibliographystyle{IEEEtran}
	\bibliography{IEEEabrv,references}

\begin{thebibliography}{10}
\providecommand{\url}[1]{#1}
\csname url@samestyle\endcsname
\providecommand{\newblock}{\relax}
\providecommand{\bibinfo}[2]{#2}
\providecommand{\BIBentrySTDinterwordspacing}{\spaceskip=0pt\relax}
\providecommand{\BIBentryALTinterwordstretchfactor}{4}
\providecommand{\BIBentryALTinterwordspacing}{\spaceskip=\fontdimen2\font plus
\BIBentryALTinterwordstretchfactor\fontdimen3\font minus
  \fontdimen4\font\relax}
\providecommand{\BIBforeignlanguage}[2]{{%
\expandafter\ifx\csname l@#1\endcsname\relax
\typeout{** WARNING: IEEEtran.bst: No hyphenation pattern has been}%
\typeout{** loaded for the language `#1'. Using the pattern for}%
\typeout{** the default language instead.}%
\else
\language=\csname l@#1\endcsname
\fi
#2}}
\providecommand{\BIBdecl}{\relax}
\BIBdecl

\bibitem{Sun2025IEM}
G.~Sun, J.~L{\"u}, K.~Liu, Z.~Wang, and G.~Chen, ``Mean-shift theory and its
  applications in swarm robotics: A new way to enhance the efficiency of
  multi-robot collaboration,'' \emph{IEEE Ind. Electron. Mag.}, 2025.

\bibitem{Zhao2023TM}
J.~Zhao, K.~Zhu, H.~Hu, X.~Yu, X.~Li, and H.~Wang, ``Formation control of
  networked mobile robots with unknown reference orientation,'' \emph{IEEE/ASME
  Trans. Mechatron.}, vol.~28, no.~4, pp. 2200--2212, 2023.

\bibitem{xu2023cooperative}
L.~Xu, X.~Cao, W.~Du, and Y.~Li, ``Cooperative path planning optimization for
  multiple uavs with communication constraints,'' \emph{Knowledge-Based Syst.},
  vol. 260, p. 110164, 2023.

\bibitem{li2024urban}
Y.~Li, T.~Guo, J.~Chen, J.~Wu, Y.~Zhang, S.~Alam, K.~Cai, and W.~Du, ``Urban
  air mobility: Review and challenges,'' \emph{IEEE Intell. Transp. Syst.
  Mag.}, 2024.

\bibitem{Ze2023TIE}
K.~Ze, W.~Wang, K.~Liu, and J.~L{\"u}, ``Time-varying formation planning and
  distributed control for multiple uavs in clutter environment,'' \emph{IEEE
  Trans. Ind. Electron.}, vol.~71, pp. 11\,305--11\,315, 2023.

\bibitem{Chu2023TASE}
W.~Chu, W.~Zhang, H.~Zhao, Z.~Jin, and H.~Mei, ``Massive shape formation in
  grid environments,'' \emph{IEEE Trans. Auto. Sci. Eng.}, vol.~20, no.~3, pp.
  1745--1759, Jul. 2023.

\bibitem{Sun2023NC}
G.~Sun, R.~Zhou, Z.~Ma, Y.~Li, R.~Gro{\ss}, Z.~Chen, and S.~Zhao, ``Mean-shift
  exploration in shape assembly of robot swarms,'' \emph{Nat. Commun.},
  vol.~14, no.~1, p. 3476, Jun. 2023.

\bibitem{Zhao2019TAC}
S.~Zhao, Z.~Li, and Z.~Ding, ``Bearing-only formation tracking control of
  multiagent systems,'' \emph{IEEE Trans. Autom. Control}, vol.~64, no.~11, pp.
  4541--4554, 2019.

\bibitem{Wang2024Auto}
H.~Wang, X.~Zhao, S.~Huang, Q.~Li, and Y.~Liu, ``A branch-and-bound based
  globally optimal solution to 2d multi-robot relative pose estimation
  problems,'' \emph{Automatica}, vol. 164, p. 111654, 2024.

\bibitem{Ning2024IJRR}
Z.~Ning, Y.~Zhang, J.~Li, Z.~Chen, and S.~Zhao, ``A bearing-angle approach for
  unknown target motion analysis based on visual measurements,'' \emph{Int. J.
  Robot. Res.}, p. 02783649241229172, 2024.

\bibitem{Xie2023TRO}
T.~H. Nguyen and L.~Xie, ``Relative transformation estimation based on fusion
  of odometry and uwb ranging data,'' \emph{IEEE Trans. Robot.}, vol.~39,
  no.~4, pp. 2861--2877, 2023.

\bibitem{Xie2019TCNS}
K.~Cao, Z.~Qiu, and L.~Xie, ``Relative docking and formation control via range
  and odometry measurements,'' \emph{IEEE Trans. Control Netw. Syst.}, vol.~7,
  no.~2, pp. 912--922, 2019.

\bibitem{dong2025tmech}
W.~Dong, S.~Chen, Z.~Mei, Y.~Ying, and X.~Zhu, ``Dwe-based sribo: An efficient
  and resilient single-range and inertial-based odometry with dimension-reduced
  wriggling estimator,'' \emph{IEEE/ASME Trans. Mechatron.}, pp. 1 -- 13, 2025.

\bibitem{Liu2023Auto}
Y.~Liu and Z.~Liu, ``Distributed adaptive formation control of multi-agent
  systems with measurement noises,'' \emph{Automatica}, vol. 150, p. 110857,
  2023.

\bibitem{xiong2025adaptive}
K.~Xiong, S.~Chen, and W.~Dong, ``An adaptive sliding window estimator for
  positioning of unmanned aerial vehicle using a single anchor,'' \emph{IEEE
  Sens. J.}, 2025.

\bibitem{Chen2023TIE}
S.~Chen, Y.~Li, and W.~Dong, ``High-performance relative localization based on
  key-node seeking considering aerial drags using range and odometry
  measurements,'' \emph{IEEE Trans. Ind. Electron.}, 2023.

\bibitem{Liu2023RAL}
J.~Liu and G.~Hu, ``Relative localization estimation for multiple robots via
  the rotating ultra-wideband tag,'' \emph{IEEE Robot. Autom. Lett.}, 2023.

\bibitem{Mehdifar2020PPB}
F.~Mehdifar, C.~P. Bechlioulis, F.~Hashemzadeh, and M.~Baradarannia,
  ``Prescribed performance distance-based formation control of multi-agent
  systems,'' \emph{Automatica}, vol. 119, p. 109086, 2020.

\bibitem{Chen2024TCNS}
J.~Chen, B.~Jayawardhana, and H.~G. de~Marina, ``Distributed distance-based
  formation-motion control of unicycle agents without orientation
  measurements,'' \emph{IEEE Trans. Control Netw. Syst.}, 2024.

\bibitem{lv2022TNNLS}
Y.~Lv, H.~Zhang, Z.~Wang, and H.~Yan, ``Distributed localization for
  multi-agent systems with random noise based on iterative learning,''
  \emph{IEEE Trans. Neural Netw. Learn. Syst.}, vol.~35, no.~1, pp. 952--960,
  2022.

\bibitem{Cossette2022IROS}
C.~C. Cossette, M.~A. Shalaby, D.~Saussi{\'e}, J.~Le~Ny, and J.~R. Forbes,
  ``Optimal multi-robot formations for relative pose estimation using range
  measurements,'' in \emph{Proc. IEEE/RSJ Int. Conf. Intell. Robots Syst.
  (IROS)}.\hskip 1em plus 0.5em minus 0.4em\relax IEEE, 2022, pp. 2431--2437.

\bibitem{Shalaby2024TRO}
M.~A. Shalaby, C.~C. Cossette, J.~Le~Ny, and J.~R. Forbes, ``Multi-robot
  relative pose estimation and imu preintegration using passive uwb
  transceivers,'' \emph{IEEE Trans. Robot.}, 2024.

\bibitem{cano2023TRO}
J.~Cano and J.~Le~Ny, ``Ranging-based localizability optimization for mobile
  robotic networks,'' \emph{IEEE Trans. Robot.}, vol.~39, no.~4, pp.
  2842--2860, 2023.

\bibitem{Ze2025TASE}
J.~L{\"u}, K.~Ze, S.~Yue, K.~Liu, W.~Wang, and G.~Sun, ``Concurrent-learning
  based relative localization in shape formation of robot swarms,'' \emph{IEEE
  Trans. Auto. Sci. Eng.}, vol.~22, pp. 11\,188--11\,204, 2025.

\bibitem{silic2019correcting}
M.~Silic and K.~Mohseni, ``Correcting current-induced magnetometer errors on
  uavs: an online model-based approach,'' \emph{IEEE Sens. J.}, vol.~20, no.~2,
  pp. 1067--1076, 2019.

\bibitem{Wang2023TM}
Y.~Wang, M.~Lin, X.~Xie, Y.~Gao, F.~Deng, and T.~L. Lam, ``Asymptotically
  efficient estimator for range-based robot relative localization,''
  \emph{IEEE/ASME Trans. Mechatron.}, vol.~28, no.~6, pp. 3525--3536, 2023.

\bibitem{Ma2024TCST}
L.~Yan, B.~Ma, Y.~Jia, and Y.~Jia, ``Observer-based trajectory tracking control
  of nonholonomic wheeled mobile robots,'' \emph{IEEE Trans. Control Syst.
  Technol.}, 2024.

\bibitem{Zou2024Auto}
Y.~Zou, L.~Zhong, W.~He, and C.~Silvestre, ``Leader--follower circumnavigation
  control of non-holonomic robots using distance-related information,''
  \emph{Automatica}, vol. 169, p. 111831, 2024.

\bibitem{Chowdhary2010CDC}
G.~Chowdhary and E.~Johnson, ``Concurrent learning for convergence in adaptive
  control without persistency of excitation,'' in \emph{in Proc. 49th IEEE
  Conf. Decis. Control (CDC)}.\hskip 1em plus 0.5em minus 0.4em\relax IEEE,
  2010, pp. 3674--3679.

\bibitem{Djaneye2019TAC}
O.~Djaneye-Boundjou and R.~Ord{\'o}{\~n}ez, ``Gradient-based discrete-time
  concurrent learning for standalone function approximation,'' \emph{IEEE
  Trans. Autom. Control}, vol.~65, no.~2, pp. 749--756, 2019.

\bibitem{muhlegg2012concurrent}
M.~M{\"u}hlegg, G.~Chowdhary, and E.~Johnson, ``Concurrent learning adaptive
  control of linear systems with noisy measurements,'' in \emph{AIAA Guidance,
  Navigation, and Control Conference}, 2012, p. 4669.

\bibitem{Cao2023TC}
K.~Cao, M.~Cao, and L.~Xie, ``Similar formation control via range and odometry
  measurements,'' \emph{IEEE T. Cybern.}, 2023.

\bibitem{zhu2024self}
W.~Zhu, S.~O{\u{g}}uz, M.~K. Heinrich, M.~Allwright, M.~Wahby, A.~L.
  Christensen, E.~Garone, and M.~Dorigo, ``Self-organizing nervous systems for
  robot swarms,'' \emph{Science Robotics}, vol.~9, no.~96, p. eadl5161, 2024.

\bibitem{cao2023similar}
K.~Cao, M.~Cao, and L.~Xie, ``Similar formation control via range and odometry
  measurements,'' \emph{IEEE Transactions on Cybernetics}, 2023.

\bibitem{Zhao2022TRO}
J.~Li, Z.~Ning, S.~He, C.-H. Lee, and S.~Zhao, ``Three-dimensional bearing-only
  target following via observability-enhanced helical guidance,'' \emph{IEEE
  Trans. Robot.}, vol.~39, no.~2, pp. 1509--1526, 2022.

\bibitem{Ze2024Arxiv}
J.~L{\"u}, K.~Ze, S.~Yue, K.~Liu, W.~Wang, and G.~Sun, ``Concurrent-learning
  based relative localization in shape formation of robot swarms,'' \emph{arXiv
  preprint arXiv:2410.06052}, 2024.

\bibitem{Zhu2023TCNS}
S.~Zhu, K.~Lv, Z.~Yang, C.~Chen, and X.~Guan, ``Bearing-based formation
  tracking control of nonholonomic mobile agents with a persistently exciting
  leader,'' \emph{IEEE Trans. Control Netw. Syst.}, vol.~11, no.~1, pp.
  307--318, 2023.

\bibitem{Xie2020TRO}
T.-M. Nguyen, Z.~Qiu, T.~H. Nguyen, M.~Cao, and L.~Xie, ``Persistently excited
  adaptive relative localization and time-varying formation of robot swarms,''
  \emph{IEEE Trans. Robot.}, vol.~36, no.~2, pp. 553--560, 2019.

\end{thebibliography}
	
\end{document}